\pdfoutput=1
\documentclass{article}

\usepackage[T1]{fontenc}
\usepackage{sspt_arxiv,times}
\usepackage{hyperref}
\usepackage{url}
\usepackage{graphicx}
\usepackage{amsmath,amssymb,amsthm}
\usepackage{booktabs}
\usepackage{longtable}
\usepackage{array}
\usepackage{microtype}
\usepackage{xcolor}
\usepackage{colortbl}
\usepackage{placeins}
\usepackage{wrapfig}
\usepackage{needspace}

\makeatletter
\newcommand{\captionof}[1]{\def\@captype{#1}\caption}
\makeatother

\newlength{\ssptSavedIntextsep}
\definecolor{protocolbg}{gray}{0.96}
\definecolor{protocoltext}{gray}{0.42}
\definecolor{ssptcolumnbg}{RGB}{242,238,248}

\newcommand{\sspt}{SSPT}
\newcommand{\ssio}{SSIO}
\newcommand{\rplanssi}{RPLAN-SSI}
\newcommand{\ipd}{IPD}
\newcommand{\pia}{PIA}
\newcommand{\piad}{PIAD}
\newcommand{\areaguard}{AreaGuard}
\newcommand{\method}{\sspt{}-PPO}

\newcommand{\housediffusion}{HouseDiffusion}

\title{Rethinking Floorplan Generation \\ with Space Syntax: From Design Theory to Model Improvement}

\author{%
  Zhuoyang Jiang\textsuperscript{1,2}\quad
  Dongqing Zhang\textsuperscript{2,\ensuremath{\dagger}}\quad
  Qingxiang Liu\textsuperscript{1}\quad
  Yuxuan Liang\textsuperscript{1,\ensuremath{\dagger}}\\[2pt]
  \normalfont\small\textsuperscript{1}Systems Hub, The Hong Kong University of Science and Technology (Guangzhou)\\
  \normalfont\small\textsuperscript{2}College of Architecture and Urban Planning, Tongji University
}
\newcommand{\arxivAuthorNote}{%
  \begingroup
  \renewcommand{\thefootnote}{\fnsymbol{footnote}}%
  \footnotetext[2]{Corresponding authors.}%
  \endgroup
}

\hypersetup{
  hidelinks,
  pdftitle={Rethinking Floorplan Generation with Space Syntax: From Design Theory to Model Improvement},
  pdfauthor={Zhuoyang Jiang, Dongqing Zhang, Qingxiang Liu, Yuxuan Liang},
  pdfsubject={Preprint: space syntax-guided evaluation and post-training for floorplan generation}
}

\begin{document}

\maketitle
\arxivAuthorNote

\begin{abstract}
Architectural design increasingly adopts generative AI (GenAI) for floorplan generation, an early-stage design task exploring how spaces should be organized for use. Yet fidelity objectives inherited from native visual and geometric generators do not automatically capture this design purpose. This organization remains largely implicit in generated representations, limiting direct evaluation and feedback. Drawing on space syntax, an established architectural design theory, the Space Syntax Integration Operator (\ssio{}) recovers convex-space access graphs across heterogeneous formats. This theory-grounded shared intermediate representation, which we call a design container, supports evaluation and model understanding of spatial design. Applying \ssio{} to four residential corpora spanning 15 countries reveals public spaces' integration levels are highest, with corpus-dependent distributions. These patterns underpin the \rplanssi{} benchmark, an empirical design reference system assessing population-level functional-profile alignment (\ipd{}) and distributional alignment of within-plan public-space advantage (\piad{}). Comparison with this residential reference reveals a design gap: public spaces in generated floorplans have a smaller integration advantage over other spaces. To address this gap, Space Syntax-Guided Post-Training (\sspt{}) provides a plug-and-play design-learning interface: \ssio{}-derived RL feedback reduces \ipd{} and \piad{} across five heterogeneous diffusion backbones while preserving native mechanisms. We further discuss how this approach may inform GenAI applications in other design domains.
\looseness=-1
\end{abstract}

\section{Introduction}

Floorplan generation brings generative AI (GenAI) into architectural design, where the goal is to create spatial layouts that support human use. This task supports early-stage exploration of spatial organization: how spaces support everyday activities, shared use, and privacy~\citep{wu2019rplan,park2024quality}. Many existing methods inherit the training objectives of their underlying GenAI models, improving visual and geometric quality; architectural requirements typically enter as input conditions~\citep{hu2020graph2plan,nauata2021houseganpp,shabani2023housediffusion}. Outputs use diverse visual and geometric formats, while the spatial organization that matters for use remains largely implicit. Common evaluation metrics assess visual similarity, geometric validity, and constraint satisfaction~\citep{regenwetter2023metrics,zeng2025floorplanmetrics}, but do not directly measure how well spaces are organized for use (Figure~\ref{fig:overview}(a)). For example, a connected generated floorplan can give a bathroom higher overall accessibility than its shared living space (Figure~\ref{fig:selected-cases-main}, B). Quantitative design theory and residential evidence accumulated in architecture let us test design-directed improvements across generators while retaining their original visual fidelity objectives.

Recent work begins to assess design performance: FloorPlan-LLaMa learns architectural judgments from expert preferences~\citep{yin2025floorplanllama}. Its evaluator, however, predicts design scores without explicitly representing the spatial objects and relations to be measured. Architectural design theories offer a way to make these objects and relations explicit and explain why they matter for use. Recovering these objects and relations across formats would make spatial organization directly measurable, providing a common architectural basis for identifying design gaps, setting improvement objectives, and guiding learning across heterogeneous generators.

\begingroup
\setlength{\intextsep}{4pt plus 1pt minus 1pt}
\begin{figure}[!h]
  \setlength{\abovecaptionskip}{4pt}
  \setlength{\belowcaptionskip}{2pt}
  \centering
  \includegraphics[width=0.94\textwidth]{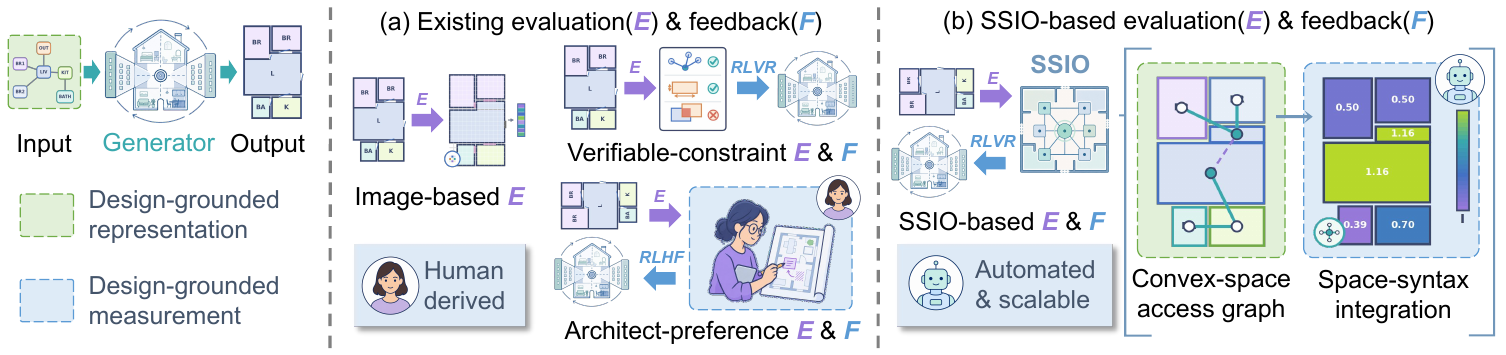}
  \caption{\textbf{Towards Design-directed Representation and Measurement in Floorplan Generation.}}
  \label{fig:overview}
\end{figure}
\endgroup

For this purpose, we draw on space syntax, an established architectural design theory linking spatial organization to building use: each space's role depends on its relations within the overall layout~\citep{hillier1984sociallogic,hillier1996space,hanson1998decoding}. It represents interiors as convex-space access graphs (CSAGs): convex spaces are nodes, access relations are edges, and integration measures relative accessibility. Our Space Syntax Integration Operator (\ssio{}) recovers CSAGs from heterogeneous sources and computes integration scores (Figures~\ref{fig:overview}(b) and~\ref{fig:ssio-branches}). This yields a shared intermediate representation (IR) of design elements and relations---which we term a \textbf{theory-defined design container}---and directly computable spatial measurements across generators.\looseness=-1

To ground these measurements empirically, we apply \ssio{} to four residential corpora spanning 15 countries~\citep{wu2019rplan,lifullhome,vanengelenburg2025msd,ziaee2024sfca68}. Public spaces have the highest integration levels in each corpus (Figure~\ref{fig:real-corpus-alignment}). This Public-first ordering grounds Public anchoring---Public spaces' relative integration advantage---while differing distributions require references matched to residential context and task. Accordingly, the RPLAN-derived Space Syntax Integration benchmark (\rplanssi{}) uses RPLAN's Chinese residences as its reference: Integration Profile Distance (\ipd{}) measures population-level functional-profile alignment, and Public-Anchor Integration Advantage Distance (\piad{}) compares distributions of within-plan Public advantage. A room-count-matched comparison of the output population from a generator pretrained on RPLAN against this corpus reference reveals preserved Public-first ordering but a substantially smaller Public integration margin over other functional domains (Figure~\ref{fig:matched-generator-gap}). These analyses establish an \textbf{empirical design reference system} for benchmarking generators and identifying design gaps.\looseness=-1

To close this design gap, we introduce Space Syntax-Guided Post-Training (\sspt{}), using grounded measurements to improve heterogeneous generators. For diffusion floorplan generators~\citep{shabani2023housediffusion,wang2025diffplanner,wang2026ce2eplan}, this requires connecting measurements of completed layouts to the native sampling decisions that produce them. Through backbone-specific policy bindings, \sspt{} applies a shared proximal policy optimization (PPO) objective~\citep{schulman2017ppo} with \ssio{}-derived terminal rewards to native reverse policies over selected lower-noise transitions (Figure~\ref{fig:sspt-ppo}). Across five diffusion backbones, both measures improve; primary results show 27.0\% lower \ipd{} and 32.2\% lower \piad{} within native-quality budgets. A connected case illustrates the gains: shared living-space anchoring replaces bathroom dominance under an unchanged room program (Figure~\ref{fig:selected-cases-main}). Shared feedback and native-policy bindings thus form a \textbf{reusable design-learning interface} for plug-and-play post-training without replacing native mechanisms.\looseness=-1

To situate this approach within GenAI for design, we review 55 studies across 14 design categories through three maps comparing design representations and criteria, where feedback acts, and cross-generator reuse. The maps highlight a distinctive connection: theory-defined measurements and empirical references guide heterogeneous-generator post-training through a shared design container. We distill this connection into a three-element framework and executable harness as a reference for design-theory translation in other domains (Section~\ref{sec:design-learning-principles}; Appendices~\ref{app:literature-review} and~\ref{app:design-learning-principles}).

Our contributions translate architectural design theory into a representation--evaluation--learning pathway for floorplan generation: \textbf{(1)} \ssio{} makes spatial organization explicit and directly measurable across heterogeneous formats, providing a unified design container for evaluation and feedback. \textbf{(2)} \rplanssi{} establishes an empirical spatial-organization benchmark, revealing weaker Public anchoring in generated layouts than in real floorplans. \textbf{(3)} \sspt{} introduces a reusable design-learning interface---to our knowledge, the first plug-and-play RL post-training framework for diffusion floorplan generators---improving both metrics across five backbones without replacing native mechanisms.\looseness=-1

\pagebreak
\section{Related Work}

\paragraph{Floorplan generation.}
Floorplan generators combine architectural conditioning with synthesis in different output spaces. Graph2Plan and House-GAN++ realize requested layouts through box prediction and mask generation, whereas HouseDiffusion denoises room polygons~\citep{hu2020graph2plan,nauata2021houseganpp,shabani2023housediffusion}. Boundary-focused and language-conditioned diffusion vary geometric representations and user controls~\citep{wang2025diffplanner,hu2025gsdiff,qin2024chathousediffusion}. Requested relations do not determine the access hierarchy realized after synthesis. \ssio{} measures this realized organization across native output formats.

\paragraph{Architectural measurement and evaluation.}
Distributional similarity, geometric validity, and condition satisfaction assess fidelity and compliance~\citep{shabani2023housediffusion,zeng2025floorplanmetrics,park2024quality}. Space syntax supplies analytical objects and accessibility measures for studying how spatial configuration supports use~\citep{hillier1984sociallogic,hillier1996space,hanson1998decoding}. Its applications include design transformation and analysis of visibility or connectivity in generated layouts~\citep{eloy2016transforming,heo2023visibility,man2026smallspace}. Our challenge is to recover a consistent analytical structure across formats and establish references for interpreting its measurements. \ssio{} and \rplanssi{} connect this recovery to residential evidence: theory specifies the property, while real designs ground numerical comparisons of the resulting spatial organization.

\paragraph{Feedback and post-training.}
FloorPlan-LLaMa learns architect-preference rewards, while verifiable-reward RL checks connectivity and area; both update autoregressive generators~\citep{yin2025floorplanllama,lara2026floorplanrlvr}. These approaches bring design judgment or specified requirements into learning. GreenPlanner targets energy and feasibility in a flow generator~\citep{zeng2026greenplanner}. \sspt{} instead targets access organization with directly computed architectural measurements. Diffusion alignment provides differentiable-reward optimization and policy gradients~\citep{clark2024draft,black2024ddpo}; we adopt policy gradients because geometric recovery and access-graph analysis do not supply reward gradients. Backbone-specific likelihood bindings let the shared feedback update native diffusion policies across different output representations and generation mechanisms.

Our broader comparison examines how design criteria are represented, where they affect learning, and how they transfer across generators (Appendix~\ref{app:literature-review}). Domain-purpose optimization and multi-backbone adaptation have precedents~\citep{zheng2023urbanplanning,lai2026posterreward,wu2024uicoder}; this study connects theory-defined analysis, empirical references, and heterogeneous-generator post-training. Section~\ref{sec:design-learning-principles} develops this combination into a reusable theory-to-learning construction workflow.

\section{Design Representation \& Evaluation: Space Syntax System}
\label{sec:measurement}

Floorplan generation synthesizes room layouts from design requirements: generator \(m\) takes native conditions \(c\) and outputs plan \(x_m^{\mathrm{nat}}\). Existing methods often focus on visual and geometric fidelity, while architectural design must organize shared use, privacy, and everyday activities through underlying spatial relations. This requires explicitly recovering the plan's underlying access structure from different visual and geometric representations. SSIO provides a common measurement for this reexamination, enabling residential reference construction and identification of generation gaps. Appendix~\ref{app:abbreviations} collects terminology and mathematical notation.

\subsection{Space Syntax Integration Operator (SSIO)}
\label{sec:ssio}
\begin{figure}[!h]
  \centering
  \includegraphics[width=0.92\textwidth]{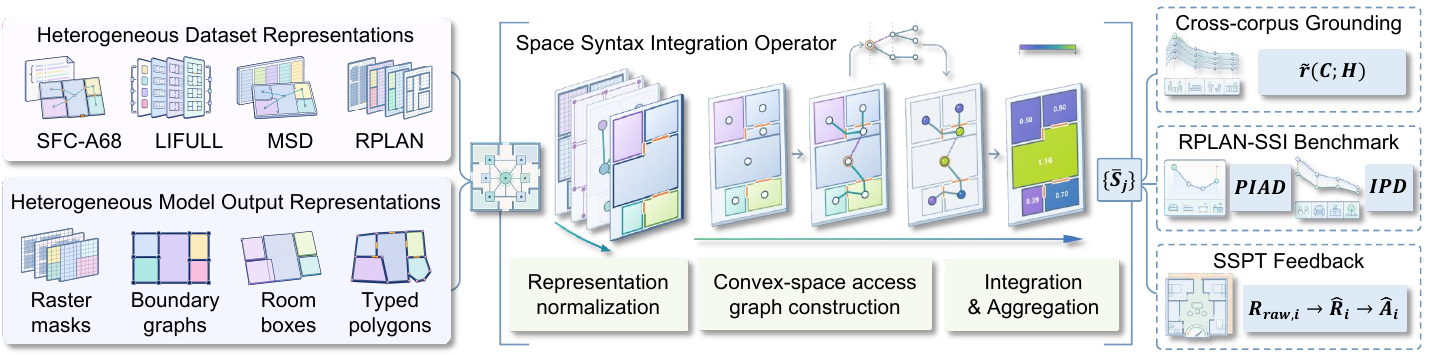}
  \caption{\textbf{Space Syntax Integration Operator (SSIO).} Recovering shared theory-defined IR.}
  \label{fig:ssio-branches}
\end{figure}
SSIO recovers CSAGs across output formats and measures their spatial organization. Its outputs serve three uses: probing real and generated layouts for functional patterns and gaps, constructing RPLAN-SSI benchmark measurements, and supplying design rewards to SSPT (Section~\ref{sec:sspt}). The computation has three steps (Figure~\ref{fig:ssio-branches}; worked example in Appendix Figure~\ref{fig:ssio-worked}).

First, normalization maps each generator's native output into room regions, labels, and interior-door incidence: \(z_m=\mathcal A_m(x_m^{\mathrm{nat}};c)=((Q_{m,j},y_{m,j})_{j=1}^{R_m},D_m^{\mathrm{int}})\). A frozen completion map supplies missing incidence, \(\widehat D_m^{\mathrm{int}}=\mathcal C_\phi(x_m^{\mathrm{nat}},c)\). Second, SSIO decomposes walkable regions into convex-atom sets \(V_j\) and represents internal access through connections between atoms. Retaining \(\mathcal R_m^+=\{j:|V_j|>0\}\), it connects atoms by within-room continuity and interior doors to obtain \(G_x=(V_x,E_x)\), with \(V_x=\biguplus_{j\in\mathcal R_m^+}V_j\). The input graph \(G_c\) specifies intended relations; \(G_x\) records access realized in the output.
Third, a scoring rule assigns each atom a finite, nonnegative accessibility score \(S_v\) from graph distances. Averaging atoms within rooms gives \(\bar S_j\), then averaging rooms within category \(k\) gives \(\bar S_{p,k}\), making the graph measurement available for functional comparison. We primarily use component-local Legacy Hillier--Hanson (HH) integration as the node scoring rule \(S_v=I_v\) (Hillier and Hanson, 1984; Hillier, 1996), yielding \(\bar I_j\) and \(\bar I_{p,k}\); Appendix Eq.~\eqref{eq:primary-hh} specifies the calculation and finite boundary conventions.

\subsection{RPLAN-derived Space Syntax Integration Benchmark (RPLAN-SSI)}
\label{sec:rplanssi}

To identify residential functional patterns, we aggregate categories into Public, Private, Service, and Outdoor domains \(\mathcal G\), normalized by their within-plan mean (Appendix~\ref{app:data}).
For plan \(p\), \(\mathcal H_p\subseteq\mathcal G\) contains the nonempty domains and \(\mathcal K_{p,g}\) their retained categories. The domain score \(S_{p,g}\) averages category scores \(\bar S_{p,k}\), weighted by retained room counts. With \(\epsilon_{\mathrm{norm}}=10^{-6}\), we normalize these domain scores within each plan and collect their domain-wise medians across cohort \(\mathcal C\), requiring nonempty support for each median:
\begin{equation}
r^{\mathrm{dom}}_{p,g}(S)
=\frac{S_{p,g}}
{\epsilon_{\mathrm{norm}}+\operatorname{mean}_{h\in\mathcal H_p}S_{p,h}},
\qquad
\widetilde{\mathbf r}(\mathcal C;S)
=\left(
\operatorname*{median}_{p\in\mathcal C:\,g\in\mathcal H_p}
r^{\mathrm{dom}}_{p,g}(S)
\right)_{g\in\mathcal G}.
\label{eq:functional-profile}
\end{equation}
The left expression compares domains within one layout; the right summarizes them across layouts. To probe Public's position in an individual layout, the Public-domain margin \(\Delta_{\mathrm{pub}}(p)=r^{\mathrm{dom}}_{p,\mathrm{Public}}(S)-\max_{g\in\mathcal H_p\setminus\{\mathrm{Public}\}}r^{\mathrm{dom}}_{p,g}(S)\) measures its lead over the strongest other domain. For layouts containing Public and another domain, a positive value indicates first rank and its magnitude records the lead. We omit \(S\) from this notation below.

\Needspace{14\baselineskip}
\begin{wrapfigure}[14]{r}{0.60\textwidth}
  \centering
  \includegraphics[width=\linewidth]{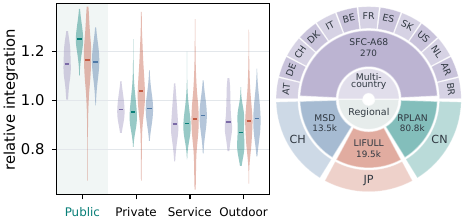}
  \caption{\textbf{Cross-Corpus Grounding: Public-first.}}
  \label{fig:real-corpus-alignment}
\end{wrapfigure}
\textbf{Cross-Corpus Grounding of Functional Regularities.} To ground a functional design target in residential evidence, we apply SSIO to RPLAN (China), LIFULL (Japan), MSD (Switzerland), and SFC-A68 (13 countries), retaining layouts with annotated Public space. Global harmonic centrality \(S_v=H_v\) measures whole-graph accessibility, with zero contributions from unreachable spaces (Appendix Eq.~\eqref{eq:global-harmonic}). Public has the highest domain median in each corpus (1.148–1.254), exceeding the next-highest median by 0.127–0.303 (Figure~\ref{fig:real-corpus-alignment}). This shared ordering and measurable lead motivate Public anchoring; differing distributions call for context-matched numerical references.

\textbf{Configurational Metrics and Evaluation.} Building on the observed residential patterns and space syntax's functional integration analysis~\citep{hillier1996space}, RPLAN-SSI reframes generator evaluation around realized functional organization. Its context-specific reference contains 20,348 retained exact-eight RPLAN layouts, \(\mathcal C_{\mathrm{ref}}\), matching generated room counts. RPLAN's predominant Public category is Living, which implements the benchmark's single-plan Public anchor. Each model attempts each condition once; measurable outputs retaining eight spatial instances and this anchor enter cohort \(\mathcal C\). The benchmark uses Legacy HH integration.
Public-Anchor Integration Advantage (PIA) measures a layout's normalized Living-category lead over its strongest eligible competitor. For plan \(p\), \(\mathcal K_p\), \(\mathcal K_{\mathrm{norm}}\), and \(\mathcal V\) denote retained, normalization, and eligible competitor categories, and \(\mu_p=\epsilon_{\mathrm{norm}}+\operatorname{mean}_{k\in\mathcal K_{\mathrm{norm}}\cap\mathcal K_p}\bar I_{p,k}\). \(F_{\mathcal C}\) and \(F_{\mathcal C_{\mathrm{ref}}}\) denote empirical PIA distributions in generated and reference layouts, respectively; \(W_1\) denotes Wasserstein-1 distance:
\begin{equation}
\mathrm{IPD}=\frac{\|\widetilde{\mathbf r}(\mathcal C)-\widetilde{\mathbf r}(\mathcal C_{\mathrm{ref}})\|_1}{|\mathcal G|},
\;
\mathrm{PIA}(p)=\frac{\bar I_{p,\mathrm{Living}}-\max\limits_{k\in\mathcal V\cap\mathcal K_p}\bar I_{p,k}}{\mu_p},
\;
\mathrm{PIAD}=W_1(F_{\mathcal C},F_{\mathcal C_{\mathrm{ref}}}).
\label{eq:metrics}
\end{equation}
Positive PIA means Living leads; negative PIA indicates a deficit. IPD compares typical four-domain profiles (Eq.~\eqref{eq:functional-profile}); PIAD compares distributions of Living's advantage. Lower values indicate closer residential alignment. PIA and its median explain individual and typical advantages (Appendix~\ref{app:metric-support}).

\Needspace{14\baselineskip}
\begin{wrapfigure}[14]{r}{0.50\textwidth}
  \centering
  \includegraphics[width=\linewidth]{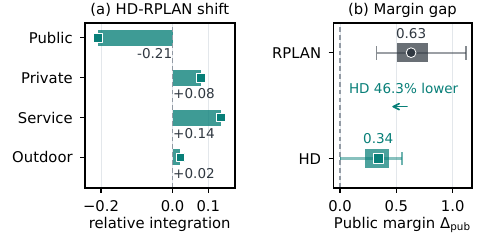}
  \caption{\textbf{Preserved Public-first Ordering, Weaker Functional Hierarchy.}}
  \label{fig:matched-generator-gap}
\end{wrapfigure}
\textbf{Design Gap between Generated and Real Floorplans.} To test whether generated layouts reproduce this organization, we compare HouseDiffusion (HD) with RPLAN under matched room counts and acceptance rules. HD preserves Public-first ordering but weakens the functional hierarchy (Figure~\ref{fig:matched-generator-gap}(a)): its median Public-domain margin is 46.3\% lower (0.3387 versus 0.6304), and its 95th percentile remains below RPLAN's median (Figure~\ref{fig:matched-generator-gap}(b)). The positive-margin fraction differs by only 4.37 percentage points. Thus, the gap concerns the strength of Public's advantage, beyond its rank, motivating the post-training target in Section~\ref{sec:sspt}.

\section{Design Learning: Space Syntax-Guided Post-Training (SSPT)}
\label{sec:sspt}

\begin{figure}[t]
  \centering
  \includegraphics[width=0.93\textwidth]{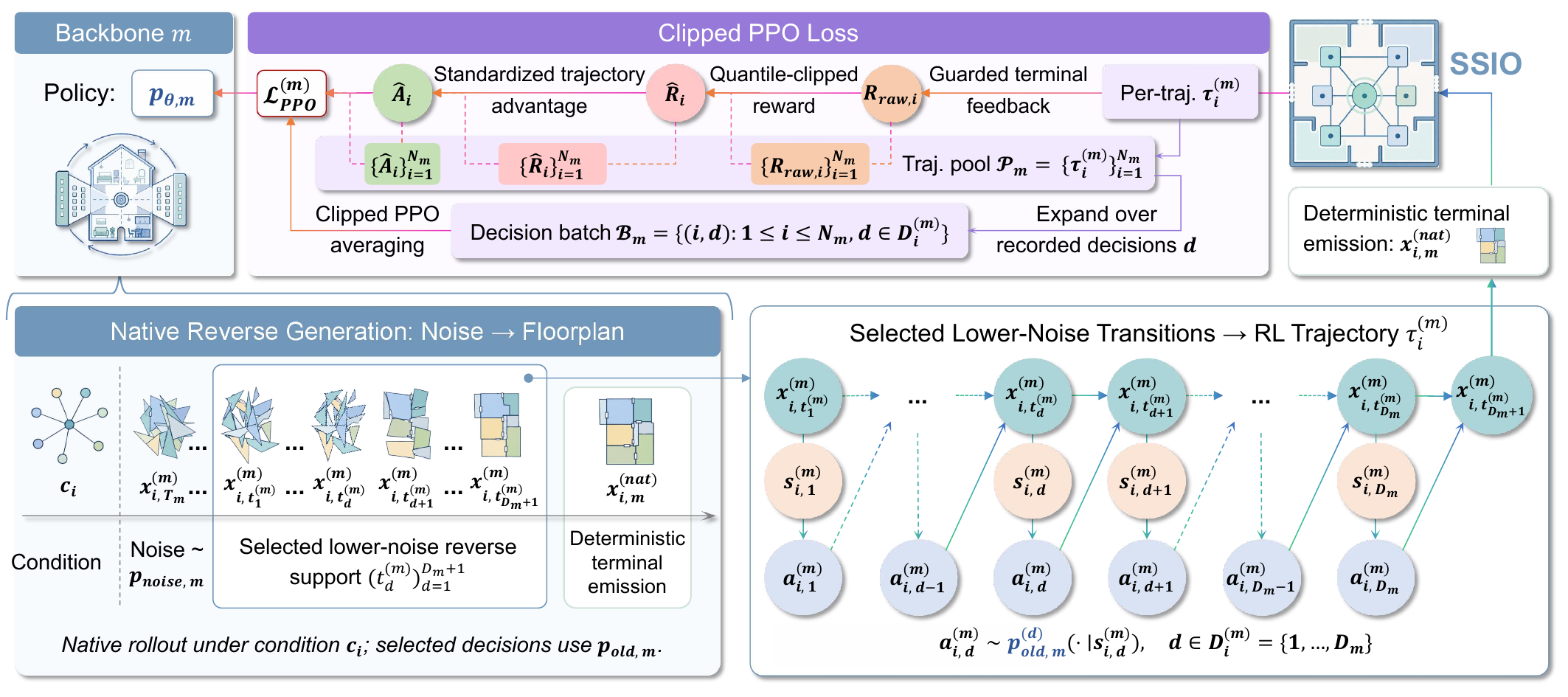}
  \caption{\textbf{SSPT-PPO trajectory-space post-training.} Guarded \ssio{} rewards are clipped and standardized over rollout pools; the resulting trajectory advantages drive PPO updates on backbone-native lower-noise transitions. The Native Reverse Generation panel illustrates HD room-corner denoising; Appendix~\ref{app:backbone-bindings} describes the native generation mechanisms of all five diffusion backbones.}
  \label{fig:sspt-ppo}
\end{figure}

\method{} turns \ssio{} measurements into terminal feedback, sharing architectural rewards and a PPO objective across native reverse policies. Learning through policy likelihoods lets discrete spatial analysis guide generator updates without differentiating through geometric recovery (Figure~\ref{fig:sspt-ppo}).

\textbf{From Selection to Signed Feedback.} \sspt{}-Iter post-trains in data space by retraining on \ssio{}-ranked real and generated examples~\citep{dong2023raft,ferbach2024selfconsuming}. The pilot retains no generated candidates, incurring sampling and scoring costs without learning from generated designs (Appendix~\ref{app:sspt-iter-pilot}). \method{} iterates in trajectory space, using signed advantages to reinforce or suppress decisions according to relative terminal rewards.

\textbf{Guarded Terminal Feedback.} We translate the empirical Public-anchor relation into the room-level Living integration lead reward; \ipd{} and \piad{} evaluate the resulting cohorts. Fix backbone \(m\) and its training measurement settings, suppressing \(m\) in reward terms. For a reward-valid output \(v_m(x)=1\), let \(\mathcal R_{\mathrm{Living}}(x)\) and \(\mathcal R_{\neg\mathrm{Living}}(x)\) contain retained Living and non-Living rooms. Define \(R_{\mathrm{anchor}}(x)=\max_{j\in\mathcal R_{\mathrm{Living}}(x)}\bar I_j -\max_{j\in\mathcal R_{\neg\mathrm{Living}}(x)}\bar I_j\), taking empty maxima as zero. A one-sided area-expansion penalty (\areaguard{}) controls the expansion shortcut exposed by optimization with only Living integration lead: \(R_{\mathrm{AreaGuard}}(x)=-[\operatorname{Area}(x)-q_A]_+/s_A\), where \([u]_+=\max(u,0)\), \(\operatorname{Area}(x)\) is total retained-atom area, and frozen Base outputs fix threshold \(q_A\) and scale \(s_A>0\). The \areaguard{} weight \(\lambda\geq0\) sets the trade-off between anchoring and expansion control:
\begin{equation}
R_{\mathrm{raw}}^{(\lambda)}(x)
=8R_{\mathrm{anchor}}(x)+\lambda R_{\mathrm{AreaGuard}}(x),
\qquad v_m(x)=1.
\label{eq:guard}
\end{equation}
For \(v_m(x)=0\), set \(R_{\mathrm{raw}}^{(\lambda)}(x)=-1\). Appendix~\ref{app:backbone-bindings} specifies measurement settings, validity, and frozen guard statistics. Below, \(R_{\mathrm{raw}}=R_{\mathrm{raw}}^{(\lambda)}\) uses the chosen weight.

\textbf{Native Reverse Policies and Rollouts.} Each backbone's selected lower-noise reverse segment defines a finite-horizon Markov decision process with decisions \(\mathcal D_i^{(m)}=\{1,\ldots,D_m\}\) at decreasing native times \(t_1^{(m)}>\cdots>t_{D_m+1}^{(m)}\). State \(s_{i,d}^{(m)}\) contains \(x_{i,t_d^{(m)}}^{(m)}\), its time index, condition \(c_i\), and the fixed masks or embeddings that define its native generation context.
Joint action \(a_{i,d}^{(m)}\) comprises the sampled active stochastic coordinates of \(x_{i,t_{d+1}^{(m)}}^{(m)}\), with fixed coordinates restored before the next step. Outside the stochastic decision set, the rollout checkpoint completes generation deterministically to obtain native output \(x_{i,m}^{\mathrm{nat}}\).
At iteration \(r\), \(p_{\mathrm{old},m}=p_{\theta_r,m}\) samples a fixed trajectory pool \(\mathcal P_m=\{\tau_i^{(m)}\}_{i=1}^{N_m}\). Terminal outputs receive \(R_{\mathrm{raw},i}=R_{\mathrm{raw}}^{(\lambda)}(x_{i,m}^{\mathrm{nat}})\) from Eq.~\eqref{eq:guard}; PPO holds outputs and rewards fixed during replay and draws a fresh pool from the updated policy at the next iteration.

\textbf{Clipped Trajectory Optimization.} Let \(\mathcal B_m=\{(i,d):1\leq i\leq N_m,\, d\in\mathcal D_i^{(m)}\}\) be the decision batch. Rollouts store \(\ell_{\mathrm{old},i,d}^{(m)}=\log p_{\mathrm{old},m}^{(d)} (a_{i,d}^{(m)}\mid s_{i,d}^{(m)})\); PPO recomputes \(\ell_{\theta,i,d}^{(m)}=\log p_{\theta,m}^{(d)} (a_{i,d}^{(m)}\mid s_{i,d}^{(m)})\) over each binding's active stochastic coordinates. For backbone \(m\), log-ratio bound \(M_m\) stabilizes the likelihood ratio between the current and rollout policies:
\begin{equation}
\widetilde\rho_{i,d}^{(m)}
=\exp\!\left(
\operatorname{clip}\!\left[
\ell_{\theta,i,d}^{(m)}-\ell_{\mathrm{old},i,d}^{(m)},
-M_m,M_m
\right]\right).
\label{eq:diffusion-ratio}
\end{equation}
Clipping terminal rewards at pool quantiles \(q_{0.01},q_{0.99}\) gives \(\widehat R_i=\operatorname{clip}(R_{\mathrm{raw},i},q_{0.01},q_{0.99})\). The detached trajectory advantage is \mbox{\(\widehat A_i=(\widehat R_i-\mu_{\widehat R})/(\sigma_{\widehat R}+10^{-8})\)}, using pool mean \(\mu_{\widehat R}\) and population standard deviation \(\sigma_{\widehat R}\). For PPO clip radius \(\epsilon\), the shared surrogate loss is
\begin{equation}
\mathcal L_{\mathrm{PPO}}^{(m)}
=-\mathbb E_{(i,d)\sim\operatorname{Unif}(\mathcal B_m)}\!\left[
\min\!\left(
\widetilde\rho_{i,d}^{(m)}\widehat A_i,
\operatorname{clip}(\widetilde\rho_{i,d}^{(m)},1-\epsilon,1+\epsilon)
\widehat A_i\right)
\right].
\label{eq:ppo}
\end{equation}
All recorded decisions in a trajectory share \(\widehat A_i\); reward quantiles and moments remain fixed during replay. This reinforcement learning with verifiable rewards (RLVR) uses computable architectural measurements to ground model improvement in design theory. Appendix~\ref{app:backbone-bindings} specifies native policy bindings, reverse-time supports, and training settings.

\FloatBarrier
\section{Experiments}
\label{sec:experiments}

\textbf{Experimental design and protocol.} We first compare spatial organization across eight generators using the RPLAN-SSI benchmark, with one attempt per condition on five disjoint blocks of 1,000 eight-room requests. Among all 5,000 attempts, we accept measurable outputs retaining exactly eight spatial instances, including Living, pool them to compute IPD and PIAD, and report their proportion among all attempts (i.e., Exact-8 acceptance). Appendix~\ref{app:block-stability} additionally reports the means and standard deviations of block IPDs, PIADs, and PIA medians. We then test whether SSPT improves HD, the highest-acceptance generator in this benchmark, within quality limits calibrated from evaluations of HD before post-training (Base). SSPT post-training uses at-most-seven-room requests; evaluation on the same eight-room suite tests generalization beyond the room counts used for post-training. Matched comparisons assess HD's design metrics before and after SSPT under identical input conditions, case studies interpret spatial changes, and controlled ablations examine update location, design reward, and expansion control. To test the SSPT framework's applicability beyond HD, we post-train four further diffusion backbones through their native policy bindings, examining gains and adaptation requirements across output representations. GSDiff's boundary-based outputs require access completion for SSIO measurement, which we examine as a separate deployment case. Appendices~\ref{app:data}, \ref{app:backbone-bindings}, and~\ref{app:baselines} detail the data and protocols.

\Needspace{18\baselineskip}
\subsection{Unified Benchmark Evaluation}
\label{sec:benchmark-results}

\setlength{\ssptSavedIntextsep}{\intextsep}
\setlength{\intextsep}{0pt}
\begin{wraptable}{r}{0.51\textwidth}
\centering
\setlength{\abovecaptionskip}{0pt}
\caption{\textbf{\rplanssi{} evaluation.}}
\label{tab:benchmark}
\setlength{\tabcolsep}{2pt}
\begin{tabular}{lcc>{\columncolor{protocolbg}\color{protocoltext}}c}
\toprule
Method & \ipd{} $\downarrow$ & \piad{} $\downarrow$ & Acc. (\%) \\
\midrule
CE2EPlan     & 0.1914 & 0.5816 & 66.5 \\
DiffPlanner  & 0.1803 & 0.5548 & 59.1 \\
House-GAN++  & 0.1612 & 0.5288 & 40.7 \\
House-GAN    & 0.1384 & 0.4136 & 36.9 \\
Graph2Plan   & 0.1078 & 0.3974 & 71.6 \\
FloorplanGAN & 0.0943 & 0.4340 & 91.5 \\
ChatHD & \textbf{0.0866} & \textbf{0.3517} & 31.2 \\
HD & 0.1111 & 0.4129 & \textbf{94.5} \\
\bottomrule
\end{tabular}
\par
\end{wraptable}
RPLAN-SSI compares spatial organization across generators with different synthesis mechanisms and output formats (Table~\ref{tab:benchmark}; Figure~\ref{fig:results-overview}(a)). As shown, diffusion models occupy both ends of the eight-model design-score ranking, motivating direct assessment of spatial organization even within one model family. ChatHouseDiffusion (ChatHD), whose diffusion output is a semantic raster, leads both metrics on accepted outputs but achieves only 31.2\% acceptance. Polygon-based HD achieves 94.5\% acceptance, while Section~\ref{sec:rplanssi} shows its smaller Public-domain margin relative to real residences (Figure~\ref{fig:matched-generator-gap}). This broad coverage and documented design gap support its use for controlled post-training experiments under native-quality constraints. Figure~\ref{fig:results-overview}(a) additionally includes GSDiff-SSIO: because native GSDiff outputs cannot directly support SSI evaluation, we develop a dedicated measurement adaptation (Appendix~\ref{app:gsdiff-completion}). We compare SSPT post-training effects on this adapted system and the other diffusion backbones.
\par\WFclear
\setlength{\intextsep}{\ssptSavedIntextsep}

\begingroup
\setlength{\intextsep}{4pt}
\setlength{\abovecaptionskip}{4pt}
\setlength{\belowcaptionskip}{0pt}
\begin{figure}[!htbp]
  \centering
  \includegraphics[width=\textwidth,trim=0 3pt 0 3pt,clip]{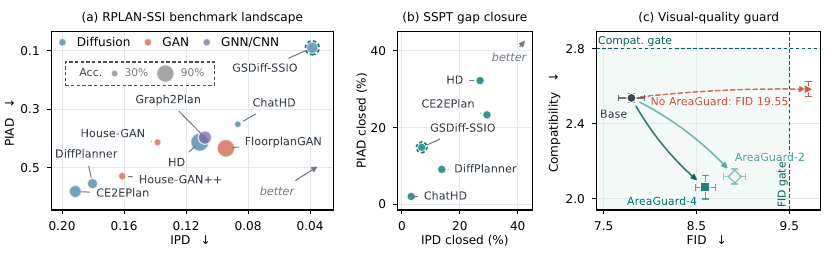}
  \caption{\textbf{Design evaluation and post-training.} (a) Model families and Exact-8 acceptance (bubble area). (b) Gap closure to zero \ipd{} and \piad{}. (c) \areaguard{} weight comparison: independent Base evaluation, Base-calibrated limits (dashes), and standard deviations (error bars).}
  \label{fig:results-overview}
\end{figure}
\endgroup

\subsection{Design Improvement under Native-Quality Constraints}
\label{sec:quality-constrained-results}
\label{sec:design-interpretation}

\textbf{Design gains and native-quality trade-off.} On independent testing, HD+\method{} lowers IPD by 27.0\% (95\% bootstrap interval: 25.2--28.6\%) and PIAD by 32.2\% (30.8--33.6\%), while Exact-8 acceptance increases from 94.5\% to 96.5\% (Table~\ref{tab:portable-full}). Public, Private, and Service move toward RPLAN's functional profile, although Outdoor moves away (Appendix~\ref{app:matched-design-results}). The overall alignment gain thus extends across several functional domains, alongside stronger Public anchoring: median PIA rises from 0.2800 to 0.3974, toward RPLAN's 0.6172.

These gains accompany higher Fr\'echet Inception Distance (FID) and Compatibility scores, indicating increased rendering-distribution mismatch and disagreement with requested access relations (Table~\ref{tab:native-evaluation}). These native-quality scores average five sampling repeats on a separate fixed cohort of 1,000 conditions (Appendix~\ref{app:native-quality}). Variability in archived five-repeat evaluations of unchanged Base HD sets FID and Compatibility limits of 9.5 and 2.8; both test means satisfy them. With validation quality constrained closer to Base, iteration 10 retains 89.8\% of the PIAD reduction on common test conditions (Appendix~\ref{app:stricter-quality}). On separate validation data, the lowest-IPD checkpoint meeting these limits and the acceptance and measurement-count requirements is \areaguard{}-4 (the Section~\ref{sec:sspt} area-expansion penalty with weight \(\lambda=4\)) at iteration 20; IPD alone decided, with no ties (Appendices~\ref{app:selection} and~\ref{app:selection-results}).

\setlength{\intextsep}{\ssptSavedIntextsep}

\setlength{\ssptSavedIntextsep}{\intextsep}
\setlength{\intextsep}{0pt}
\begin{wraptable}[9]{r}{0.48\textwidth}
\centering
\setlength{\abovecaptionskip}{0pt}
\caption{\textbf{Selected HD checkpoint: final test.} Native metrics are five-repeat mean $\pm$ std.}
\label{tab:native-evaluation}
\footnotesize
\setlength{\tabcolsep}{2pt}
\begin{tabular}{@{}lccc@{}}
\toprule
Metric & Base HD & +\method{} & Limit \\
\midrule
\ipd{} $\downarrow$ & 0.1111 & \textbf{0.0811} & -- \\
\piad{} $\downarrow$ & 0.4129 & \textbf{0.2798} & -- \\
FID $\downarrow$ & $7.798{\pm}0.140$ & $9.087{\pm}0.098$ & $9.500$ \\
Compatibility $\downarrow$ & $2.537{\pm}0.026$ & $2.572{\pm}0.049$ & $2.800$ \\
\bottomrule
\end{tabular}
\par\vspace{14pt}
\end{wraptable}
\textbf{Gains under identical design requirements.} For each test condition, we compare floorplans generated by HD before and after SSPT with the same room counts, types, and connection requirements. Both outputs meet the acceptance rules for 4,588 of the original 5,000 conditions. On this common condition set, IPD decreases from 0.1112 to 0.0812; PIA increases for 70.42\% of conditions, with a median within-condition change of +0.1164. Resampling each condition's before/after outputs together within the test blocks estimates sampling uncertainty in this median change, giving a 95\% confidence interval of [0.1082, 0.1257]. PIA gains persist when this comparison is restricted to conditions whose output graphs are connected both before and after SSPT, indicating improvement beyond connectivity repair (Figure~\ref{fig:selected-cases-main}; Appendix~\ref{app:matched-design-results}).

\setlength{\intextsep}{\ssptSavedIntextsep}

\textbf{Robustness to centrality choice.} We rescore the full accepted HD outputs and residential reference using four accessibility measures (Section~\ref{sec:ssio}): Legacy HH within connected components; bounded HH with scores in \([0,1]\); bounded HH excluding one- and two-node components from aggregation; and global harmonic centrality over the whole graph, with unreachable nodes contributing zero. IPD and PIAD decrease under every setting (Appendix~\ref{app:centrality-sensitivity}).

\begingroup
\setlength{\intextsep}{4pt}
\setlength{\abovecaptionskip}{4pt}
\setlength{\belowcaptionskip}{0pt}
\begin{figure}[!htbp]
  \centering
  \includegraphics[width=\textwidth]{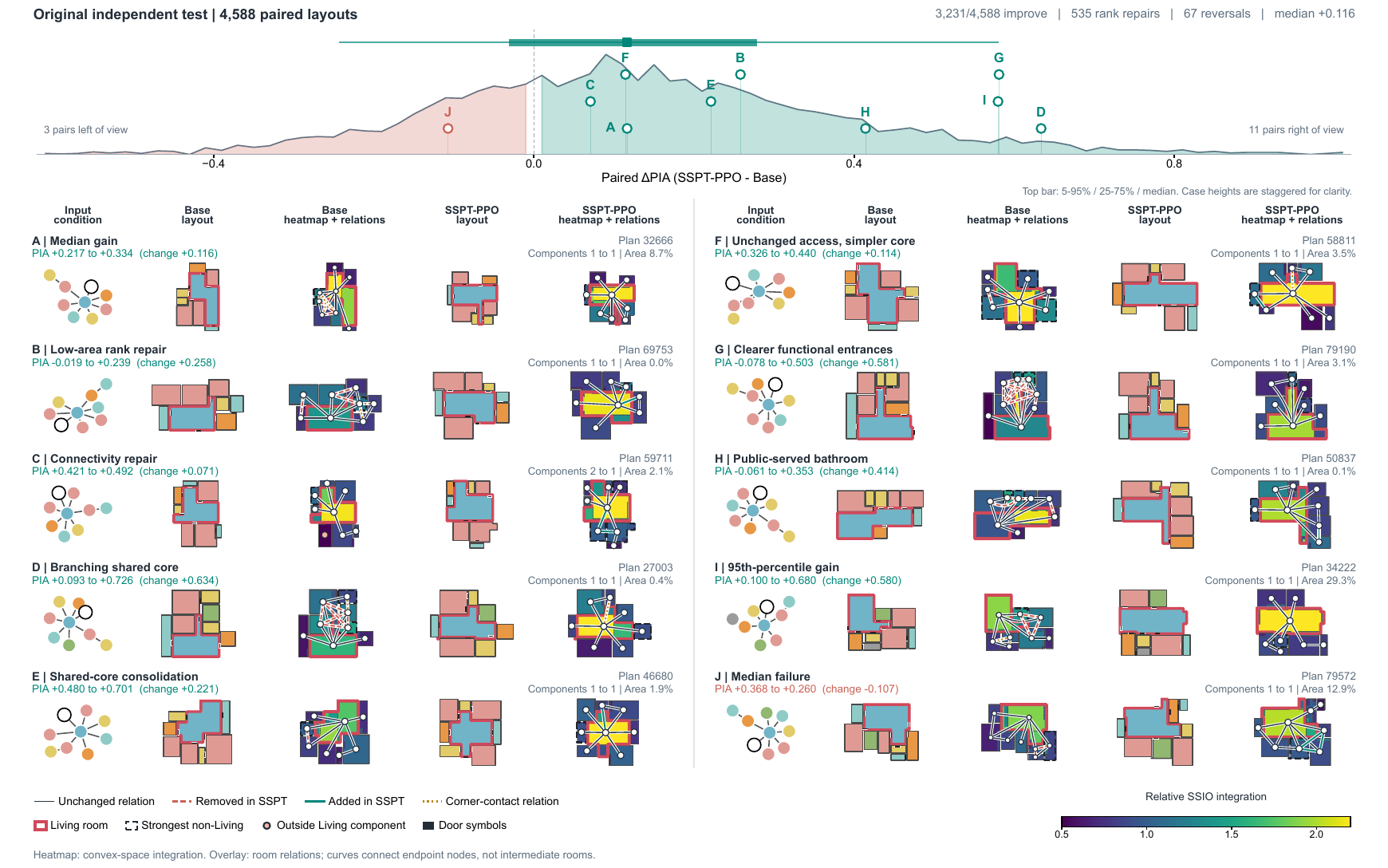}
  \caption{\textbf{HD+\method{} reorganizes functional configuration under fixed programs.} Top: paired \pia{} changes with ten cases marked. Bottom: shared inputs, layouts, integration heatmaps, and room-access relations; the legend identifies retained, removed, and added links.}
  \label{fig:selected-cases-main}
\end{figure}
\endgroup

\Needspace{5\baselineskip}
\textbf{Case studies of spatial organization.} To explain the spatial changes behind the gains, we analyze 200 pairs stratified by test block and PIA change. Most changes concern room-access grouping or the living-space core and branches. Of 144 PIA-increasing pairs, 131 retain the same number of recovered room-level links or fewer: gains predominantly accompany spatial reorganization rather than denser connectivity. Figure~\ref{fig:selected-cases-main} illustrates ten pairs; Appendix~\ref{app:systematic-case-analysis} reports the full analysis.
Among the illustrated cases, B replaces bathroom dominance with Living anchoring in connected layouts with the same eight-room program and total area; C repairs disconnection and improves anchoring. A, B, and I improve with fewer projected cross-room links, whereas J's integration advantage weakens despite an added link.

\Needspace{10\baselineskip}
\subsection{Ablation Studies}
\label{sec:ablations}

HD development ablations test update location, reward, and expansion control (Appendix~\ref{app:ablation-development}).

\setlength{\ssptSavedIntextsep}{\intextsep}
\setlength{\intextsep}{0pt}
\begin{wraptable}{r}{0.48\textwidth}
\centering
\setlength{\abovecaptionskip}{0pt}
\caption{\textbf{HouseDiffusion reverse-time ablation.}}
\label{tab:region-dev}
\footnotesize
\setlength{\tabcolsep}{3.5pt}
\begin{tabular}{@{}lcc@{}}
\toprule
Reverse-time support & \ipd{} $\downarrow$ & \piad{} $\downarrow$ \\
\midrule
Lower-noise segment & \textbf{0.0945} & \textbf{0.3501} \\
Uniform full horizon & 0.1026 & 0.3873 \\
Higher-noise segment & 0.1126 & 0.4197 \\
\bottomrule
\end{tabular}
\end{wraptable}
\textbf{(1) Where should post-training act?} At iteration 3, lower-noise updates outperform uniform full-horizon and higher-noise updates with profile-distance reward and retained-step budget fixed (Table~\ref{tab:region-dev}). Among these schedules, feedback is most effective late in HD's denoising process.
\par\WFclear
\setlength{\intextsep}{\ssptSavedIntextsep}

\textbf{(2) Which design reward should guide learning?} We compare three targets: Living integration lead rewards the difference between the highest Living and non-Living room integration scores; RPLAN profile alignment matches each layout's relative category-integration profile to RPLAN; and Standardized Living contrast measures Living's category score relative to the non-Living category median, normalized by a robust scale. The first two perform similarly and outperform the third. Their sum, Living lead + RPLAN alignment, yields the best design scores but exceeds the FID limit (Table~\ref{tab:reward-dev}), motivating the following area-control ablation.

\textbf{(3) Does controlling expansion preserve design gains?} With Living integration lead fixed, we compare weights \(\lambda=0,2,4\) for the AreaGuard penalty defined in Section~\ref{sec:sspt}, evaluating all variants at iteration 10. Relative to no penalty, weight 2 retains design gains while sharply reducing FID and Compatibility (Table~\ref{tab:guard-dev}, Figure~\ref{fig:results-overview}(c)); weight 4 further improves native quality at some design cost. AreaGuard curbs a reward shortcut: excessive expansion can increase the Living integration lead reward while degrading geometric quality. The AreaGuard weight and training duration are then selected jointly.

\begingroup
\setlength{\intextsep}{4pt}
\setlength{\abovecaptionskip}{2pt}
\par\noindent\begin{minipage}{\linewidth}
\centering\footnotesize
\captionof{table}{\textbf{\areaguard{} improves the quality--design trade-off.} Iteration-10 development.}
\label{tab:guard-dev}
\begin{tabular}{lcccc}
\toprule
Variant & \ipd{} $\downarrow$ & \piad{} $\downarrow$ &
FID $\downarrow$ & Compatibility $\downarrow$ \\
\midrule
Living integration lead & 0.0830 & \textbf{0.2855} & \(19.55\pm0.18\) & \(2.583\pm0.041\) \\
+\areaguard{}-2 & \textbf{0.0827} & 0.2890 &
\(8.91\pm0.12\) & \(2.118\pm0.040\) \\
+\areaguard{}-4 & 0.0858 & 0.2976 &
\textbf{8.59}\(\pm\)\textbf{0.11} &
\textbf{2.062}\(\pm\)\textbf{0.063} \\
\bottomrule
\end{tabular}
\end{minipage}\par

\FloatBarrier
\endgroup
\subsection{Residential-Use Assessment}
\label{sec:human-residential-use}

To test whether integration gains correspond to perceived residential-use benefits, a real-estate designer and an architecture student outside the project independently compared 50 condition-matched HD Base/SSPT pairs with identities and scores hidden, judging everyday-use suitability. Both favored SSPT and higher-PIA layouts in most definite choices (Table~\ref{tab:human-main}). The sample spans test blocks and PIA changes; architecture researchers conducted a separate author-involved discussion (Appendix~\ref{app:human-residential-use}).
Preference for SSPT also persists among the 22 pairs whose native Compatibility increases (Appendix~\ref{app:compatibility-use}).

\begingroup
\setlength{\abovecaptionskip}{2pt}
\par\noindent\begin{minipage}{\linewidth}
\setlength{\parskip}{0pt}
\centering\footnotesize
\captionof{table}{\textbf{Residential-use judgments.} Definite preferences among 50 pairs. Last column: Base has higher PIA among Base-preferred pairs. Complete counts: Appendix~\ref{app:human-native-use}.}
\label{tab:human-main}
\setlength{\tabcolsep}{3pt}
\begin{tabular}{@{}lccc@{}}
\toprule
Assessment & Prefer SSPT & Prefer higher PIA & Base higher PIA when preferred\\
\midrule
Real-estate designer & 83\% (34/41) & 73\% (30/41) & 43\% (3/7)\\
Architecture student & 67\% (28/42) & 67\% (28/42) & 43\% (6/14)\\
Architecture researchers & 74\% (29/39) & 72\% (28/39) & 40\% (4/10)\\
\bottomrule
\end{tabular}
\end{minipage}\par
\endgroup

\Needspace{6\baselineskip}
\subsection{Cross-Backbone Portability}
\label{sec:portability-results}

We test whether \sspt{} improves spatial organization across diffusion generators with different output representations. Each of the five backbones is post-trained separately, applying the shared design objective through its native sampling policy (Section~\ref{sec:sspt}). For each backbone, Base and post-trained models are evaluated on the same 5,000 conditions with \ssio{} measurement settings held fixed. \ipd{} and \piad{} decrease across all five backbones (Table~\ref{tab:portable-diffusion}, Figure~\ref{fig:results-overview}(b)).
\looseness=-1
\setlength{\intextsep}{\ssptSavedIntextsep}

\setlength{\ssptSavedIntextsep}{\intextsep}
\setlength{\intextsep}{0pt}
\begin{wraptable}{r}{0.55\textwidth}
\centering
\setlength{\abovecaptionskip}{0pt}
\caption{\textbf{SSPT-PPO across five diffusion backbones.}}
\label{tab:portable-diffusion}
\footnotesize
\setlength{\tabcolsep}{3pt}
\begin{tabular}{@{}lc>{\columncolor{ssptcolumnbg}}cc>{\columncolor{ssptcolumnbg}}c@{}}
\toprule
& \multicolumn{2}{c}{\ipd{} $\downarrow$} & \multicolumn{2}{c}{\piad{} $\downarrow$} \\
\cmidrule(lr){2-3}\cmidrule(l){4-5}
Backbone & Base & +\method{} & Base & +\method{} \\
\midrule
HD & 0.1111 & \textbf{0.0811} & 0.4129 & \textbf{0.2798} \\
CE2EPlan & 0.1914 & \textbf{0.1352} & 0.5816 & \textbf{0.4460} \\
DiffPlanner & 0.1803 & \textbf{0.1555} & 0.5548 & \textbf{0.5043} \\
ChatHD & 0.0866 & \textbf{0.0838} & 0.3517 & \textbf{0.3446} \\
GSDiff-SSIO & 0.0386 & \textbf{0.0360} & 0.0903 & \textbf{0.0769} \\
\bottomrule
\end{tabular}
\par\vspace{6pt}
\end{wraptable}
Exact-8 acceptance rises for HD and DiffPlanner; CE2EPlan improves both design metrics while accepting fewer outputs, whereas ChatHD changes little in either metrics or acceptance (Appendix~\ref{app:portability-results}). GSDiff's boundary outputs leave access relations unspecified and require completion for \ssio{} measurement. Section~\ref{sec:gsdiff-case} examines the resulting GSDiff-SSIO system separately.
\looseness=-1

\setlength{\intextsep}{\ssptSavedIntextsep}

\subsection{Completing Design Relations: The GSDiff Case}
\label{sec:gsdiff-case}

GSDiff~\citep{hu2025gsdiff} exemplifies boundary-based generation without explicit access relations. Unlike outputs with door information, its room boundaries specify adjacency but not direct access. Evaluating spatial organization for use therefore requires completing these missing relations. We train an access predictor on RPLAN-derived local access labels, using room geometry, types, and local context. The resulting GSDiff-SSIO system uses the same frozen predictor for Base and +\method{}, with \ssio{} measuring integration on the completed graph. Updating the generator through \sspt{} improves both design metrics under this fixed completion map (Table~\ref{tab:portable-diffusion}).
\looseness=-1

We examine both completion fidelity and its influence on design scores. On 500 held-out RPLAN plans, redrawing predicted doors before decomposition gives access-pair F1 0.8879 and median absolute PIA error 0.3611; redrawing annotated doors still gives 0.3692. The deployed pipeline fixes the decomposition and inserts access edges, giving error 0.0724 on the same plans (Appendix~\ref{app:access-route-control}). Separately, on 538 conditions accepted by both models under all four completion routes, withholding room-type inputs from the frozen predictor lowers Base median \pia{} from 0.5882 to 0.2781 with generated geometry fixed. GSDiff-SSIO scores thus depend on both generated geometry and type-informed access completion (Appendix~\ref{app:gsdiff-completion}). For such generators, translating design theory into learning requires both relation recovery and measurement validation.
\looseness=-1

\section{Cross-Domain Discussion and Implications}
\label{sec:design-learning-principles}

We situate our floorplan approach within a review of 55 studies across 14 design categories (Appendix~\ref{app:literature-review}). Three comparative maps distinguish what feedback measures, where it acts, and its reuse across generators (Figures~\ref{fig:design-survey-scatter} and~\ref{fig:design-survey-posttraining-scope}). Theory-defined analysis rarely guides post-training across heterogeneous formats in these studies. The floorplan study suggests three considerations for exploring this direction elsewhere: a \emph{design container} specifying what to represent, an \emph{empirical reference system} grounding evaluation, and a \emph{design-learning interface} connecting feedback to generator updates. Appendix~\ref{app:design-learning-principles} proposes a theory-translation workflow and provides an agent-assisted harness, using worked user-interface (UI), scientific-diagram, and poster examples to discuss choices in representation, measurement, and feedback.
\looseness=-1

\section{Conclusion and Future Work}

We make spatial organization explicit across heterogeneous formats through a shared IR for space-syntax analysis. Across five diffusion backbones, empirically grounded post-training turns architectural knowledge into stronger design capabilities while preserving native mechanisms. Directly understanding designers' and users' intentions and needs, and iterating through simulated user feedback, remain beyond this theory-grounded scope. Future work should embed design-native representations and objectives in pretraining and investigate how design capability scales with data, model capacity, and compute. It should also validate the three-element workflow and harness through generator training and independent task evaluation across design domains.
\looseness=-1

\label{end:main-content}

\section*{Reproducibility Statement}

The main paper specifies the frozen measurement and evaluation protocol, terminal feedback, and shared native-policy PPO objective. The appendix documents \ssio{} construction, data and metric support, five backbone policy bindings, independent checkpoint selection and final evaluation, and GSDiff-SSIO access completion. Five-block, condition-matched, connectivity, centrality, functional-profile, and reward-development analyses support the reported results.

\section*{AI use statement}

The authors established the narrative, planned the exposition, and prepared an initial draft. AI tools then assisted with iterative checks of logical coherence and language polishing. Within the established code framework, they assisted with implementation details following the authors' algorithm and experiment design. They helped review the correctness and completeness of mathematical reasoning and notation and suggested additions and refinements. Building on an author-curated literature collection, they checked related-work coverage and identified recent relevant studies to supplement it.

Generative AI tools also supported descriptive thematic coding and interpretation of the 200-pair case analysis. The human judgments were collected separately.

\bibliography{references}
\bibliographystyle{iclr2027_conference}

\clearpage
\appendix
\numberwithin{equation}{section}
\section{Measurement Definitions}
\label{app:proofs}

\subsection{Quick Reference}
\label{app:abbreviations}
\begingroup
\normalsize
\setlength{\tabcolsep}{5pt}
\setlength{\LTpre}{8pt}
\setlength{\LTpost}{8pt}
\setlength{\extrarowheight}{1pt}
\renewcommand{\arraystretch}{1.13}
\begin{longtable}{@{}>{\raggedright\arraybackslash}p{0.21\textwidth}>{\raggedright\arraybackslash}p{\dimexpr0.79\textwidth-10pt\relax}@{}}
\caption{Quick reference: core terms, evaluation measures, and models.}\label{tab:abbreviations}\\
\toprule Term & Meaning in this paper \\ \midrule
\endfirsthead
\multicolumn{2}{l}{Table \thetable\ continued.}\\
\toprule Term & Meaning in this paper \\ \midrule
\endhead
\midrule\multicolumn{2}{r}{Continued on next page.}\\\endfoot
\bottomrule\endlastfoot
SSIO & Space Syntax Integration Operator: recovers access graphs and measures spatial organization across output formats. \\
RPLAN-SSI & RPLAN-derived Space Syntax Integration benchmark: evaluates generated layouts against room-count-matched real layouts using IPD and PIAD. \\
SSPT & Space Syntax-Guided Post-Training: updates pretrained generators using SSIO feedback. \\
IR & Intermediate representation: shared, theory-defined design \mbox{objects and relations}. \\
CSAG & Convex-space access graph: convex spaces as nodes and access \mbox{relations as edges}. \\
HH & Hillier--Hanson integration: the component-local Legacy HH score used for benchmark evaluation and post-training. \\
IPD & Integration Profile Distance: population-level distance from RPLAN's four-domain functional profile; lower is better. \\
PIA & Public-Anchor Integration Advantage: normalized within-plan lead of Living's category mean over the strongest eligible non-Living category; positive when Living leads, read against RPLAN's distribution. \\
PIAD & PIA Distance: Wasserstein-1 distance between a cohort's and RPLAN's PIA distributions; lower is better. \\
Exact-8 acceptance & Share of all generation attempts that are measurable and retain exactly eight spatial instances, including Living. \\
FID & Fr\'echet Inception Distance: discrepancy between generated and real rendering distributions; lower is better. \\
Compatibility & Mean number of missing or extra access edges relative to the input requirements; lower is better. \\
Base & Generator before SSPT, evaluated under the stated sampling and measurement protocol. \\
HD & HouseDiffusion: polygon-based diffusion; used for the main \mbox{experiments and ablations}. \\
ChatHD & ChatHouseDiffusion: semantic-raster diffusion; SSPT uses a frozen prefix and trainable suffix. \\
CE2EPlan & Room-tensor diffusion; SSPT updates room centers and sizes. \\
DiffPlanner & Room-box diffusion; SSPT updates its PartitioningDiff component. \\
GSDiff & Structural-boundary diffusion without explicit access relations. \\
GSDiff-SSIO & GSDiff with access completion before SSIO measurement; the completion model is frozen for both Base and post-training. \\
\end{longtable}

\begin{longtable}{@{}>{\raggedright\arraybackslash}p{0.28\textwidth}>{\raggedright\arraybackslash}p{\dimexpr0.72\textwidth-10pt\relax}@{}}
\caption{Quick reference: training terms and general abbreviations.}\label{tab:training-terms}\\
\toprule Term & Meaning in this paper \\ \midrule
\endfirsthead
\multicolumn{2}{l}{Table \thetable\ continued.}\\
\toprule Term & Meaning in this paper \\ \midrule
\endhead
\midrule\multicolumn{2}{r}{Continued on next page.}\\\endfoot
\bottomrule\endlastfoot
RL & Reinforcement learning. \\
RLVR & Reinforcement learning with verifiable rewards. \\
PPO & Proximal policy optimization: the shared policy-update \mbox{objective in SSPT}. \\
SSPT-PPO & Main SSPT method: PPO updates using sampled \mbox{diffusion trajectories}. \\
SSPT-Iter & Pilot selection-and-retraining variant using SSIO-ranked real and generated examples. \\
Living integration lead & Reward: maximum Living room score minus maximum non-Living room score; distinct from category-level PIA. \\
RPLAN profile alignment & Reward: negative mean absolute distance from a layout's relative category scores to RPLAN category medians. \\
Standardized \mbox{Living contrast} & Reward: Living's category score centered on the non-Living category median and divided by a robust scale. \\
Living lead + \mbox{RPLAN alignment} & Sum of the two named rewards with their established scaling, rather than a sum of PIA and IPD. \\
AreaGuard & One-sided penalty on total retained-atom area above a frozen Base-derived threshold; scale fixed separately for each backbone. \\
AreaGuard-2 / -4 & AreaGuard configurations with penalty weights \(\lambda=2\) and \(\lambda=4\). \\
MAD & Median absolute deviation: robust scale used in \mbox{Standardized Living contrast}. \\
GenAI & Generative artificial intelligence. \\
GAN & Generative adversarial network. \\
DDIM & Denoising diffusion implicit model: sampler used for ChatHD's frozen prefix. \\
IQR & Interquartile range: spread of total retained-atom area across Base outputs, used to set the AreaGuard scale. \\
\end{longtable}
\endgroup

\begingroup
\small
\setlength{\tabcolsep}{4pt}
\setlength{\LTpre}{8pt}
\setlength{\LTpost}{8pt}
\renewcommand{\arraystretch}{1.12}
\begin{longtable}{@{}>{\raggedright\arraybackslash}p{0.24\textwidth}>{\raggedright\arraybackslash}p{\dimexpr0.76\textwidth-8pt\relax}@{}}
\caption{Notation used in the paper.}\label{tab:notation}\\
\toprule Symbol & Meaning \\ \midrule
\endfirsthead
\multicolumn{2}{l}{Table \thetable\ continued.}\\
\toprule Symbol & Meaning \\ \midrule
\endhead
\midrule\multicolumn{2}{r}{Continued on next page.}\\\endfoot
\bottomrule\endlastfoot
\multicolumn{2}{l}{\textbf{Design recovery and measurement}} \\
\(c=(B,\mathcal M_c,G_c)\) & boundary, requested room multiset, optional relation graph \\
\(\mathcal A_m\) & deterministic native-output normalization internal to \ssio{} \\
\(\mathcal C_\phi\) & frozen access-completion map when native access is absent \\
\(z_{p,m}\) & canonical measurement record for plan \(p\) and method \(m\) \\
\(G_{p,m}=(V_{p,m},E_{p,m})\) & room-labeled rectangular convex-atom access graph \\
\(\mathcal R^+_{p,m}; V_j\) & retained rooms; retained atoms of room \(j\) \\
\(S_v; I_v,H_v\) & node centrality; Legacy HH integration and global harmonic centrality \\
\(\bar S_j,\bar S_{p,k},S_{p,g}\) & room, category, and domain scores \\
\(\eta^{\mathrm{eval}},\eta_m^{\mathrm{train}}\) & common evaluation settings and backbone-specific \mbox{training-time measurement settings} \\
\addlinespace[3pt]
\multicolumn{2}{l}{\textbf{Functional organization and benchmark evaluation}} \\
\(\mathcal G,\mathcal H_p\) & four functional domains and those with score support in plan \(p\) \\
\(\mathcal K_p,\mathcal K_{p,g}\) & categories with retained score support in plan \(p\), and those within domain~\(g\) \\
\(r^{\mathrm{dom}}_{p,g}(S)\) & relative domain score within plan \(p\) \\
\(\widetilde{\mathbf r}(\mathcal C;S)\) & functional profile: four-domain median vector of cohort \(\mathcal C\) \\
\(\Delta_{\mathrm{pub}}(p)\) & Public-domain margin within plan \(p\) \\
\(\mathrm{IPD},\mathrm{PIAD}\) & population-profile and anchor-distribution benchmark metrics \\
\(\mathrm{PIA}(p);\,F_{\mathcal C}\) & plan-level anchor advantage; its distribution over cohort \(\mathcal C\) \\
\addlinespace[3pt]
\multicolumn{2}{l}{\textbf{Feedback and native-policy learning}} \\
\(R_{\mathrm{anchor}},R_{\mathrm{AreaGuard}}\) & anchor reward and \areaguard{} penalty \\
\(\lambda,q_A,s_A\) & \areaguard{} weight, frozen Base-derived area threshold, and area scale \\
\(\widehat R_i,\widehat A_i\) & quantile-clipped terminal reward and detached standardized trajectory advantage \\
\((m,r,i,d)\) & backbone, outer iteration, rollout, and execution-order decision indices \\
\((t_d^{(m)},t_{d+1}^{(m)})\) & native reverse-time indices before and after stochastic decision \(d\) \\
\(s_{i,d}^{(m)},a_{i,d}^{(m)}\) & native RL state and joint reverse action at decision \(d\) \\
\(\Omega_{i,d}^{(m)}\) & active stochastic coordinates in the joint policy density \\
\(D_m,\mathcal D_i^{(m)}\) & number and ordered set \(\{1,\ldots,D_m\}\) of retained stochastic decisions \\
\(\tau_i^{(m)},\mathcal P_m\) & replayable trajectory and its outer-iteration rollout pool \\
\(p_{\mathrm{old},m}\) & iteration-start policy snapshot used to collect \(\mathcal P_m\) \\
\(\mathcal B_m\) & rollout--decision pairs included in PPO averaging \\
\(N_m,M_m\) & rollouts per outer iteration and backbone-specific log-ratio bound \\
\(L_m\) & padded room-slot capacity for room-tensor backbones \\
\(\Theta_E\) & eligible post-training checkpoints \\
\end{longtable}
\endgroup

\subsection{Canonical SSIO Measurement}
\label{app:ssio}

\ssio{} converts native formats into access graphs and aggregates graph scores into design measurements. Appendices~\ref{app:grounding-robustness} and~\ref{app:centrality-sensitivity} test these measurement choices.

For plan \(p\) with condition \(c_p\), source-specific normalization \(\mathcal A_m\) produces a common record of room regions, labels, and interior access:
\begin{equation}
z_{p,m}=\mathcal A_m(x^{\mathrm{nat}}_{p,m};c_p)=\big((Q_{p,m,j},y_{p,m,j})_{j=1}^{R_{p,m}},
D^{\mathrm{int}}_{p,m}\big),
\label{eq:ssio-canonical-record}
\end{equation}
where \(Q_{p,m,j}\) is a walkable room region, \(y_{p,m,j}\) its semantic label, and \(D^{\mathrm{int}}_{p,m}\) the interior access. For GSDiff-SSIO, the frozen completion map \(\mathcal C_\phi\) supplies missing access relations before measurement.

\begin{table}[!htbp]
\centering
\renewcommand{\arraystretch}{1.12}
\caption{Semantic crosswalk for the RPLAN benchmark.}
\label{tab:labels}
\small
\setlength{\tabcolsep}{4pt}
\begin{tabular}{@{}>{\raggedright\arraybackslash}p{0.23\textwidth}>{\raggedright\arraybackslash}p{0.34\textwidth}>{\raggedright\arraybackslash}p{0.38\textwidth}@{}}
\toprule
Canonical category & Source labels & Benchmark role \\
\midrule
Living & Living & Public \\
Dining & Dining & Public \\
Bedroom & Master, Child, Second, Guest & Private \\
Study & Study & Private \\
Kitchen & Kitchen & Service \\
Bathroom & Bathroom & Service \\
Balcony & Balcony & Outdoor \\
Entrance & Entrance & \pia{} normalization only \\
Storage & Storage & \pia{} competitor \\
Wall-in/Unknown & Wall-in & graph/count only \\
\bottomrule
\end{tabular}
\end{table}

All final evaluations use the same settings \(\eta^{\mathrm{eval}}\): a \(256\times256\) lattice, a minimum retained-atom area of 50 pixels\(^2\), and a within-room continuity threshold of 1.5 pixels. Vector coordinates are mapped and clipped to the lattice before rasterization. Repeated generation attempts are counted separately even when their geometry is identical.

For each room, \ssio{} repeatedly extracts the largest axis-aligned rectangle in its walkable mask; each rectangle becomes a convex-space atom. Rooms and rectangles are visited in fixed source and raster orders, with ties resolved by the first candidate visited. Atoms within a room are connected if they touch or lie within the continuity threshold. Any remaining disconnected components within that room are joined through their closest atom pair. An interior door connects the closest atoms of its two incident rooms. Each atom retains its room identity, including when atom bounds coincide.

Table~\ref{tab:labels} maps RPLAN labels to measurement categories. Bedroom subtypes merge after room-level integration. Entrance, Storage, and unassigned structural regions can affect graph distances but do not enter the functional profile; Section~\ref{app:metric-support} specifies their roles in other statistics.

The fixed traversal order, tie rules, label mapping, and finite-score conventions make each defined measurement deterministic. Graph isomorphisms that preserve room membership and semantic labels also preserve shortest-path distances and all defined room, category, domain, \ipd{}, and \pia{} scores. Thus the scores depend on the recovered graph; door evidence and access completion determine its edges.

\paragraph{Door evidence in RPLAN.}\label{app:rplan-door-audit} We audited how all 80,788 RPLAN plans encode passages between rooms. Distinct room instances never touch directly: wall or door pixels separate all 646,859 adjacent instance pairs. Every annotated passage between rooms is therefore an interior-door region, which \ssio{} reads as access; rooms that only share a wall remain unconnected. Separate Living and Dining rooms follow the same encoding. A Dining room is labeled in 1,311 plans (1.6\%); of the 1,312 Living--Dining pairs in these plans, 1,089 are joined by an annotated opening, 185 are separated by a wall, and 38 are not adjacent. Door-based recovery can thus miss access only where an opening is unannotated. Such an omission is evident when a room has no opening at all, as for 68 Dining rooms (0.08\% of plans); 15 lie in the 20,348-plan reference and 3 in the 5,000 benchmark conditions.

\begin{figure}[!htbp]
\centering
\includegraphics[width=\linewidth]{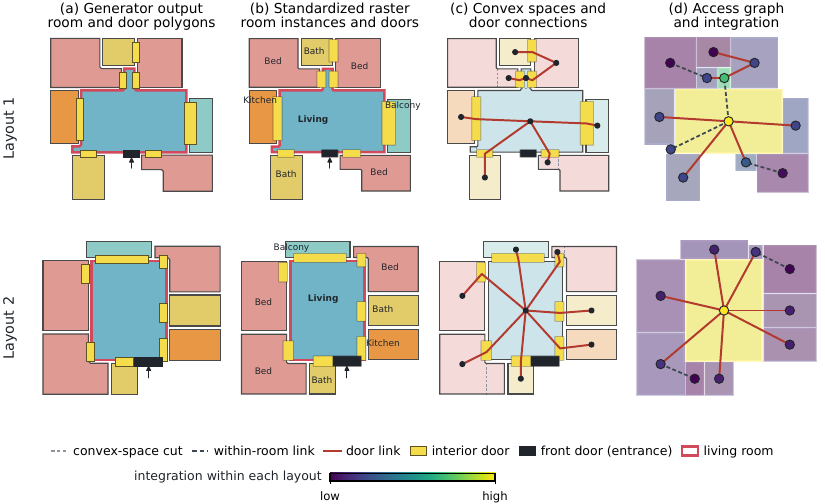}
\caption{\textbf{SSIO's graph construction, illustrated on two generated layouts.} Each row is one layout: (a) room and door polygons; (b) standardized raster; (c) convex spaces of each room (maximal rectangles), with red lines through each door to the spaces it connects; (d) access graph of convex spaces with within-room and door links, colored by integration within each layout. Exact rules are given in the text.}
\label{fig:ssio-worked}
\end{figure}

Figure~\ref{fig:ssio-worked} separates the two sources of graph edges: continuity within a room and openings between rooms. Room membership remains attached to each convex atom, so integration can be aggregated back to rooms and categories. Formats without explicit access information require the additional recovery step examined in Appendix~\ref{app:gsdiff-completion}.

\FloatBarrier
\subsection{Node Scores and Aggregation}

Node centrality measures accessibility through the access graph. For fixed \(p,m\), write \(G=(V,E)\) and omit these indices from node and room scores. Every node included in scoring receives a finite, nonnegative value. Let \(C(v)\) be node \(v\)'s connected component, \(n=|C(v)|\), and \(d(v,u)\) the unweighted shortest-path distance. For \(n>2\), mean depth and relative asymmetry define Legacy HH integration within that component; the zero branch handles the remaining cases:
\begin{equation}
\begin{gathered}
\mathrm{MD}_v=\frac{1}{n-1}\sum_{u\in C(v)\setminus\{v\}}d(v,u),
\qquad \mathrm{RA}_v=\frac{2(\mathrm{MD}_v-1)}{n-2},\\[4pt]
D_n=\frac{2\{n[\log_2((n+2)/3)-1]+1\}}{(n-1)(n-2)},
\qquad I_v=\begin{cases}
D_n/\mathrm{RA}_v,&n>2\ \text{and}\ \mathrm{RA}_v>0,\\
0,&\text{otherwise}.
\end{cases}
\end{gathered}
\label{eq:primary-hh}
\end{equation}
The zero branch covers one- and two-node components and universal nodes connected directly to all others in their component. Averages include only retained, scored atoms and rooms.

Cross-corpus grounding uses global harmonic centrality~\citep{boldi2014axioms}, measuring whole-graph accessibility with normalization by \(|V|-1\):
\begin{equation}
H_v=\frac{1}{|V|-1}
\sum_{\substack{u\in V\setminus\{v\}\\d(v,u)<\infty}}\frac{1}{d(v,u)}
\quad\text{for }|V|>1,
\label{eq:global-harmonic}
\end{equation}
with \(H_v=0\) for \(|V|\leq1\).  Unreachable nodes thus contribute zero, whereas universal nodes retain maximal finite centrality.

Room score \(\bar S_j\) averages node scores \(S_v\) over room \(j\)'s retained atoms. Category score \(\bar S_{p,k}\) averages retained room scores in category \(k\). Domain score \(S_{p,g}\) averages category scores in domain \(g\), weighted by their numbers of scored rooms. Categories and domains contribute only when they contain a scored room. Equation~\eqref{eq:functional-profile} defines within-plan relative domain scores and their cohort medians, forming the four-domain functional profile; Section~\ref{sec:rplanssi} defines the Public-domain margin, \ipd{}, and \pia{}.

\par\addvspace{8pt}\noindent\begin{minipage}{\linewidth}
\centering
\renewcommand{\arraystretch}{1.12}
\captionof{table}{Distinct summaries and learning signals derived from SSIO scores.}
\label{tab:measurement-levels}
\small
\setlength{\tabcolsep}{4pt}
\newlength{\ssptQuantityWidth}
\settowidth{\ssptQuantityWidth}{Public-domain margin \(\Delta_{\mathrm{pub}}\)}
\begin{tabular}{@{}l>{\raggedright\arraybackslash}p{0.40\textwidth}>{\raggedright\arraybackslash}p{\dimexpr0.60\textwidth-\ssptQuantityWidth-16pt\relax}@{}}
\toprule
Quantity & Aggregation or contrast & Role \\
\midrule
Functional profile \(\widetilde{\mathbf r}\) & Within-plan relative domain scores, then \mbox{cohort medians} & Population \mbox{organization; \ipd{}} \\
Public-domain margin \(\Delta_{\mathrm{pub}}\) & Public minus strongest competing \mbox{relative domain score} & Individual \mbox{domain hierarchy} \\
\pia{} & Living category mean minus strongest eligible category mean; category-normalized & Individual \mbox{RPLAN anchor} \\
\piad{} & Per-plan \pia{} values, then Wasserstein-1 distance to the reference distribution & Population \mbox{anchor distribution} \\
\(R_{\mathrm{anchor}}\) & Largest Living room score minus largest non-Living room score; unnormalized & Living integration \mbox{lead reward} \\
\bottomrule
\end{tabular}
\end{minipage}\par\addvspace{8pt}

To test sensitivity to the centrality definition, we rescore the same graphs. For \(n>2\), bounded HH uses \(1/(1+\mathrm{RA}_v/D_n)\), including one when \(\mathrm{RA}_v=0\); a connected pair scores one and an isolated node zero. A second variant excludes atoms in components of size at most two from aggregation. Room means, category weights, and the set of scored domains are recomputed using the remaining atoms and rooms. The original Exact-8 acceptance counts stay fixed, and a statistic is undefined if no scored elements remain. Appendix~\ref{app:centrality-sensitivity} reports the resulting task-matched gap and post-training gains.

Plans without retained atoms cannot be measured. Statistics for categories, functional domains, or \pia{} competitors without scored elements remain undefined, with no imputed values. Appendix~\ref{app:metrics} specifies acceptance and metric requirements.

\FloatBarrier
\section{Residential Reference and Benchmark}
\label{app:data}

\subsection{Data Roles}

\par\addvspace{8pt}\noindent\begin{minipage}{\linewidth}
\centering
\captionof{table}{Data used for analysis and evaluation. Development, validation, calibration, and matched endpoint pairs refer to the primary HD study; Table~\ref{tab:training-bindings} gives training scales for the other backbones.}
\label{tab:data-roles}
\small
\renewcommand{\arraystretch}{1.08}
\begin{tabular}{>{\raggedright\arraybackslash}p{0.29\textwidth}>{\raggedright\arraybackslash}p{0.18\textwidth}>{\raggedright\arraybackslash}p{0.43\textwidth}}
\toprule
Role & Size & Use \\
\midrule
RPLAN real-corpus prior & 80,788 & cross-corpus functional regularity \\
LIFULL Living/Dining cohort & 19,677 / 19,541 & relative-domain-score / margin support \\
MSD Swiss corpus & 13,472 & cross-corpus functional regularity \\
SFC-A68 multi-country corpus & 270 & cross-corpus functional regularity \\
RPLAN exact-eight reference & 20,348 & frozen four-domain benchmark profile \\
\method{} rollout conditions & \(\leq7\)-room requests & post-training only \\
Functional validation & 2,000 & checkpoint eligibility and ordering \\
\areaguard{} calibration & 1,000 & fixed area-scale statistics \\
Base quality calibration & \(3\times5\times1,000\) & common FID/Compatibility ceilings; \mbox{two protocols} \\
Native validation & \(1,000\times5\) & candidate FID and Compatibility \\
Native evaluation & \(1,000\times5\) & post-selection FID and Compatibility \\
Shared evaluation suite & \(5\times1,000\) & final benchmark and block stability \\
Matched endpoint pairs & 4,588 & condition-paired Base-to-\method{} analysis \\
\bottomrule
\end{tabular}
\end{minipage}\par\addvspace{8pt}

The five final-evaluation blocks have disjoint condition identifiers absent from the recorded post-training rollout and checkpoint-selection sets. After checkpoint selection, each method makes one attempt per condition, without replacing failures. The RPLAN reference describes the residential population rather than a real layout matched to each test condition; its exact-eight median \pia{} is 0.61718.

\subsection{Cross-Corpus Mapping}

We map each corpus to four functional domains: Living and Dining to Public; bedrooms and Study to Private; Kitchen and Bathroom to Service; and Balcony to Outdoor. Circulation and Storage affect access graphs but do not enter domain averages. In LIFULL, closet/laundry maps to Storage and corridor to Entrance/circulation. Of 145,811 released records, 19,691 annotate Living or Dining; polygon reconstruction retains 19,677 for functional-profile estimation, of which 19,541 also support \(\Delta_{\mathrm{pub}}\). The same mapping covers 13,472 Swiss floorplans from MSD and 270 apartments recovered from SFC-A68 access graphs across 13 countries.

Appendix~\ref{app:grounding-robustness} reports how centrality and connectivity choices affect the RPLAN and LIFULL Public-domain margins.

\subsection{Metric Support and Acceptance}
\label{app:metric-support}
\label{app:metrics}

Each functional-profile entry is computed from plans that contain the corresponding domain and have a finite relative domain score. In cohort \(\mathcal C\), these plans form \(\mathcal C_g=\{p\in\mathcal C:g\in\mathcal H_p\}\). The profile is defined only when all four domains are represented. Reporting \ipd{} requires at least 100 contributing plans per domain overall and 20 per domain in each of five blocks; checkpoint validation uses the overall minimum only.

For \pia{}, normalization categories are Living, Bedroom, Kitchen, Bathroom, Dining, Study, Balcony, Entrance, and Storage. Eligible non-Living competitors are Bedroom, Kitchen, Bathroom, Dining, Study, Balcony, and Storage. Entrance normalizes scores but is not a competitor; Storage is. \pia{} requires Living and at least one eligible competitor.

All 20,348 reference plans contain Living; 429 (2.11\%) also contain Dining. Living thus provides a consistently observed Public anchor. \pia{} retains Dining in both normalization and competition, whereas domain-level functional profiles and Public-domain margins aggregate Living and Dining as Public.

\paragraph{PIA Distance and metric roles.} Space-syntax analyses connect domestic organization to recurring integration inequalities between functional spaces~\citep[Ch.~1, pp.~25--27]{hillier1996space}. PIA quantifies one such relation: the lead of Living over its strongest eligible competitor. Comparing its distribution evaluates both the strength of this relation and its variation across residences. For cohort \(\mathcal C\), \(F_{\mathcal C}(t)\) is the fraction of defined per-plan PIA values no greater than \(t\); \(F_{\mathcal C}^{-1}\) denotes its quantile function. PIA Distance compares this distribution with the fixed reference:
\begin{equation}
\mathrm{PIAD}(\mathcal C,\mathcal C_{\mathrm{ref}})
=W_1\!\left(F_{\mathcal C},F_{\mathcal C_{\mathrm{ref}}}\right)
=\int_0^1\left|F_{\mathcal C}^{-1}(u)-F_{\mathcal C_{\mathrm{ref}}}^{-1}(u)\right|du,
\label{eq:piad}
\end{equation}
that is, the mean absolute gap between corresponding quantiles, in PIA's normalized-integration units~\citep[Section~2.6]{peyre2019computational}. PIAD is zero for identical distributions. It measures discrepancies in anchoring strength and spread relative to RPLAN, including departures toward stronger advantages. The reference therefore preserves residential variation rather than imposing one PIA target on every layout. In the 20,348 reference plans, median PIA is 0.6172 [25th, 75th percentiles: 0.4838, 0.7794], and Living leads in 98.98\% of them.

IPD compares four domain-wise cohort medians, describing the typical balance of integration across Public, Private, Service, and Outdoor functions. These medians do not determine which category leads within each layout or the distribution of its lead. PIAD first measures this within-plan contrast and then compares its full distribution; similar medians can conceal differences in the lower and upper tails. Conversely, the same Living advantage can coexist with different configurations of the remaining functions; IPD remains sensitive to changes in their domain medians. The two distances thus assess complementary properties of the same spatial organization. Across the eight benchmark systems, the two distances are correlated (Pearson \(r=0.93\), Spearman \(\rho=0.86\)) but order three pairs differently. For example, FloorplanGAN has lower IPD than Graph2Plan (0.0943 versus 0.1078) but higher PIAD (0.4340 versus 0.3974): its typical functional profile is closer to RPLAN, whereas Living fails to lead in 28.5\% of its accepted layouts, compared with 8.3\% for Graph2Plan (Table~\ref{tab:benchmark-pia}). We therefore report both distances rather than combining them into one score. Per-plan PIA explains individual layouts and paired changes, and median PIA describes a cohort's typical anchoring strength relative to the reference median. For generation, a condition-matched real plan provides one feasible design reference: its PIA need not be reproduced by another valid layout. A PIA difference describes a change in anchoring strength. In measurement-fidelity tests, the original plan instead anchors reconstruction error for the same underlying layout (Appendix~\ref{app:access-route-control}).

\par\addvspace{8pt}\noindent\begin{minipage}{\linewidth}
\centering
\renewcommand{\arraystretch}{1.12}
\captionof{table}{PIA distributions behind PIAD on the five-block benchmark. Living leads when \({\mathrm{PIA}>10^{-6}}\).}
\label{tab:benchmark-pia}
\small
\begin{tabular}{lrcrr}
\toprule
System & \piad{} $\downarrow$ & median \pia{} [25th, 75th percentiles] & Living leads & Accepted \\
\midrule
RPLAN reference & 0 & 0.6172 [0.4838, 0.7794] & 99.0\% & 20,348 plans \\
\midrule
CE2EPlan & 0.5816 & 0.0726 [-0.0862, 0.2375] & 61.4\% & 3,324/5,000 \\
DiffPlanner & 0.5548 & 0.1113 [-0.0503, 0.2603] & 67.5\% & 2,956/5,000 \\
House-GAN++ & 0.5288 & 0.1583 [0.0239, 0.2724] & 77.7\% & 2,036/5,000 \\
House-GAN & 0.4136 & 0.2607 [0.1628, 0.3479] & 94.2\% & 1,843/5,000 \\
Graph2Plan & 0.3974 & 0.2886 [0.1619, 0.3939] & 91.7\% & 3,582/5,000 \\
FloorplanGAN & 0.4340 & 0.2296 [-0.0557, 0.4714] & 71.5\% & 4,574/5,000 \\
ChatHD & 0.3517 & 0.3295 [0.1391, 0.5060] & 85.7\% & 1,561/5,000 \\
HD & 0.4129 & 0.2800 [0.1302, 0.3936] & 87.8\% & 4,723/5,000 \\
HD+\method{} & 0.2798 & 0.3974 [0.2873, 0.4931] & 98.1\% & 4,825/5,000 \\
\bottomrule
\end{tabular}
\end{minipage}\par\addvspace{8pt}

Let \(e_i\) indicate whether attempt \(i\) emits native output. After \ssio{} normalization and measurement, acceptance is
\begin{equation}
\mathbf 1_{\mathrm{acc}}(i)
=\mathbf 1\!\left[
e_i=1,\ z_i\ \text{is measurable},\
N_i^{\mathrm{ret}}=8,\ n_{i,\mathrm{Living}}>0
\right],\; \mathrm{Acc}
=\frac{\sum_i\mathbf 1_{\mathrm{acc}}(i)}{N_{\mathrm{attempted}}}.
\label{eq:benchmark-acceptance}
\end{equation}
Here \(N_i^{\mathrm{ret}}\) counts retained spatial instances and \(n_{i,\mathrm{Living}}\) counts retained Living rooms. Exactly eight instances including Living are required; graph connectivity is not. Missing or malformed outputs count toward total attempts. \ipd{} uses accepted outputs; \pia{}, and hence \piad{}, additionally requires an eligible competitor. Outputs failing a metric's requirements receive no imputed score.

\subsection{Comparison Scope}
\label{app:baselines}

The main benchmark compares eight systems under common measurement rules, with one output per condition: CE2EPlan~\citep{wang2026ce2eplan}, DiffPlanner~\citep{wang2025diffplanner}, House-GAN++~\citep{nauata2021houseganpp}, House-GAN~\citep{nauata2020housegan}, Graph2Plan~\citep{hu2020graph2plan}, FloorplanGAN~\citep{luo2022floorplangan}, ChatHouseDiffusion~\citep{qin2024chathousediffusion}, and \housediffusion{}. Separately, GSDiff~\citep{hu2025gsdiff} is evaluated as GSDiff--SSIO, with missing access supplied by a frozen completion model (Appendix~\ref{app:gsdiff-completion}). Figure~\ref{fig:results-overview}(a) shows all nine systems. GSDiff--SSIO also enters the five-backbone post-training comparison with completion fixed.

CE2EPlan, DiffPlanner, House-GAN, and FloorplanGAN use checkpoints trained or retrained on the common fewer-than-eight-room split.  \housediffusion{}, House-GAN++, ChatHouseDiffusion, and GSDiff use released checkpoints. Graph2Plan uses its released checkpoint with a reproduction of the released alignment procedure.

The released GSDiff checkpoint was pretrained on a split containing 4,241 of the 5,000 evaluation conditions. Its Base-to-post-training comparison therefore tests improvement from the same pretrained model with fixed access completion; it does not test generalization to conditions absent from pretraining.

Appendix~\ref{app:portability-results} reports each backbone's Base and post-training design scores and the number of accepted outputs.

\subsection{Cross-Corpus Grounding}\label{app:grounding-robustness}

We test whether the shared Public-space advantage and the difference between residential corpora persist when the accessibility measure changes or disconnected layouts are excluded. Table~\ref{tab:lifull-connected} compares RPLAN and LIFULL Public-domain margins using the scoring definitions in Appendix~\ref{app:proofs}.

\par\addvspace{8pt}\noindent\begin{minipage}{\linewidth}
\centering
\renewcommand{\arraystretch}{1.12}
\captionof{table}{Robustness of \(\Delta_{\mathrm{pub}}\) in RPLAN and LIFULL across centrality and connectivity choices.}
\label{tab:lifull-connected}
\small
\setlength{\tabcolsep}{2.5pt}
\begin{tabular}{llrrrr}
\toprule
Centrality & Cohort & \shortstack{Plan counts\\RPLAN / LIFULL} & \shortstack{RPLAN\\median} & \shortstack{LIFULL\\median} & Ratio \\
\midrule
Legacy HH & all eligible & 80,788 / 19,541 & 0.7162 & 0.3052 & 42.61\% \\
Legacy HH & connected only & 73,997 / 6,625 & 0.7062 & 0.3013 & 42.66\% \\
Bounded HH, \(n>2\) & all eligible & 80,710 / 18,554 & 0.2602 & 0.1210 & 46.48\% \\
Bounded HH, \(n>2\) & connected only & 73,997 / 6,590 & 0.2583 & 0.1268 & 49.08\% \\
Global harmonic & all eligible & 80,788 / 19,541 & 0.2791 & 0.1086 & 38.90\% \\
Global harmonic & connected only & 73,997 / 6,625 & 0.2759 & 0.1048 & 37.99\% \\
\bottomrule
\end{tabular}
\end{minipage}\par\addvspace{8pt}

The median Public-domain margin \(\Delta_{\mathrm{pub}}\) remains positive in every setting (Table~\ref{tab:lifull-connected}), while LIFULL's median is consistently lower: 37.99--49.08\% of RPLAN's. These results support a shared Public-first ordering, with numerical reference values matched to the residential context.

\subsection{Variation across Test-Condition Blocks}
\label{app:block-stability}

The pooled benchmark combines five disjoint blocks of test conditions. Table~\ref{tab:block-results} reports the mean and population standard deviation of the five block IPDs, PIADs, and PIA medians, to show how results vary across blocks. It covers the eight-system benchmark and HD post-training; GSDiff--SSIO block results appear with its separate analysis in Appendix~\ref{app:gsdiff-completion}.

\par\addvspace{8pt}\noindent\begin{minipage}{\linewidth}
\centering
\renewcommand{\arraystretch}{1.12}
\captionof{table}{Mean and population standard deviation across five evaluation blocks for the benchmark and primary post-training result. The main table pools all five blocks.}
\label{tab:block-results}
\small
\begin{tabular}{lcccc}
\toprule
& \multicolumn{3}{c}{Configurational metrics} &
\multicolumn{1}{c}{Protocol outcome} \\
\cmidrule(lr){2-4}\cmidrule(l){5-5}
Method & Block \ipd{} $\downarrow$ & Block \piad{} $\downarrow$ & Block median \pia{} &
Accepted \\
\midrule
CE2EPlan & 0.1916$\pm$0.0020 & 0.5815$\pm$0.0070 & 0.0745$\pm$0.0114 & 3,324/5,000 \\
DiffPlanner & 0.1806$\pm$0.0028 & 0.5550$\pm$0.0051 & 0.1119$\pm$0.0080 & 2,956/5,000 \\
House-GAN++ & 0.1611$\pm$0.0034 & 0.5292$\pm$0.0020 & 0.1599$\pm$0.0100 & 2,036/5,000 \\
House-GAN & 0.1385$\pm$0.0023 & 0.4139$\pm$0.0060 & 0.2617$\pm$0.0070 & 1,843/5,000 \\
Graph2Plan & 0.1074$\pm$0.0018 & 0.3974$\pm$0.0072 & 0.2899$\pm$0.0095 & 3,582/5,000 \\
FloorplanGAN & 0.0983$\pm$0.0084 & 0.4384$\pm$0.0211 & 0.2288$\pm$0.0079 & 4,574/5,000 \\
ChatHD & 0.0845$\pm$0.0046 & 0.3530$\pm$0.0245 & 0.3298$\pm$0.0171 & 1,561/5,000 \\
HD Base & 0.1112$\pm$0.0009 & 0.4130$\pm$0.0034 & 0.2804$\pm$0.0027 & 4,723/5,000 \\
HD+\method{} & \textbf{0.0813$\pm$0.0018} & \textbf{0.2800$\pm$0.0039} & 0.3991$\pm$0.0098 & \textbf{4,825/5,000} \\
\bottomrule
\end{tabular}
\end{minipage}\par\addvspace{8pt}

For HD, mean block \ipd{} decreases from 0.1112 to 0.0813, mean block \piad{} from 0.4130 to 0.2800, and mean block median \pia{} increases from 0.2804 to 0.3991. All three changes substantially exceed the corresponding standard deviations across the five disjoint condition blocks (Table~\ref{tab:block-results}).

\FloatBarrier
\section{SSPT Learning Mechanisms and Quality Control}

SSPT can select examples for diffusion retraining or update sampled trajectories directly. The following policy bindings specify each backbone's state, action, and trainable component.

\subsection{Data-Space Selection}
\label{app:sspt-iter}

\begin{figure}[!htbp]
  \centering
  \includegraphics[width=\linewidth]{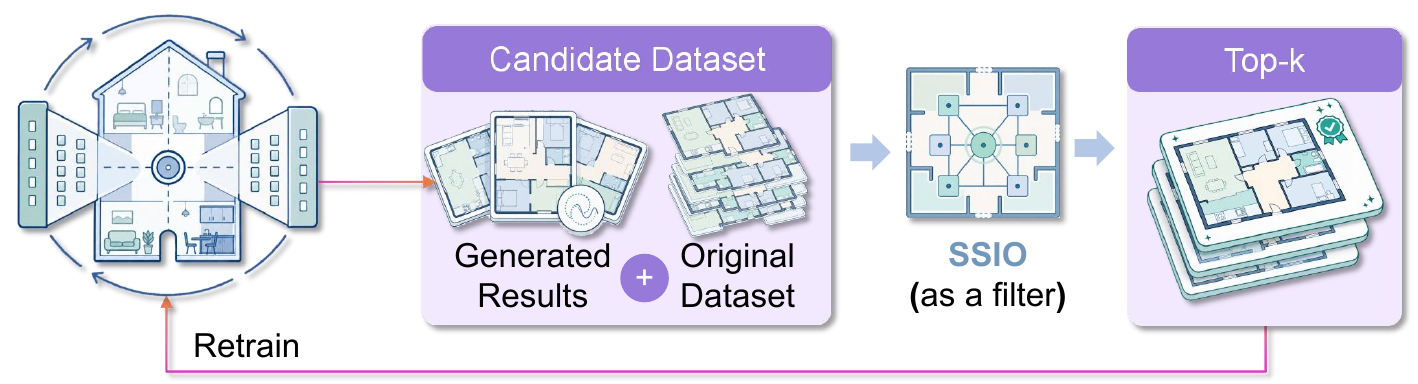}
  \caption{\textbf{SSPT-Iter: design feedback through data selection.} Generated candidates and a fixed real replay pool are ranked by SSIO scores. The top-$K$ selection supplies available targets for ordinary diffusion retraining before the next round.}
  \label{fig:sspt-iter}
\end{figure}

\sspt{}-Iter ranks a fixed set of real examples together with \(N\) generated candidates by terminal reward (Figure~\ref{fig:sspt-iter}). The top \(K\) examples provide ordinary diffusion training targets where these are available. Examples without targets are skipped and not replaced. Ranking retains source identities and breaks ties deterministically; rejected candidates provide no negative feedback.

Appendix~\ref{app:sspt-iter-pilot} reports the controlled selection pilot and its contribution to the development of trajectory-space post-training.

\subsection{Backbone Policy Bindings}
\label{app:backbone-bindings}

\subsubsection{Terminal Measurement and Reward Support}

Backbone \(m\) computes rewards with training settings \(\eta_m^{\mathrm{train}}\) (Table~\ref{tab:training-bindings}). Final evaluation rescores all outputs with common settings \(\eta^{\mathrm{eval}}\), including a 50 pixels\(^2\) minimum atom area.

A training reward is valid, \(v_m(x)=1\), only after native-output normalization succeeds. HD and ChatHD also require a scored Living space. CE2EPlan and DiffPlanner require finite scores for every native room and successful identification and realization of all official doors and access openings, with no separate Living check. GSDiff requires finite measurements and consistent room counts. Empty maxima in \(R_{\mathrm{anchor}}\) are zero. These training rules differ from benchmark acceptance, which requires retained Living for every backbone.

For each backbone, a frozen set of Base outputs fixes \(q_A\) at the 90th percentile of total retained-atom area and \(s_A=\max\{\mathrm{IQR}^{\mathrm{base}},1\ \mathrm{pixel}^2\}\), where IQR is the interquartile range of that area. For the primary \housediffusion{} run, \(q_A=22{,}134.4\) pixels\(^2\) and \(\mathrm{IQR}^{\mathrm{base}}=3{,}621.5\) pixels\(^2\); the selected anchor weight is 8 and the \areaguard{} weight is \(\lambda=4\).

\subsubsection{Shared Trajectory Objective}

At outer iteration \(r\), the frozen model snapshot \(p_{\mathrm{old},m}=p_{\theta_r,m}\) samples trajectory set \(\mathcal P_m=\{\tau_i^{(m)}\}_{i=1}^{N_m}\). We save each condition, retained state--action--log-probability tuple, and final native output:
\begin{equation}
\tau_i^{(m)}=
\left(c_i,
\left(s_{i,d}^{(m)},a_{i,d}^{(m)},
\ell_{\mathrm{old},i,d}^{(m)}\right)_{d\in\mathcal D_i^{(m)}},
x_{i,m}^{\mathrm{nat}}\right).
\label{eq:appendix-trajectory}
\end{equation}
The recorded state and fixed mask determine which stochastic coordinates \(\Omega_{i,d}^{(m)}\) enter the policy likelihood. PPO reuses this trajectory set for every epoch in the iteration, then samples a new set from the updated model.

PPO averages the clipped objective over the retained trajectory--decision pairs in this fixed set:
\begin{equation}
\widehat J_m(\theta)=\frac{1}{|\mathcal B_m|}
\sum_{(i,d)\in\mathcal B_m}
\min\!\left\{
\widetilde\rho_{i,d}^{(m)}\widehat A_i,\,
\operatorname{clip}\!\left(\widetilde\rho_{i,d}^{(m)},1-\epsilon,1+\epsilon\right)
\widehat A_i
\right\},
\label{eq:appendix-finite-ppo}
\end{equation}
Training minimizes \(-\widehat J_m\). Rewards are clipped, centered, and standardized before replay. The resulting advantage \(\widehat A_i\) is detached from gradients and fixed across all recorded decisions in trajectory \(i\).

Each backbone assigns a joint diagonal-Gaussian likelihood to its active stochastic coordinates (Table~\ref{tab:backbone-bindings}):
\begin{equation}
\log p_{\theta,m}^{(d)}(a_{i,d}^{(m)}\mid s_{i,d}^{(m)})
=-\frac{1}{2}\sum_{j\in\Omega_{i,d}^{(m)}}
\left[
\log(2\pi v_{m,d,j})+
\frac{(a_{i,d,j}^{(m)}-\mu_{\theta,m,d,j})^2}{v_{m,d,j}}
\right],
\label{eq:shared-native-density}
\end{equation}
where \(\mu_{\theta,m,d}\) is produced by the native reverse kernel from \(s_{i,d}^{(m)}\), and \(v_{m,d}\) is its fixed reverse variance.

\par\addvspace{8pt}\noindent\begin{minipage}{\linewidth}
\centering
\renewcommand{\arraystretch}{1.12}
\captionof{table}{Native policy bindings used by the five reported diffusion backbones.}
\label{tab:backbone-bindings}
\small
\setlength{\tabcolsep}{4pt}
\begin{tabular}{>{\raggedright\arraybackslash}p{0.14\textwidth}>{\raggedright\arraybackslash}p{0.28\textwidth}>{\raggedright\arraybackslash}p{0.27\textwidth}>{\raggedright\arraybackslash}p{0.21\textwidth}}
\toprule
Backbone & Native variable & Active coordinates & Trainable component \\
\midrule
HD &
\(2\times100\) coordinate tensor &
all 200 coordinates &
denoiser \\
CE2EPlan &
\(6\times L_m\) room tensor \([\mathrm{pres},\mathrm{type},x,y,w,h]\) &
unpadded \mbox{\(x,y,w,h\) coordinates} &
denoiser \\
DiffPlanner &
\(4\times L_m\) room-box tensor \(xyxy\) &
unpadded room boxes &
PartitioningDiff \\
ChatHD &
\(1\times64\times64\) semantic raster &
interior raster pixels &
suffix U-Net \\
GSDiff--SSIO &
\(53\times10\) corner/semantic tensor &
joint reverse tensor &
main diffusion model \\
\bottomrule
\end{tabular}
\end{minipage}\par\addvspace{8pt}

\subsubsection{Sampling Entry and Reverse-Time Support}

Each training trajectory contains \(D_m=99\) stochastic decisions between 100 retained reverse-time points. The \((80,20,0,0)\) schedule retains 80, 20, 0, and 0 points from the four equal sections of the original 1,000-step axis. Sampling visits these retained points in reverse order, from \(499\to0\).

HD, CE2EPlan, DiffPlanner, and GSDiff initialize this shortened chain with standard Gaussian noise, subject to native condition masks, without first running a high-noise prefix. ChatHD (ChatHouseDiffusion) instead enters through the frozen prefix described below. Each terminal output then undergoes \ssio{} normalization \(\mathcal A_m\).

Final evaluation keeps the same respaced sampler for CE2EPlan, DiffPlanner, and GSDiff, and the same prefix--suffix sampler for ChatHD. HD is evaluated with its original full 1,000-step ancestral sampler, for both the benchmark and native-quality metrics; its lower-noise schedule determines post-training trajectories rather than replacing the final native sampler.

\paragraph{Concrete reverse kernel.} Let \(\kappa_0<\cdots<\kappa_{99}\) be the original timesteps retained by \((80,20,0,0)\). Specifically, \(\kappa_e=\operatorname{round}(249e/79)\) for \(0\leq e<80\), and \(\kappa_{80+j}=250+\operatorname{round}(249j/19)\) for \(0\leq j<20\). Sampling proceeds from effective position \(e=99\) to \(e=0\).

To connect retained timestep positions to the main-paper decision index, suppress \(i\) and \(m\) and write \(x_e:=x_{i,\kappa_e}^{(m)}\). During a stochastic transition \(e\to e-1\), the decision index is \(d=100-e\), so \(t_d^{(m)}=\kappa_e\) and \(t_{d+1}^{(m)}=\kappa_{e-1}\). The last recorded action is \(e=1\to0\).

The deterministic final \(e=0\) kernel uses rollout parameters \(\theta_r\) and emits or decodes \(x_{i,m}^{\mathrm{nat}}\). PPO covers only the recorded stochastic decisions; saved outputs and rewards remain fixed during replay.

For \housediffusion{}, the state contains \((x_e,\kappa_e,c)\). Write \(a_e:=a_{i,d}^{(m)}=x_{e-1}\) and \(p_\theta^{(e)}:=p_{\theta,m}^{(d)}\) for \(d=100-e\). With fixed respaced variance \(v_e>0\), its policy is
\begin{equation}
\begin{aligned}
p_\theta^{(e)}(a_e\mid x_e,c)
&=\mathcal N\!\left(
\operatorname{vec}(a_e);
\operatorname{vec}(\mu_\theta^{(e)}(x_e,c)),v_e I_{200}\right),\\
\log p_\theta^{(e)}(a_e\mid x_e,c)
&=-\frac12\sum_{j=1}^{200}
\left[\log(2\pi v_e)+
\frac{\big(a_{e,j}-[\mu_\theta^{(e)}(x_e,c)]_j\big)^2}{v_e}\right].
\end{aligned}
\label{eq:implemented-kernel}
\end{equation}
The primary \housediffusion{} experiment averages the surrogate over all 99 stochastic transitions. At \(e<32\), its binary branch replaces the continuous output and is hard-quantized before the policy mean is computed, giving zero parameter gradient. Nonzero gradients therefore arise from \(e=32,\ldots,99\) through the shared Transformer and continuous-coordinate head; their retained original timesteps span \(101\) through \(499\).

\subsubsection{Backbone Instantiations}

In CE2EPlan's type-conditioned mode, presence and category are restored from the condition after every transition. The likelihood covers only unpadded center/size coordinates \(x,y,w,h\), updating the shared trunk, condition encoders, and \(xywh\) output rows. Deterministic \ssio{} postprocessing derives adjacency from the generated boxes.

DiffPlanner fixes the upstream outputs of NodeDiff and AdjacencyDiff and trains PartitioningDiff. Deterministic alignment and decoration are applied during \ssio{} normalization.

ChatHD first executes 25 denoising diffusion implicit model (DDIM) transitions with \(\eta=0\), from \(t=999\) to \(t=499\), with a frozen Base U-Net. The ancestral suffix trains a separate U-Net copy through interior-pixel actions; terminal class quantization precedes \ssio{} normalization.

GSDiff trains its topology-conditioned main diffusion model through continuous corner coordinates. The topology encoder and terminal edge stage are frozen; hard semantic thresholds block direct gradients through the semantic channels, whose predicted values can still change as the diffusion model is updated. Its \ssio{} normalization applies \(\mathcal C_\phi\) before convex-space graph construction.

Table~\ref{tab:training-bindings} gives the training settings; all bindings use \(\epsilon=0.2\).

\par\addvspace{8pt}\noindent\begin{minipage}{\linewidth}
\centering
\renewcommand{\arraystretch}{1.12}
\captionof{table}{Numerical settings of the backbone training bindings.}
\label{tab:training-bindings}
\small
\setlength{\tabcolsep}{4pt}
\begin{tabular}{lrrrrr}
\toprule
Backbone & \shortstack{Iterations $\times$\\rollouts per iteration} & \shortstack{PPO\\epochs} & \shortstack{Learning\\rate} & \(M_m\) &
\shortstack{Min. atom\\area (px$^2$)} \\
\midrule
HD & \(20\times10{,}000\) & 2 & \(5\times10^{-6}\) & 10 & 50 \\
CE2EPlan & \(20\times128\) & 2 & \(5\times10^{-6}\) & 10 & 16 \\
DiffPlanner & \(60\times512\) & 4 & \(10^{-6}\) & 10 & 16 \\
ChatHD & \(5\times512\) & 2 & \(5\times10^{-6}\) & 10 & 50 \\
GSDiff--SSIO & \(10\times512\) & 2 & \(5\times10^{-6}\) & 20 & 16 \\
\bottomrule
\end{tabular}
\end{minipage}\par\addvspace{8pt}

\FloatBarrier
\subsection{Native-Quality Evaluation}
\label{app:native-quality}

We evaluate HD's native output quality using its automatic benchmark~\citep{shabani2023housediffusion}. Generated room and door polygons and their condition-matched real counterparts are rendered at \(256\times256\) pixels with the same semantic colors, black outlines, and white background. FID compares their 2,048-dimensional Inception features using \texttt{pytorch-fid}. The real images come from this native evaluation cohort, separate from the 20,348-plan reference used for \ipd{}.

Compatibility counts missing or extra access edges in generated layouts relative to the requested graph, then averages this count over layouts. Generated edges are recovered from door--room overlaps; the exterior door connects to an outside node. Since the input fixes room identities, the modified graph-edit score compares edges directly without searching over node permutations. Lower scores mean closer agreement with the requested connections; functional organization is evaluated separately by the design metrics.

Checkpoint validation and final testing each use a separate fixed cohort of 1,000 exact-eight conditions. We generate five outputs per condition, using seeds 0--4 consistently across checkpoints, and report the mean and population standard deviation of the five evaluation scores. Final testing uses generation batches of 1,000 and FID feature batches of 64. Native metrics use all generated outputs, before \rplanssi{} acceptance filtering. Base and the selected checkpoint both use HD's original full 1,000-step ancestral sampler.

For the AreaGuard development comparison in Figure~\ref{fig:results-overview}(c), all weights are evaluated at iteration 10 on the native loader's exact-eight cohort, with five repeats and generation batch 2,048. The Base marker is an independent evaluation, so it is a quality reference rather than a matched development result. The dashed limits are the Base-calibrated ceilings. Table~\ref{tab:native-evaluation} reports the matched final Base--SSPT comparison using the fixed conditions and seeds above.

\subsection{Base-Referenced Quality Calibration and Model Selection}
\label{app:selection}

\paragraph{Purpose and quality criteria.} We select HD checkpoints for improved functional organization subject to native-quality limits. FID measures discrepancy from real renderings; Compatibility measures disagreement with requested access conditions. We constrain their five-repeat means using upper prediction limits calibrated from unchanged Base HD, following the \(\varepsilon\)-constraint principle~\citep{mavrotas2009epsilon}. Calibration accounts for variation across repeated evaluations of the same generator. \areaguard{} controls area expansion during training; the quality limits are applied afterward to select checkpoints.

\paragraph{Calibration data.} We retrospectively calibrate limits from three archived Base batches, each containing five evaluations of 1,000 exact-eight conditions. A and B use the loader-default protocol and share the saved real-image cohort; V uses the explicit native-validation condition list. Batch-specific means and variances determine one ceiling per metric covering both protocols. Repeated seeded records of V are counted once; the code provides all source scores. Calibration excludes post-trained design scores and final-test outcomes. The resulting limits replace the earlier empirical limits, and selection is reapplied to the archived validation candidates.

\paragraph{Prediction target and assumptions.} Checkpoint eligibility is based on five-repeat means, so calibration predicts a new five-repeat Base mean. Let \(X_{b,j,r}\) denote repeat \(r\) of metric \(j\in\{\mathrm{FID},\mathrm{Comp}\}\) in batch \(b\in\{A,B,V\}\), with mean \(\bar X_{b,j}\) and sample standard deviation \(s_{b,j}\) computed using denominator \(n_b-1\). Assuming independent normally distributed repeat scores within a batch, an independent new \(m\)-repeat Base mean \(\bar X^{\mathrm{new}}_{b,j}\) satisfies the Student prediction relation~\citep{hahn1991statisticalintervals}
\begin{equation}
\frac{\bar X^{\mathrm{new}}_{b,j}-\bar X_{b,j}}
{s_{b,j}\sqrt{1/n_b+1/m}}
\sim t_{n_b-1},
\qquad n_b=m=5.
\label{eq:base-prediction-pivot}
\end{equation}
The variance terms account for uncertainty in both the estimated Base mean and the new evaluation. Using only \(1/n_b\) would omit the latter. Normality and independence are assumptions, and five repeats cannot establish the accuracy of the predicted tails. The early batches lack recorded repeat seeds; V records seeds 0--4. The prediction requires independent new sampling; paired seeds can still be used to compare candidate models.

\paragraph{Deriving the joint limits.} We allocate a nominal 5\% upper-tail error budget across the three batches and two metrics. This bounds the probability of excluding a new unchanged-Base mean; it is not a percentage allowance for model degradation. With \(\alpha=0.05\), \(B=3\), and \(J=2\), Bonferroni allocation gives
\begin{equation}
U_{b,j}
=\bar X_{b,j}+t_{1-\alpha/(BJ),\,n_b-1}\,
s_{b,j}\sqrt{1/n_b+1/m},\; \tau_j^{\mathrm{raw}}=\max_b U_{b,j},
\;
\tau_j=\frac{\lceil10\tau_j^{\mathrm{raw}}\rceil}{10}.
\label{eq:base-reference-limit}
\end{equation}
Taking the largest batch-specific bound gives a common ceiling for the observed development settings while keeping their distributions separate. Independence between calibration batches is unnecessary because the union bound gives
\begin{equation}
\Pr\!\left(\exists\,b,j:\bar X^{\mathrm{new}}_{b,j}>\tau_j\right)
\leq\sum_{b,j}\Pr\!\left(\bar X^{\mathrm{new}}_{b,j}>U_{b,j}\right)
\leq\alpha
\label{eq:base-reference-coverage}
\end{equation}
under the stated prediction model, across repeated calibration and new Base evaluations. We choose the 5\% budget and round each limit upward to one decimal; this preserves nominal coverage and simplifies its use. Table~\ref{tab:base-reference-calibration} retains the unrounded bounds. Here \(t_{0.9916667,4}=3.9607865\), giving \(\tau_{\mathrm{FID}}^{\mathrm{raw}}=9.4243\) and \(\tau_{\mathrm{Comp}}^{\mathrm{raw}}=2.7822\), which round to \(\tau_{\mathrm{FID}}=9.5\) and \(\tau_{\mathrm{Comp}}=2.8\).

\par\addvspace{8pt}\noindent\begin{minipage}{\linewidth}
\centering
\renewcommand{\arraystretch}{1.12}
\captionof{table}{\textbf{Base-only calibration of the native-quality range.} Each entry uses five 1,000-condition repeats. Standard deviations \(s\) use denominator four. The upper bounds jointly cover six future Base means under Eq.~\eqref{eq:base-reference-limit}; bold entries determine the common ceilings.}
\label{tab:base-reference-calibration}
\small
\setlength{\tabcolsep}{5pt}
\begin{tabular}{lrrrrrr}
\toprule
& \multicolumn{3}{c}{FID} & \multicolumn{3}{c}{Compatibility} \\
\cmidrule(lr){2-4}\cmidrule(lr){5-7}
Base batch & Mean & \(s\) & \(U_{b,j}\) & Mean & \(s\) & \(U_{b,j}\) \\
\midrule
A: loader-default & 8.5157 & 0.1670 & 8.9342 & 2.5240 & 0.0650 & 2.6869 \\
B: loader-default & 8.6177 & 0.3220 & \textbf{9.4243} & 2.5066 & 0.0207 & 2.5586 \\
V: native validation & 7.4739 & 0.0996 & 7.7234 & 2.6020 & 0.0719 & \textbf{2.7822} \\
\midrule
Adopted upper limit & \multicolumn{3}{c}{9.5} & \multicolumn{3}{c}{2.8} \\
\bottomrule
\end{tabular}
\end{minipage}\par\addvspace{8pt}

\paragraph{Scope of the calibrated range.} These limits describe variation in repeated unchanged-Base measurements across the recorded development settings. A post-trained checkpoint can satisfy them yet have larger errors than Base on the same cohort; final evaluation therefore reports the matched differences alongside design gains. Passing the limits does not establish perceptual equivalence. Changes to rendering, sample counts, or condition populations require recalibration. Smaller tail-error budgets widen the range: \(\alpha=0.01,0.05,0.10\) gives rounded FID/Compatibility limits of 9.9/2.9, 9.5/2.8, and 9.3/2.8, respectively. Appendix~\ref{app:selection-results} compares the checkpoints selected under these ceilings and alternative quality requirements.

\paragraph{Checkpoint selection.} In the primary \housediffusion{} experiment, an eligible checkpoint has at least 100 validation plans per functional-profile entry, at least one valid \pia{} score, and acceptance of at least 0.90. All five native-metric repeats must be finite. Mean FID~\citep{heusel2017fid} must not exceed \(\tau_{\mathrm{FID}}\), and mean Compatibility must not exceed \(\tau_{\mathrm{Comp}}\). From the nonempty finite set of eligible checkpoints \(\Theta_E\), we select by the following ordered criteria:
\begin{equation}
\theta^*=\operatorname*{arg\,min}^{\mathrm{lex}}_{\theta_j\in\Theta_E}
\big(
\mathrm{IPD}_{\mathrm{val}}(\theta_j),
-\operatorname{med}\mathrm{PIA}_{\mathrm{val}}(\theta_j),
\mathrm{FID}_{\mathrm{val}}(\theta_j),
\mathrm{Comp}_{\mathrm{val}}(\theta_j),j
\big),
\label{eq:selection}
\end{equation}
where \(\mathrm{FID}_{\mathrm{val}}\) and \(\mathrm{Comp}_{\mathrm{val}}\) are five-repeat means. We first minimize validation IPD, then break ties by higher median PIA, lower FID, lower Compatibility, and finally the fixed candidate order \(j\). All comparisons use unrounded values. No IPD ties occurred, so median PIA was never used. Archived validation records keep cohort summaries, not per-plan PIA values, so validation PIAD is unavailable; we keep the executed rule and report final-test PIAD for the selected checkpoint.

\paragraph{Independent final evaluation.} We compare the selected checkpoint with matched Base HD on design and native-quality measures (Tables~\ref{tab:native-evaluation} and~\ref{tab:portable-full}). Final-test outcomes and candidate design rankings are excluded from Base-only calibration. Appendix~\ref{app:selection-results} reports the checkpoint selected after recalibration and compares alternative validation choices.

\subsection{Ablations and Method Development}
\label{app:ablation-development}

We first report the selection-and-retraining pilot, then isolate the effects of update location, reward, and \areaguard{} weight in HD. Validation subsequently selects the weight and training duration. These development and validation experiments use different data from the final tests in Tables~\ref{tab:native-evaluation} and~\ref{tab:portable-full}.

\subsubsection{Data-Space Selection Pilot}
\label{app:sspt-iter-pilot}

The selection-and-retraining pilot runs for three rounds. Each round ranks \(N=10{,}000\) generated candidates together with 80,788 real replay examples and selects the top \(K=4{,}000\). All 12,000 selections come from the real examples; 141 selected examples per round lack a training target, leaving 3,859 update targets per round. Thus none of the 30,000 generated candidates contributes to learning. This pilot motivates using sampled trajectories directly: SSPT can assign positive or negative advantages to generated outputs instead of discarding every output ranked below the real examples.

\subsubsection{Reverse-Time Support Ablation}
\label{app:reverse-support}

To compare update locations, we fix the profile-distance reward and retain 100 of HouseDiffusion's original 1,000 timesteps. Across four equal sections ordered from low to high noise, schedules \((80,20,0,0)\), \((25,25,25,25)\), and \((0,0,20,80)\) give lower-noise, uniform full-horizon, and higher-noise updates. Each uses 5,000 trajectories per iteration, seed 0, two PPO epochs, microbatch size 32, and gradient accumulation over four microbatches. All are evaluated at iteration 3 on 1,000 exact-eight conditions each (Table~\ref{tab:region-dev}).

Lower-noise updates give lower \ipd{} and \piad{} than either alternative (Table~\ref{tab:region-dev}). With the reward and retained-step budget fixed, the feedback is more effective late in HD's denoising process.

\subsubsection{Reward Identity}
\label{app:reward-development}

We compare rewards with the sampling schedule and training budget fixed: 10,000 rollout trajectories per iteration, the same lower-noise timesteps, and evaluation at iteration 10 on a 5,000-condition development set separate from final testing. Living integration lead is Eq.~\eqref{eq:guard} with \(\lambda=0\). RPLAN profile alignment is the negative mean absolute distance between a layout's available category coordinates and their RPLAN medians. Standardized Living contrast compares Living's category score with the non-Living category median, divided by a robust scale. Living lead + RPLAN alignment adds the first two signals. If a reward's required scores are unavailable, we assign the fixed failure value rather than average the available terms.

\par\addvspace{8pt}\noindent\begin{minipage}{\linewidth}
\centering
\renewcommand{\arraystretch}{1.12}
\captionof{table}{Reward identity on the development pool at iteration 10.}
\label{tab:reward-dev}
\small
\begin{tabular}{lrrrr}
\toprule
& \multicolumn{3}{c}{Configurational metrics} &
\multicolumn{1}{c}{Protocol outcome} \\
\cmidrule(lr){2-4}\cmidrule(l){5-5}
Reward & \ipd{} $\downarrow$ & \piad{} $\downarrow$ & median \pia{} & Accepted \\
\midrule
Living integration lead & 0.0830 & 0.2855 & 0.3912 & 4753/5000 \\
RPLAN profile alignment & 0.0828 & 0.2869 & 0.3903 & 4782/5000 \\
Standardized Living contrast & 0.0927 & 0.3243 & 0.3554 & 4747/5000 \\
Living lead + RPLAN alignment & \textbf{0.0807} & \textbf{0.2754} & 0.4027 & 4764/5000 \\
\bottomrule
\end{tabular}
\end{minipage}\par\addvspace{8pt}

Living integration lead and RPLAN profile alignment yield similar design scores, both better than Standardized Living contrast (Table~\ref{tab:reward-dev}). Their sum improves both scores further, but FID rises from the matched Base value of \(7.474\pm0.089\) to \(21.470\pm0.244\), above the 9.5 limit. The next experiment holds Living integration lead fixed and tests whether penalizing area expansion reduces this quality cost.

\subsubsection{AreaGuard Weight}
\label{app:area-guard-ablation}

\textbf{What does AreaGuard contribute?} We hold Living integration lead and training duration at iteration 10 fixed while varying the \areaguard{} weight (Table~\ref{tab:guard-dev}, Figure~\ref{fig:quality-guard-tradeoff}). Weight 2 reduces FID from 19.55 to 8.91 and Compatibility from 2.583 to 2.118 while keeping design scores close: \ipd{} changes from 0.0830 to 0.0827 and \piad{} from 0.2855 to 0.2890. Weight 4 further improves these native-quality scores but yields weaker design scores than weight 2. We then use independent validation to select the weight and training duration jointly under the quality limits.

\begin{figure}[!htbp]
  \centering
  \includegraphics[page=2,width=\linewidth]{figures/supp_quality_tradeoff.pdf}
  \caption{\textbf{AreaGuard weight at a fixed training duration.} Living integration lead and iteration 10 are held fixed. Design metrics use 5,000 development conditions; native metrics use five repeats of 1,000 conditions. Error bars show repeat standard deviations; dashed lines mark the calibrated Base-reference ceilings.}
  \label{fig:quality-guard-tradeoff}
\end{figure}

\subsubsection{AreaGuard Weight and Training Duration}
\label{app:selection-results}

We apply the Base-calibrated selection rule in Eq.~\eqref{eq:selection} to nine archived checkpoints. Each has design scores on 2,000 validation conditions and five native-quality evaluations of 1,000 conditions. Four satisfy the FID/Compatibility limits of 9.5/2.8, and \areaguard{}-4 at iteration 20 has the lowest \ipd{} among them (Table~\ref{tab:selection-path}). Replacing the earlier limits with the calibrated limits changes neither eligibility nor selection: candidates that now satisfy Compatibility still exceed the FID ceiling. The model and outputs used for independent final testing therefore remain unchanged.

\par\addvspace{8pt}\noindent\begin{minipage}{\linewidth}
\centering
\renewcommand{\arraystretch}{1.12}
\captionof{table}{\textbf{Validation separates eligibility from IPD-based selection.} Native metrics are five-repeat means; acceptance is a fraction. All candidates satisfy support and acceptance requirements. Status applies the Base-calibrated FID/Compatibility ceilings of 9.5/2.8; the selected row is bold. Here, \(\lambda\) is the \areaguard{} weight.}
\label{tab:selection-path}
\small
\setlength{\tabcolsep}{3.5pt}
\begin{tabular}{lcccccc}
\toprule
\(\lambda\) & Iter. & \ipd{} $\downarrow$ & Acc. & FID $\downarrow$ & Comp. $\downarrow$ & Status \\
\midrule
2 & 10 & 0.0842 & 0.9665 & 8.594 & 2.210 & Eligible \\
2 & 15 & 0.0812 & 0.9610 & 10.061 & 2.430 & Excluded \\
2 & 19 & 0.0778 & 0.9560 & 12.330 & 2.640 & Excluded \\
2 & 22 & 0.0811 & 0.9510 & 13.557 & 2.787 & Excluded \\
2 & 25 & 0.0822 & 0.9445 & 15.084 & 2.967 & Excluded \\
2 & 30 & 0.0835 & 0.9210 & 19.145 & 3.343 & Excluded \\
\addlinespace[3pt]
4 & 10 & 0.0843 & 0.9705 & 8.176 & 2.184 & Eligible \\
4 & 15 & 0.0839 & 0.9690 & 8.462 & 2.331 & Eligible \\
\textbf{4} & \textbf{20} & \textbf{0.0835} & \textbf{0.9630} & \textbf{9.359} & \textbf{2.581} & \textbf{Selected} \\
\bottomrule
\end{tabular}
\end{minipage}\par\addvspace{8pt}

\begin{figure}[!htbp]
  \centering
  \includegraphics[page=1,width=\linewidth]{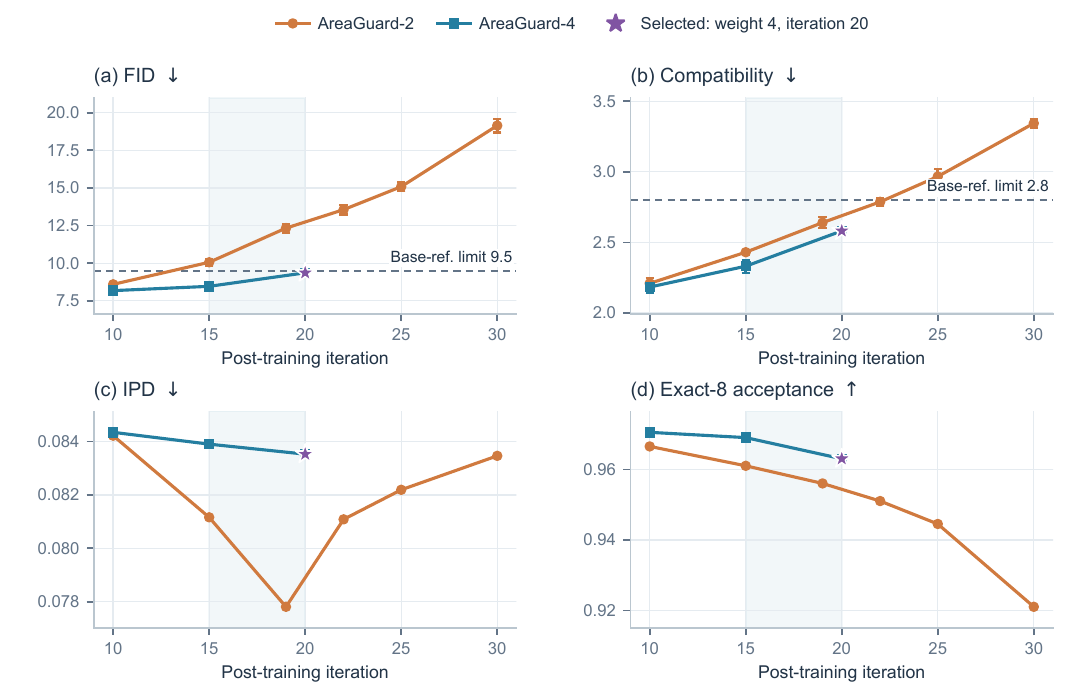}
  \caption{\textbf{Training duration and the quality--design trade-off.} Lines connect the nine measured validation checkpoints in Table~\ref{tab:selection-path}; native-metric error bars show repeat standard deviations. Shading marks iterations 15--20, where \areaguard{}-4 yields smaller additional design gains at increasing native cost. The star identifies the selected checkpoint. Dashed lines mark the Base-calibrated ceilings.}
  \label{fig:quality-training-tradeoff}
\end{figure}

\textbf{How does training duration change the trade-off?} With \areaguard{}-2, validation \ipd{} improves through iteration 19, but FID exceeds its limit from iteration 15; later checkpoints worsen on both design and native-quality metrics. With \areaguard{}-4, all three checkpoints satisfy the limits, and validation \ipd{}, the selection criterion, keeps improving through iteration 20 (Figure~\ref{fig:quality-training-tradeoff}). Between iterations 15 and 20, \ipd{} falls from 0.08390 to 0.08352, while FID increases from 8.462 to 9.359, Compatibility from 2.331 to 2.581, and acceptance falls from 0.9690 to 0.9630. Iteration 15 offers better native quality; iteration 20 is selected for its slightly lower validation \ipd{} within the permitted quality range.

\textbf{How sensitive is selection to the quality requirement?} Figure~\ref{fig:quality-gate-sensitivity} changes the ceilings while keeping the nine measured candidates and ranking rule fixed. FID at 9.0, or Compatibility at 2.5, selects \areaguard{}-4 at iteration 15. FID at 10.1 with Compatibility at 2.8 selects \areaguard{}-2 at iteration 15, improving design scores at higher FID. The adopted 9.5/2.8 limits select \areaguard{}-4 at iteration 20, as do the unrounded limits of 9.4243/2.7822. Rounding therefore does not affect selection. This analysis shows how alternative quality requirements change the selected model; the limits themselves come from the Base calibration. Table~\ref{tab:native-evaluation} reports the selected model's independent final test.

\begin{figure}[!htbp]
  \centering
  \includegraphics[page=3,width=\linewidth]{figures/supp_quality_tradeoff.pdf}
  \caption{\textbf{Checkpoint selection under different quality requirements.} (a) Validation FID and Compatibility; shading marks the admissible region and error bars show repeat standard deviations. (b) Selection with the candidate set and Eq.~\eqref{eq:selection} fixed; labels denote the \areaguard{} weight and training iteration. The star marks the Base-calibrated ceilings, 9.5/2.8, selecting \areaguard{}-4 at iteration 20. Stricter ceilings can select iteration 15.}
  \label{fig:quality-gate-sensitivity}
\end{figure}

\subsubsection{Interpreting the Base-Reference Range}
\label{app:quality-range-diagnostics}

To interpret the FID ceiling, we measure each saved output's foreground and axis-aligned bounding-box areas relative to its condition-matched real image, without design-score selection; a ratio of one means equal area. Mean foreground ratios are 0.954 and 0.951 for the two loader-default Base cohorts, 1.380 for Living integration lead alone, and 1.028 and 0.994 with \areaguard{}-2 and \areaguard{}-4, respectively. Foreground reaches the outermost border in 0.1\% of Base and 5.9\% of unguarded renderings. These measurements link higher FID to rendered expansion, not physical floor area.

\begin{figure}[!htbp]
\centering
\includegraphics[width=\linewidth]{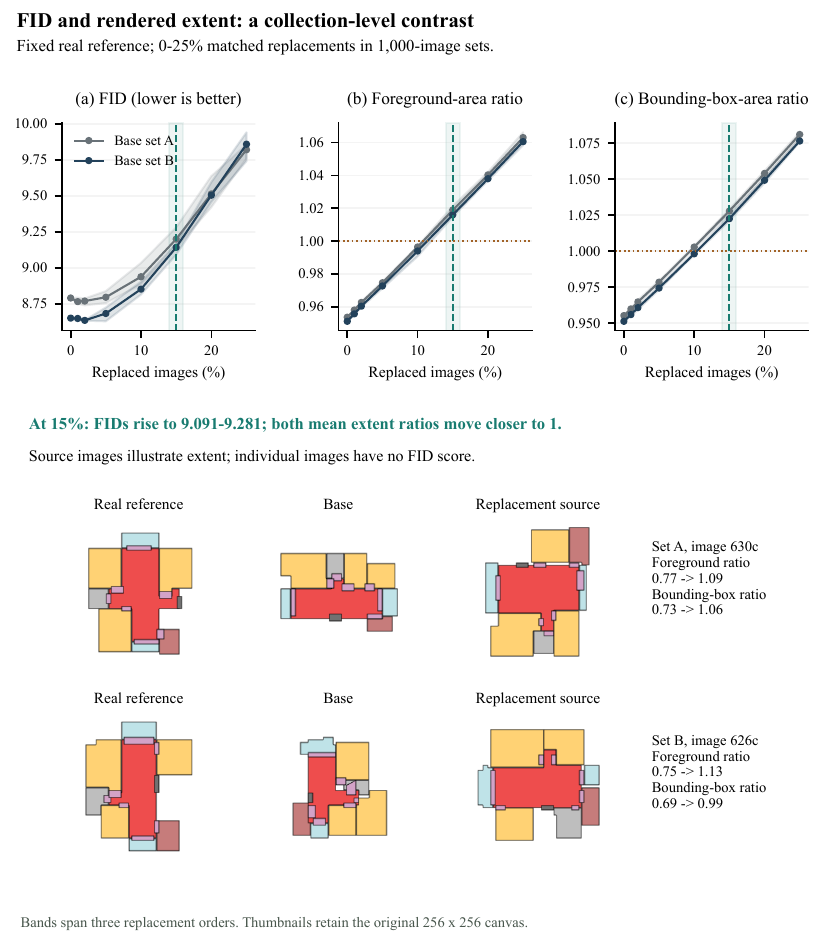}
\caption{\textbf{FID and rendered extent under fixed-reference collection changes.} Curves average three replacement orders for each of two Base sets; bands show their ranges. All six 15\% collections have higher FID but mean foreground- and bounding-box-area ratios closer to the real reference. Thumbnails are unchanged source images from the actual replacement subset, selected near the median extent-error reduction under the documented conditions. FID applies to collections; thumbnails illustrate extent. These development cohorts are separate from the selected-checkpoint final test.}
\label{fig:base-range-response}
\end{figure}

We then form mixed image sets by replacing selected Base outputs with outputs from the unguarded model under the same conditions. All 1,000 real reference images remain fixed. Two Base cohorts, three fixed nested replacement orders, and fractions of 0, 1, 2, 5, 10, 15, 20, 25, 50, and 100\% produce 60 sets. We compute FID from each set's 2,048-dimensional features, rather than interpolate the Base and unguarded scores. Recomputed saved-endpoint FIDs differ from the archived values by less than 0.01. Figure~\ref{fig:base-range-response} shows the 0--25\% range near the ceiling. Each mixed set yields one FID; these replacements are not independent generation repeats and do not recalibrate the five-repeat limits.

A limit of 8.5 excludes both early Base five-repeat FID means. All six mixed sets with 15\% replacement have FID 9.091--9.281, although their mean foreground- and bounding-box-area ratios are closer to one than Base. Thus a single-cohort FID above 9.0 can accompany more accurate geometric extent. Among mixed sets below FID 9.0, 9.5, and 10.0, the largest mean bounding-box-area ratios are 1.003, 1.056, and 1.083, respectively: 9.5 admits some expansion beyond the real images, and 10.0 admits more. These tests illustrate the geometric consequences of the calibrated limits; checkpoint eligibility still uses five-repeat scores. Compatibility separately measures agreement with requested access relations.

\subsubsection{Design Gains under Tighter Quality Limits}
\label{app:stricter-quality}

To test how much design gain remains when native quality stays closer to Base, we retrospectively constrain validation FID to at most 1.10 times Base and Compatibility to at most Base across the nine evaluated checkpoints. The matched Base in Table~\ref{tab:base-reference-calibration} gives limits of 8.2212 and 2.6020; AreaGuard-4 at iteration 10 meets both. We compare Base, iteration 10, and the selected iteration 20 on their 4,508 jointly accepted held-out test conditions.

\par\addvspace{8pt}\noindent\begin{minipage}{\linewidth}
\centering\small
\captionof{table}{\textbf{Tighter quality limits retain most of the design gain.} Native-quality means use the same validation cohort; PIAD uses the same 4,508 jointly accepted held-out test conditions.}
\label{tab:stricter-quality}
\setlength{\tabcolsep}{5pt}
\begin{tabular}{@{}lrrr@{}}
\toprule
Evaluation & Base & Iteration 10 & Iteration 20 \\
\midrule
Validation FID $\downarrow$ & 7.4739 & 8.1762 & 9.3590 \\
Validation Compatibility $\downarrow$ & 2.6020 & 2.1838 & 2.5812 \\
Held-out PIAD $\downarrow$ & 0.4127 & 0.2951 & 0.2818 \\
\bottomrule
\end{tabular}
\end{minipage}\par\addvspace{8pt}

The earlier checkpoint retains 89.8\% of the selected checkpoint's PIAD reduction: \((0.4127-0.2951)/(0.4127-0.2818)\), calculated from unrounded values. Joint resampling of the three outputs within each of the five test blocks gives a 95\% bootstrap interval of [85.8\%, 94.0\%] over 5,000 replicates with the residential reference fixed. Most of the design gain therefore remains at an earlier checkpoint with native quality closer to Base. This supplements the original selection in Eq.~\eqref{eq:selection}; quality uses validation data and design gains use held-out conditions.

\FloatBarrier
\section{Spatial Organization Gains and Residential-Use Assessment}
\label{app:full-results}

We examine HD improvements under identical design requirements, their spatial interpretation across 200 pairs, and residential-use judgments on 50 pairs. The systematic case analysis and the three assessment groups provide distinct forms of evidence.

\subsection{Matched Comparisons and Design Interpretation}
\label{app:matched-design-results}

\subsubsection{Condition-Matched Sensitivity}
\label{app:condition-matched}

Base and post-trained HD can differ in which outputs pass the acceptance rules. To compare the same design requests, we retain the 4,588 of 5,000 conditions for which both outputs are accepted, then recompute IPD, PIAD, and PIA statistics on this shared subset (Table~\ref{tab:condition-matched}). IPD and PIAD both decrease, and the share of layouts in which Living leads approaches the reference. Living's\label{app:living-share} mean fraction of retained-atom area rises from 36.2\% to 39.9\% (matched RPLAN plans: 40.0\%).

\par\addvspace{8pt}\noindent\begin{minipage}{\linewidth}
\centering
\renewcommand{\arraystretch}{1.12}
\captionof{table}{Base and post-training results on the same 4,588 conditions, with both outputs accepted. Brackets give 25th and 75th percentiles; the reference column describes all 20,348 RPLAN plans.}
\label{tab:condition-matched}
\small
\begin{tabular}{lrrr}
\toprule
Statistic & Base & +\method{} & RPLAN reference \\
\midrule
median \pia{} [25th, 75th] & 0.2809 [0.1306, 0.3937] & 0.3968 [0.2869, 0.4916] & 0.6172 [0.4838, 0.7794] \\
\(\Pr[\mathrm{PIA}>10^{-6}]\) & 87.90\% & 98.10\% & 98.98\% \\
\ipd{} on joint subset & 0.1112 & 0.0812 & 0 \\
\piad{} on joint subset & 0.4123 & 0.2807 & 0 \\
\bottomrule
\end{tabular}
\end{minipage}\par\addvspace{8pt}

PIA increases for 3,231 of these 4,588 conditions (70.42\%), with median within-condition change \(\Delta\mathrm{PIA}\) of 0.1164. We estimate uncertainty by resampling each condition's two outputs together within its original test block. Across 10,000 bootstrap replicates, the median change has a 95\% interval of \([0.1082,0.1257]\); the improvement fraction has Wilson interval \([69.09\%,71.73\%]\).

For the overall \ipd{} reduction, bootstrap sampling includes all attempted conditions within each block. Each replicate recomputes acceptance separately for Base and post-training, followed by the four population medians. The resulting 95\% interval for the reduction is \([25.2\%,28.6\%]\). The same replicates give a 95\% interval of \([30.8\%,33.6\%]\) for the 32.2\% \piad{} reduction.

\subsubsection{Connectivity Sensitivity}
\label{app:connectivity-sensitivity}

We test whether improvement remains when disconnected output graphs are excluded. Table~\ref{tab:connectivity} first evaluates the connected outputs of each model; we then compare only conditions whose Base and post-training graphs are both connected.

\par\addvspace{8pt}\noindent\begin{minipage}{\linewidth}
\centering
\renewcommand{\arraystretch}{1.12}
\captionof{table}{Single-component sensitivity on the five-block benchmark.}
\label{tab:connectivity}
\small
\setlength{\tabcolsep}{4pt}
\begin{tabular}{lrrrrr}
\toprule
Method & \shortstack{Connected /\\accepted} & \shortstack{Connected /\\attempted} & \ipd{} & \piad{} & median \pia{} \\
\midrule
Base & 4,223/4,723 (89.41\%) & 4,223/5,000 (84.46\%) & 0.1124 & 0.4182 & 0.2742 \\
+\method{} & 4,311/4,825 (89.35\%) & 4,311/5,000 (86.22\%) & 0.0833 & 0.2906 & 0.3883 \\
\bottomrule
\end{tabular}
\end{minipage}\par\addvspace{8pt}

Among each model's connected outputs, \ipd{} falls by 25.96\%, \piad{} by 30.52\%, and median \pia{} rises by 41.61\% (Table~\ref{tab:connectivity}). The stricter comparison uses 3,914 conditions with connected outputs both before and after SSPT: PIA improves for 70.49\%, with median within-condition change \(+0.1119\), and \piad{} decreases from 0.4167 to 0.2906. Improvement in this already-connected subset shows that the gain includes changes in functional organization beyond repairing disconnected graphs.

\subsubsection{Functional Profiles}
\label{app:functional-profiles}

To identify which functional domains contribute to the IPD change, Table~\ref{tab:profiles} reports each coordinate of the RPLAN, Base, and post-training profiles.

\par\addvspace{8pt}\noindent\begin{minipage}{\linewidth}
\centering
\renewcommand{\arraystretch}{1.12}
\captionof{table}{Median relative-integration profile coordinates.}
\label{tab:profiles}
\small
\begin{tabular}{lrrrr}
\toprule
Source & Public & Private & Service & Outdoor \\
\midrule
RPLAN reference & 1.5548 & 0.8772 & 0.7860 & 0.7445 \\
HD Base & 1.3465 & 0.9565 & 0.9210 & 0.7664 \\
HD+\method{} & 1.4044 & 0.9068 & 0.8686 & 0.8064 \\
\bottomrule
\end{tabular}
\end{minipage}\par\addvspace{8pt}

Post-training moves Public, Private, and Service closer to the RPLAN reference. Outdoor moves farther away even though overall \ipd{} decreases. The profile therefore shows both the domains that improve and the remaining mismatch hidden by the aggregate score.

\subsubsection{Task-Matched Gap and Centrality Sensitivity}
\label{app:centrality-sensitivity}

To check whether the gap to real residences and the post-training gains persist under other centrality definitions, we rescore the same HD and RPLAN graphs (Table~\ref{tab:matched-centrality-gap}). Eq.~\eqref{eq:metrics} replaces \(I_v\) and its averages with the chosen \(S_v\) and its room, category, and domain averages, for both generated outputs and the reference. Legacy HH remains the benchmark definition.

\par\addvspace{8pt}\noindent\begin{minipage}{\linewidth}
\centering
\renewcommand{\arraystretch}{1.12}
\captionof{table}{Centrality sensitivity of the task-matched functional gap and \method{} gains.}
\label{tab:matched-centrality-gap}
\small
\setlength{\tabcolsep}{3.5pt}
\begin{tabular}{lrrrrrrr}
\toprule
 & & \multicolumn{2}{c}{\ipd{} $\downarrow$} & \multicolumn{2}{c}{\piad{} $\downarrow$} & \multicolumn{2}{c}{median \pia{}} \\
\cmidrule(lr){3-4}\cmidrule(lr){5-6}\cmidrule(lr){7-8}
Centrality & \shortstack{Public-margin\\deficit} & Base & +\method{} & Base & +\method{} & Base & +\method{} \\
\midrule
Legacy HH & 46.27\% & 0.1111 & 0.0811 & 0.4129 & 0.2798 & 0.2800 & 0.3974 \\
Bounded HH & 38.68\% & 0.0368 & 0.0230 & 0.1180 & 0.0648 & 0.1060 & 0.1573 \\
Bounded HH, \(n>2\) & 38.54\% & 0.0376 & 0.0223 & 0.1130 & 0.0578 & 0.1060 & 0.1576 \\
Global harmonic & 42.36\% & 0.0503 & 0.0178 & 0.1456 & 0.0607 & 0.1186 & 0.2022 \\
\bottomrule
\end{tabular}
\end{minipage}\par\addvspace{8pt}

All four settings start from the same exact-eight sets: 20,348 RPLAN plans, 4,723 HD Base outputs, and 4,825 +\method{} outputs. The Public-margin deficit is the difference between the RPLAN and Base medians of \(\Delta_{\mathrm{pub}}\), divided by the RPLAN median. Excluding small components leaves 20,338 RPLAN plans with the scores needed for this comparison. In every setting, the Base Public margin remains below the residential reference, and post-training lowers IPD and PIAD while median PIA increases. PIAD compares each setting's outputs with RPLAN scored by the same centrality (20,336 plans have defined PIA when small components are excluded). These directions therefore persist beyond the Legacy HH score used for learning.

\subsubsection{Spatial Reorganization across 200 Paired Layouts}
\label{app:systematic-case-analysis}

\paragraph{Purpose and findings.}
The ten illustrations locate particular changes in spatial organization; the 200-pair analysis examines whether these changes recur across a broader range of PIA outcomes. Among 144 PIA-increasing pairs, 131 retain the same number of recovered room-level access links or fewer. The two dominant themes concern room access and grouping (96 pairs) and the living-space core and branches (71 pairs). Together, these findings characterize the gains primarily as changes in spatial organization rather than simply more links. Both themes also occur in PIA-decreasing pairs, so their effect must be interpreted through the relative roles of Living and the other functions.

\paragraph{Sample.}
The 200 pairs come from the same 4,588 common-acceptance conditions. Five test blocks and within-block PIA-change quintiles form 25 strata of eight pairs, including the ten original illustrations. The sample contains 144 increases and 56 decreases. Counts describe this stratified case set; the full-condition comparison in Section~\ref{sec:quality-constrained-results} provides the aggregate improvement estimate.

\paragraph{Organization of the observations.}
Each pair is assigned one primary descriptive theme from its input relations, room and door geometry, integration heatmap, and before/after scores (Table~\ref{tab:case-themes}). The themes locate where organization changes: room-access grouping describes which functions share an entry or are reached through public space; core-and-branch changes locate integration within Living's geometry. Component changes are compared with the requested grouping, and door details locate boundary-sensitive recovered relations.

\begin{table}[!htbp]
\centering\small
\caption{Primary descriptive themes in the 200-pair analysis. Each pair occurs once; PIA direction is cross-tabulated after thematic coding.}
\label{tab:case-themes}
\setlength{\tabcolsep}{6pt}
\begin{tabular}{@{}lrrr@{}}
\toprule
Primary theme & Pairs & PIA increase & PIA decrease\\
\midrule
Room access and room grouping & 96 & 71 & 25\\
Living-space core and branches & 71 & 51 & 20\\
Graph components and input relations & 20 & 15 & 5\\
Door pixels and recovered relations & 13 & 7 & 6\\
\midrule
Total & 200 & 144 & 56\\
\bottomrule
\end{tabular}
\end{table}

\paragraph{Case records.}
Each record retains condition and review identifiers, before/after scores, observed changes, the primary theme, and supporting locations. The separate native Compatibility audit additionally requires cached vectors and condition files and covers 195 pairs. Its five unavailable records do not reduce this 200-pair visual analysis. Residential-use judgments are assessed separately in Appendix~\ref{app:human-residential-use}.

\subsection{Residential-Use Assessment and Native-Quality Interpretation}
\label{app:human-native-use}

\subsubsection{Human Assessment of Residential Use}
\label{app:human-residential-use}
\paragraph{Purpose and sample.}
We assess whether the observed spatial changes correspond to judgments of residential-use suitability beyond SSIO scores. We use 50 condition-matched HD Base/SSPT pairs, selecting two pairs from each of five test blocks by five within-block strata of paired PIA change. The sample includes PIA increases and decreases. Counts describe this stratified sample rather than a uniformly sampled population preference.

\begin{figure}[!htbp]
\centering
\includegraphics[width=\linewidth]{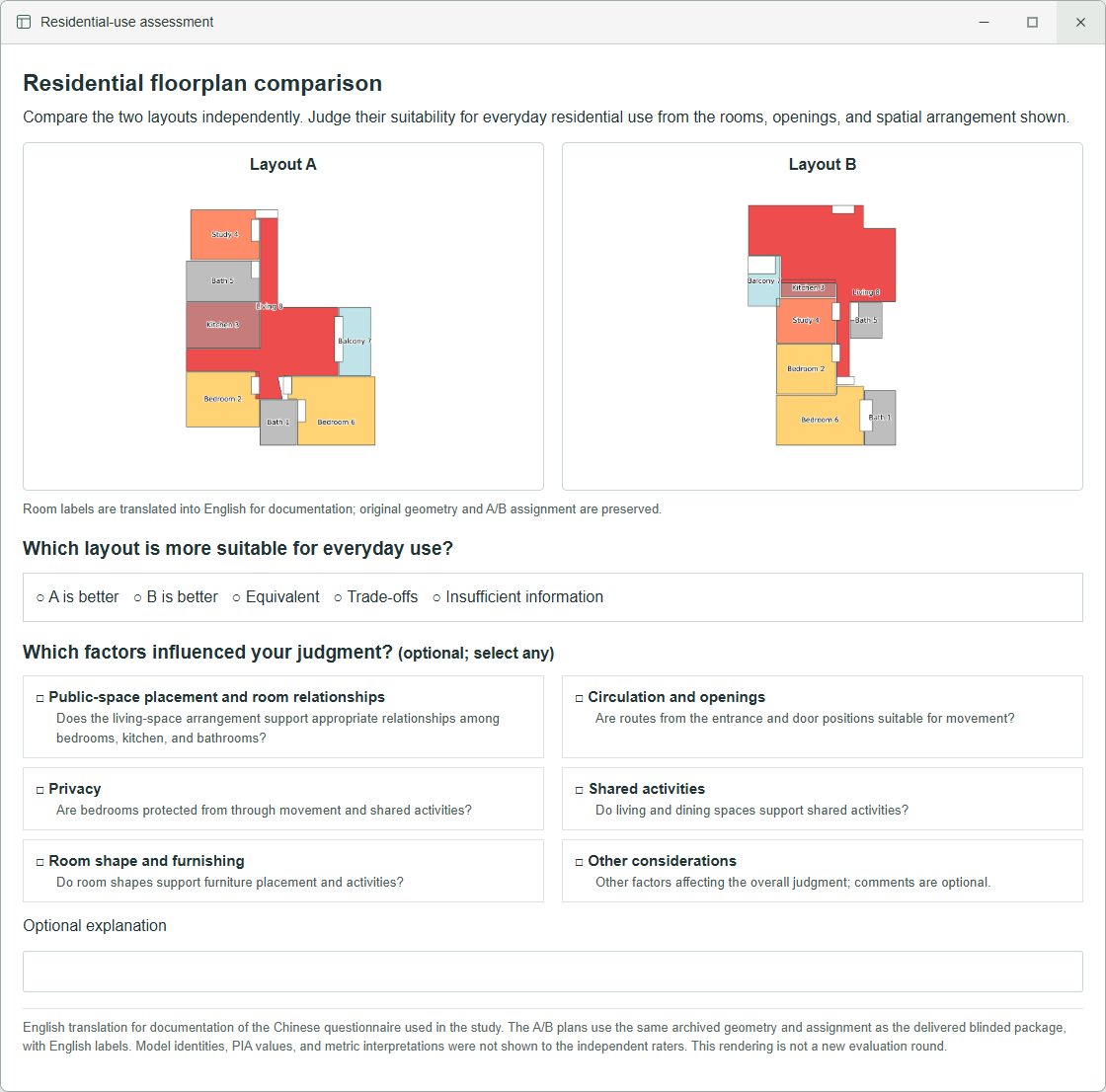}
\caption{English documentation rendering of the Chinese residential-use questionnaire. Room labels are translated; archived geometry and blinded A/B assignment are preserved. Independent raters compared A/B plans, selected an overall judgment, and could mark reasons. Model identities and SSIO scores were hidden. The figure translates the instrument used; it does not represent a new evaluation.}
\label{fig:human-review-interface}
\par\medskip
\begin{minipage}{\linewidth}
\centering\small
\captionof{table}{Residential-use assessments of the same 50 pairs. Preference and agreement with higher PIA are distinct outcomes. Percentages use definite preferences; other judgments remain in the complete counts. The discussion review is author-involved.}
\label{tab:human-residential-use}
\setlength{\tabcolsep}{3pt}
\begin{tabular}{@{}lrrrrrc@{}}
\toprule
Assessment & SSPT & Base & Equal & Trade-off & Insuff. & SSPT preference\\
\midrule
Real-estate designer (blind) & 34 & 7 & 2 & 5 & 2 & 83\% (34/41)\\
Architecture student (blind) & 28 & 14 & 2 & 1 & 5 & 67\% (28/42)\\
Architecture researchers (discussion) & 29 & 10 & 2 & 6 & 3 & 74\% (29/39)\\
\bottomrule
\end{tabular}
\par\smallskip
\begin{tabular}{@{}lcc@{}}
\toprule
Assessment & Prefer higher PIA & Base higher when preferred\\
\midrule
Real-estate designer & 73\% (30/41) & 43\% (3/7)\\
Architecture student & 67\% (28/42) & 43\% (6/14)\\
Architecture researchers & 72\% (28/39) & 40\% (4/10)\\
\bottomrule
\end{tabular}
\end{minipage}
\end{figure}

\Needspace{3\baselineskip}
\paragraph{Question and procedure.}
A real-estate designer and an architecture student, neither involved in the project, independently answered: ``Which layout is more suitable for everyday use?'' Model identities, scores, and metric explanations were hidden. Available judgments were A better, B better, equivalent, trade-offs, and insufficient information. Optional reasons covered public-space placement and relationships to other rooms, circulation/openings, privacy, shared activities, furnishing, and other considerations. These were considerations for an overall judgment, not separately scored dimensions. Figure~\ref{fig:human-review-interface} gives an English translation of the Chinese form actually delivered.

A separate author--architectural-postdoc discussion review used the same pairs and consulted metric/graph explanations on 17 cases. It provides descriptive judgments and qualitative interpretation and is not pooled with the independent assessments. We use final explicitly saved judgments and retain initial records and revisions in the evaluation archive.

\paragraph{Reading the evidence.}
Table~\ref{tab:human-residential-use} distinguishes method preference from agreement with PIA. The designer's 30 higher-PIA choices include 27 SSPT and 3 Base choices, whereas 34 choices favor SSPT. Overall preference rates use definite Base/SSPT choices; the last column uses only Base-preferred cases. Both independent assessments favor SSPT and higher-PIA layouts in most definite choices, connecting this spatial attribute to perceived everyday-use suitability.

The discussion review supplies explanations: in case R083, its recorded rationale favors the separation of kitchen and bedrooms through the public-space arrangement while keeping the sole bathroom closer to bedrooms. These are judgments from the layouts rather than observations of actual inhabitation.

\paragraph{Examples of recorded reasons.}
In R097, the designer favored SSPT's kitchen and bathroom positions relative to bedrooms, attributing this to the living-room configuration. In R199, the student favored its spatial variety for different furnishing needs. These paraphrases illustrate recorded reasons, rather than separately scored dimensions.

\subsubsection{Compatibility and Residential-Use Judgments}
\label{app:compatibility-use}

We cross-tabulate the existing residential-use judgments with native Compatibility, which counts missing and extra relations relative to the requested graph, including its exterior relation. We recompute these counts from the archived polygons for all 50 assessed pairs, using the unchanged native rule. Compatibility increases for 22 pairs. Table~\ref{tab:compatibility-use} reports every existing judgment for this subset; no pair is selected by its preference outcome. The two independent assessors still prefer SSPT in most definite choices. These results describe the stratified human-study sample, separately from the fixed native-quality cohort used for final testing.

\begin{table}[!htbp]
\centering\small
\caption{Existing judgments on all 22 assessed pairs whose native Compatibility increases. Percentages use definite Base/SSPT preferences; other judgments are retained.}
\label{tab:compatibility-use}
\setlength{\tabcolsep}{5pt}
\begin{tabular}{@{}lccc@{}}
\toprule
Assessment & Prefer SSPT & Prefer Base & Other judgments\\
\midrule
Real-estate designer (blind) & 83\% (15/18) & 17\% (3/18) & 4/22\\
Architecture student (blind) & 71\% (12/17) & 29\% (5/17) & 5/22\\
Architecture researchers (discussion) & 76\% (13/17) & 24\% (4/17) & 5/22\\
\bottomrule
\end{tabular}
\end{table}

\paragraph{FID is interpreted at collection level.}
Appendix~\ref{app:quality-range-diagnostics} retains the fixed-reference collection experiment and adds matched source-image examples (Figure~\ref{fig:base-range-response}). At 15\% replacement, all six collections have higher FID but mean foreground- and bounding-box-area ratios closer to the real reference. This concerns rendered extent under a controlled change in collection composition, rather than a per-image FID or an isolated causal effect of extent alone.

\subsubsection{Visual Interpretation of Native-Quality Changes}
\label{app:native-quality-visuals}

\begin{figure}[!htbp]
\paragraph{Local boundary sensitivity.}
For R037, R038, and R100, the displayed door is separated from one required room by one unit on the native 256-grid (Figure~\ref{fig:rb-boundary-sensitivity}). The native evaluator requires positive door--room overlap, so such a local change can remove a recovered link and raise Compatibility. The full layouts provide context for the magnified window; no polygons are shifted or repaired for display. This explains a specific count response, rather than assigning architectural acceptability from a grid-unit distance.
\smallskip
\centering
\includegraphics[width=0.94\linewidth]{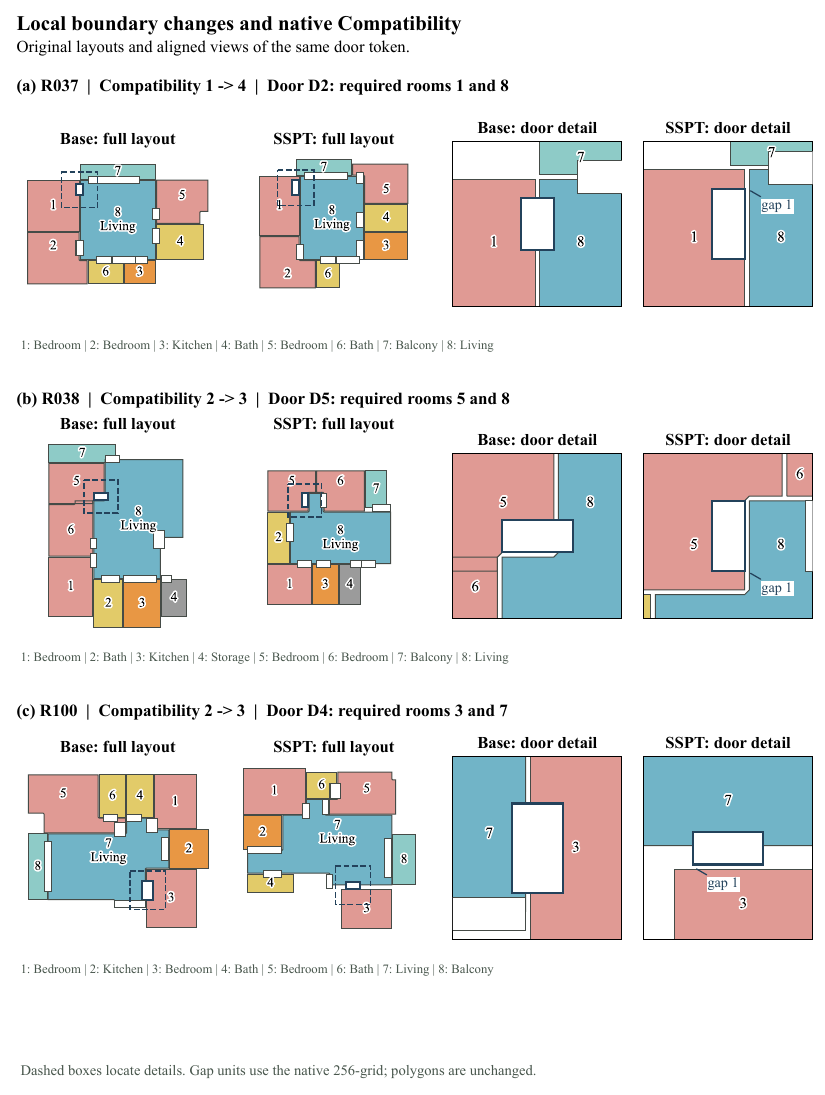}
\caption{Complete layouts and aligned door-detail views for three condition-matched pairs. Dashed boxes locate the details. One-unit door/room gaps prevent required links from being recovered by the native overlap rule and contribute to higher Compatibility counts. Full-layout scales match within each pair, as do local scales. Units refer to the native 256-grid; exterior contributions are unchanged within pairs.}
\label{fig:rb-boundary-sensitivity}
\end{figure}

\begin{figure}[!htbp]
\paragraph{Recognized benefits despite higher counts.}
For R083, R113, and R142, both independent assessors prefer SSPT although native Compatibility increases (Figure~\ref{fig:rb-use-preference}). Recorded discussion explanations concern the organization of kitchen, bedrooms and bathroom through public space; a more regular overall layout; and the spatial arrangement of other rooms. R142 also removes the Base kitchen/living overlap. Thus the count and the broader residential-use judgment can favor different outputs. The source images and graph requirements remain visible, while independent preferences and discussion interpretations are identified separately.
\smallskip
\centering
\includegraphics[width=0.94\linewidth]{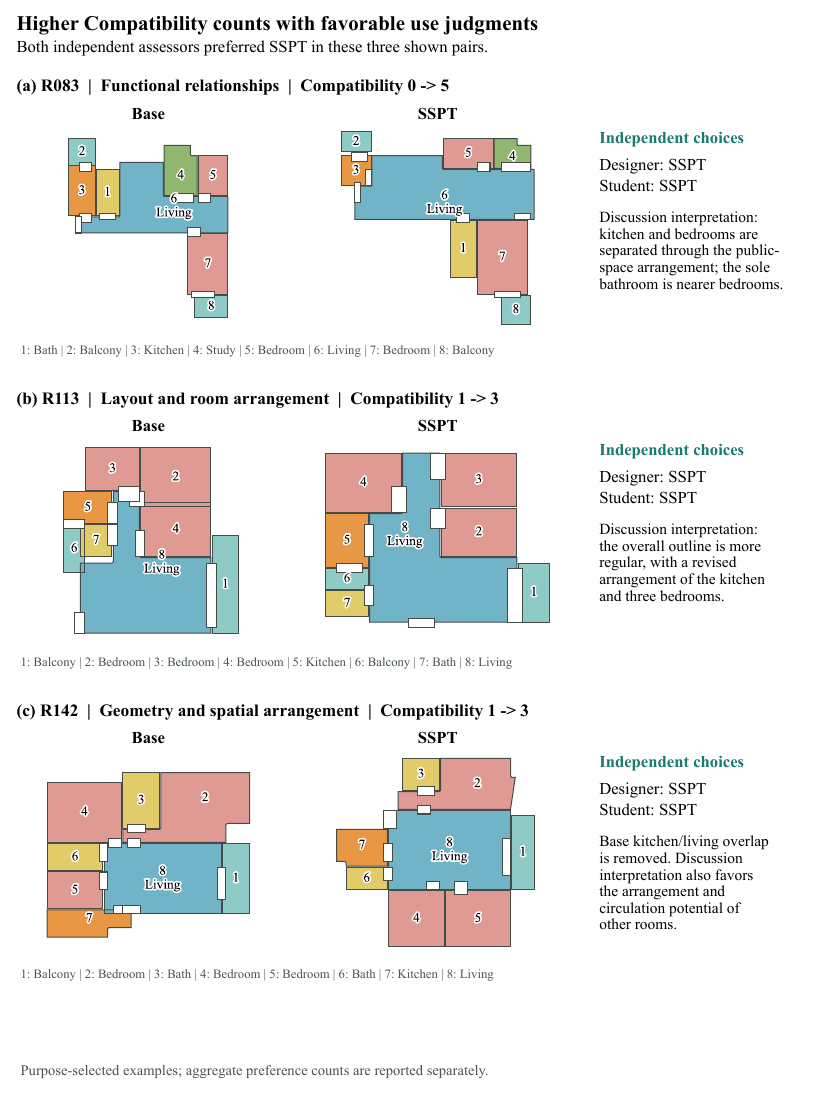}
\caption{Condition-matched examples with higher Compatibility counts that both independent blind assessors preferred for everyday residential use. Side notes distinguish independent choices from interpretations in the separate author/postdoc discussion. These selected examples illustrate functional arrangement, overall layout, and a local overlap change. Frequency results are reported for the full 50-pair study and its 22-pair higher-Compatibility subset.}
\label{fig:rb-use-preference}
\end{figure}

\paragraph{Source and selection.}
Of the 200 pairs stratified by test block and PIA change, 195 have the cached polygons and condition records needed for the native-rule diagnostic. Figures~\ref{fig:rb-boundary-sensitivity} and~\ref{fig:rb-use-preference} show purpose-selected examples with complete layouts and fixed room numbering: three boundary-sensitive pairs whose native interior reference matches the requested relations, and three pairs favored by both independent assessors. Preference frequencies use the full counts above.

\FloatBarrier

\Needspace{8\baselineskip}
\subsubsection{Case-Level Human-Assessment Records}
\label{app:human-case-records}
Table~\ref{tab:human-case-records} retains every final saved judgment. Pair numbers follow the blinded questionnaire; R identifiers link to the case-analysis records. S/B denotes a definite preference for SSPT/Base, = denotes equivalent, T denotes trade-offs, and ? denotes insufficient information. The first two columns of judgments are independent blind assessments; the researchers' column records the separate author-involved discussion. The Compatibility flag identifies all 22 pairs used in Table~\ref{tab:compatibility-use}.

\begingroup\footnotesize
\renewcommand{\arraystretch}{0.94}
\setlength{\tabcolsep}{5pt}
\begin{longtable}{@{}lrc ccc@{}}
\caption{Complete final judgments on the 50 assessed pairs. \(\Delta\mathrm{PIA}\) is SSPT minus Base.}\label{tab:human-case-records}\\
\toprule
Pair/R ID & \(\Delta\mathrm{PIA}\) & \shortstack{Compatibility\\increases} & \shortstack{Real-estate\\designer} & \shortstack{Architecture\\student} & \shortstack{Architecture\\researchers}\\
\midrule
\endfirsthead
\multicolumn{6}{l}{\tablename\ \thetable\ continued}\\
\toprule
Pair/R ID & \(\Delta\mathrm{PIA}\) & \shortstack{Compatibility\\increases} & \shortstack{Real-estate\\designer} & \shortstack{Architecture\\student} & \shortstack{Architecture\\researchers}\\
\midrule
\endhead
\midrule\multicolumn{6}{r}{Continued on the next page}\\\endfoot
\bottomrule\endlastfoot
01/R137 & +0.418 & Yes & S & S & S \\
02/R031 & +0.119 & Yes & S & S & S \\
03/R103 & +0.427 & Yes & ? & ? & S \\
04/R225 & +0.356 & Yes & S & S & S \\
05/R097 & +0.075 & Yes & S & = & S \\
06/R055 & -0.141 & No & S & ? & ? \\
07/R052 & +0.117 & No & S & S & S \\
08/R017 & -0.131 & Yes & T & B & B \\
09/R021 & +0.309 & Yes & S & B & S \\
10/R150 & +0.164 & No & S & S & S \\
11/R051 & -0.077 & Yes & B & B & = \\
12/R053 & +0.109 & Yes & ? & B & B \\
13/R138 & -0.188 & No & S & S & S \\
14/R199 & +0.168 & Yes & S & S & T \\
15/R090 & +0.203 & Yes & S & ? & S \\
16/R108 & +0.176 & Yes & S & S & S \\
17/R130 & -0.008 & No & S & B & S \\
18/R189 & -0.077 & No & T & B & B \\
19/R083 & +0.195 & Yes & S & S & S \\
20/R059 & -0.095 & No & B & B & T \\
21/R096 & +0.516 & No & S & S & S \\
22/R167 & -0.044 & No & S & S & S \\
23/R142 & +0.205 & Yes & S & S & S \\
24/R098 & +0.005 & Yes & S & = & = \\
25/R113 & +0.412 & Yes & S & S & S \\
26/R074 & +0.562 & No & S & S & S \\
27/R120 & +0.059 & No & S & T & S \\
28/R194 & +1.078 & No & S & S & S \\
29/R095 & +0.010 & No & B & B & T \\
30/R099 & +0.197 & No & B & S & B \\
31/R058 & +0.050 & Yes & B & B & S \\
32/R206 & -0.558 & Yes & S & S & B \\
33/R123 & +0.649 & No & S & S & B \\
34/R151 & -0.007 & Yes & T & S & S \\
35/R048 & -0.250 & No & S & S & S \\
36/R191 & +0.109 & Yes & B & S & B \\
37/R224 & +0.254 & Yes & S & S & T \\
38/R149 & +0.067 & No & S & B & S \\
39/R132 & -0.339 & No & = & ? & ? \\
40/R072 & -0.038 & No & B & B & T \\
41/R143 & +0.314 & No & S & B & S \\
42/R159 & +0.341 & No & S & S & S \\
43/R144 & -0.046 & Yes & S & ? & ? \\
44/R104 & +0.270 & No & S & S & T \\
45/R060 & +0.779 & No & S & B & S \\
46/R084 & +0.325 & No & S & S & S \\
47/R170 & +0.240 & No & T & S & B \\
48/R092 & +0.068 & No & = & S & S \\
49/R015 & -0.524 & No & T & S & B \\
50/R041 & +0.059 & No & S & B & B \\
\end{longtable}
\endgroup

\FloatBarrier
\section{Cross-Backbone Transfer and Access Completion}

\subsection{Cross-Backbone Results and Measurement Coverage}
\label{app:portability-results}

We evaluate Base and +\method{} for each backbone on the same 5,000 conditions, using the same measurement procedure before and after post-training. Table~\ref{tab:portable-full} reports design scores and accepted-output counts. HD supplies the detailed quality, ablation, and case-study results; the other four backbones test whether SSPT also improves their spatial organization.

\par\addvspace{8pt}\noindent\begin{minipage}{\linewidth}
\centering
\renewcommand{\arraystretch}{1.12}
\captionof{table}{Base and post-training results across five backbones, including Exact-8 acceptance.}
\label{tab:portable-full}
\small
\setlength{\tabcolsep}{5pt}
\begin{tabular}{llrrrr}
\toprule
Backbone & Variant & \ipd{} $\downarrow$ & \piad{} $\downarrow$ & median \pia{} &
Accepted \\
\midrule
HD & Base & 0.1111 & 0.4129 & 0.2800 & 4,723/5,000 \\
 & +\method{} & \textbf{0.0811} & \textbf{0.2798} & 0.3974 & 4,825/5,000 \\
\addlinespace
CE2EPlan & Base & 0.1914 & 0.5816 & 0.0726 & 3,324/5,000 \\
 & +\method{} & \textbf{0.1352} & \textbf{0.4460} & 0.2441 & 2,956/5,000 \\
\addlinespace
DiffPlanner & Base & 0.1803 & 0.5548 & 0.1113 & 2,956/5,000 \\
 & +\method{} & \textbf{0.1555} & \textbf{0.5043} & 0.1721 & 3,227/5,000 \\
\addlinespace
ChatHD & Base & 0.0866 & 0.3517 & 0.3295 & 1,561/5,000 \\
 & +\method{} & \textbf{0.0838} & \textbf{0.3446} & 0.3336 & 1,556/5,000 \\
\addlinespace
GSDiff--SSIO & Base & 0.0386 & 0.0903 & 0.5766 & 3,186/5,000 \\
 & +\method{} & \textbf{0.0360} & \textbf{0.0769} & 0.5934 & 3,304/5,000 \\
\bottomrule
\end{tabular}
\end{minipage}\par\addvspace{8pt}

Both design metrics improve for all five backbones. Acceptance increases for HD, DiffPlanner, and GSDiff--SSIO, decreases for CE2EPlan, and changes little for ChatHD. The size of the gains and the number of measurable outputs therefore depend on the backbone. Equation~\eqref{eq:reference-gap-closure} expresses these gains relative to each model's initial gap. We next examine GSDiff--SSIO, where learned access completion also influences the scores.

\subsection{Reference-Gap Summary}

Figure~\ref{fig:results-overview}(b) expresses each backbone's improvement as the fraction of its Base distance to the reference that post-training closes. B and S denote Base and +\method{}; both distances are zero for the reference itself:

\begin{equation}
g_{\mathrm{IPD}}=\frac{\mathrm{IPD}_{B}-\mathrm{IPD}_{S}}{\mathrm{IPD}_{B}},
\qquad
g_{\mathrm{PIAD}}=\frac{\mathrm{PIAD}_{B}-\mathrm{PIAD}_{S}}
{\mathrm{PIAD}_{B}}.
\label{eq:reference-gap-closure}
\end{equation}

\subsection{Completing Design Relations: The GSDiff Case}
\label{app:gsdiff-completion}

We evaluate GSDiff--SSIO in three experiments: access-completion accuracy on 500 held-out real RPLAN plans; the effects of completion inputs and decoding on 1,000 conditions, of which 538 are accepted in all eight model--completion combinations; and final post-training performance on 5,000 conditions, of which 2,462 are accepted both before and after training.

\subsubsection{From Native Structure to Access Relations}

GSDiff~\citep{hu2025gsdiff} generates corner and structural-edge graphs learned from RPLAN. Its preprocessing merges wall and door pixels into one mask, so recovered room cycles specify partitions and bubble-graph edges specify shared boundaries, but neither determines whether a boundary is passable. This differs from HD's generated door polygons and ChatHouseDiffusion's door-semantic pixels. CE2EPlan and DiffPlanner infer access from geometry (Appendix~\ref{app:backbone-bindings}); GSDiff--SSIO instead learns to predict the missing access relations.

During normalization \(\mathcal A_m\), a frozen predictor \(\mathcal C_\phi\) scores feasible room pairs from geometry, room types, and local context. The deployed pipeline decomposes room regions once and connects the resulting atoms through geometry-anchored openings for selected pairs. The predictor uses 2,000 RPLAN plans for training, 500 for calibration, and a disjoint 500 for fidelity evaluation. It learns local access labels; \ssio{} computes integration and the terminal reward from the completed graph. The same predictor \(\mathcal C_\phi\) is frozen for Base and +\method{}. SSPT updates only GSDiff's main diffusion model, keeping its topology encoder and terminal edge stage fixed.

\subsubsection{Access Recovery and Measurement Fidelity}

The raster-reconstruction test redraws predicted doors into held-out real plans, then recomputes the convex decomposition and integration scores (Table~\ref{tab:completion-fidelity}). Candidate recall measures how many reference access pairs appear in the geometric candidate set before prediction; these reference pairs come from door components touching exactly two rooms. An opening-realization issue occurs if a selected access pair is missing after rendering or an internal opening cannot be constructed.

Three of nine prespecified criteria are met. Access-pair recall is 0.9792 and precision is 0.8121; 97.2\% of plans have no opening-realization issues. Yet median absolute paired \pia{} error is 0.3611 and anchor-score correlation is 0.2726. Recovering most access pairs therefore does not ensure that integration is preserved. GSDiff--SSIO is reported separately because its absolute scores depend on learned access completion as well as generated geometry.

\par\addvspace{8pt}\noindent\begin{minipage}{\linewidth}
\centering
\captionof{table}{\textbf{Access-completion fidelity on 500 held-out plans.} Thresholds were fixed before evaluation. Pair recovery, opening realization, and preservation of integration are tested separately.}
\label{tab:completion-fidelity}
\small
\renewcommand{\arraystretch}{1.08}
\begin{tabular}{@{}lrrc@{}}
\toprule
Criterion & Required & Observed & Met \\
\midrule
\multicolumn{4}{l}{\emph{Opening realization}} \\

Extra pairs introduced by opening realization & $=0$ & 0 & Yes \\
Plans with opening-realization issues & $\leq0.020$ & 0.028 & No \\
\addlinespace[3pt]
\addlinespace[3pt]
\multicolumn{4}{l}{\emph{Candidate and access-pair recovery}} \\
Access-pair precision & $\geq0.900$ & 0.8121 & No \\
Access-pair recall & $\geq0.900$ & 0.9792 & Yes \\
Access-pair F1 & $\geq0.900$ & 0.8879 & No \\
Unambiguous-candidate recall & $\geq0.995$ & 1.000 & Yes \\
\addlinespace[3pt]
\addlinespace[3pt]
\multicolumn{4}{l}{\emph{Agreement of spatial measurements}} \\
Median absolute paired \pia{} error & $\leq0.050$ & 0.3611 & No \\
$R_{\mathrm{anchor}}$ Spearman correlation & $\geq0.900$ & 0.2726 & No \\
Room-integration Spearman correlation & $\geq0.900$ & 0.5405 & No \\
\bottomrule
\end{tabular}
\end{minipage}\par\addvspace{8pt}

\subsubsection{Separating Access Prediction from Raster Reconstruction}
\label{app:access-route-control}

A discrepancy in PIA can arise from predicting access or from reconstructing the raster used to measure it. We separate these effects on the same 500 held-out RPLAN plans, with the predictor frozen and RPLAN room categories fixed. One route redraws doors and reruns decomposition; the deployed route keeps decomposition fixed and inserts geometry-anchored access edges. Each uses annotated or predicted openings and compares PIA against the original plan.

\par\addvspace{8pt}\noindent\begin{minipage}{\linewidth}
\centering\small
\captionof{table}{\textbf{Access representation changes measurement fidelity.} Median absolute PIA error against the original plan on the same 500 held-out plans. Brackets give 95\% plan-bootstrap intervals.}
\label{tab:access-route-control}
\setlength{\tabcolsep}{5pt}
\begin{tabular}{@{}llr@{}}
\toprule
Access construction & Openings & PIA error [95\% interval] \\
\midrule
Redraw and decompose & Annotated & 0.3692 [0.3344, 0.4036] \\
Redraw and decompose & Predicted & 0.3611 [0.3268, 0.3940] \\
Fixed decomposition & Annotated & 0.0729 [0.0621, 0.0900] \\
Fixed decomposition & Predicted & 0.0724 [0.0646, 0.0829] \\
\bottomrule
\end{tabular}
\end{minipage}\par\addvspace{8pt}

The large error persists when the raster route uses annotated doors: access-pair accuracy alone cannot explain this measurement discrepancy. The fixed-decomposition route avoids that reconstruction step. Predicted versus annotated access within that fixed decomposition still gives median absolute PIA difference 0.0764, so relation recovery remains relevant. Including GSDiff's coarse room categories gives end-to-end error 0.0767. These checks use held-out plans with at most seven rooms; eight-room benchmark results retain their original measurements.

\subsubsection{Which Information Determines the Completed Scores?}

To determine how completion affects the measured organization, we hold generated geometry fixed and vary the room-type inputs and the rule used to select access edges. For a separate set of 1,000 conditions, we generate Base and +\method{} layouts once, then score each using four frozen completion configurations. Both models pass acceptance under all four configurations for 538 conditions; Table~\ref{tab:gsdiff-attribution} compares this common subset.

Each configuration uses the same trained room-pair predictor. Excluded features are replaced by their training means, making their standardized values zero. Geometry/context removes room-type and same-type features; Type removes the other predictor features. Both still use geometry to identify candidate pairs and place openings. The decoding rule then selects access edges: raw decoding keeps pairs above the predictor's threshold, while connected decoding also adds scored feasible pairs to join disconnected components where possible.

\par\addvspace{8pt}\noindent\begin{minipage}{\linewidth}
\centering
\renewcommand{\arraystretch}{1.12}
\captionof{table}{\textbf{Access-source attribution on 538 common conditions from a separate 1,000-condition experiment.} Feature restrictions apply to the frozen pair predictor; every route retains geometric candidate and opening construction.}
\label{tab:gsdiff-attribution}
\small
\setlength{\tabcolsep}{5pt}
\begin{tabular}{lrrrr}
\toprule
& \multicolumn{2}{c}{\ipd{} $\downarrow$} & \multicolumn{2}{c}{median \pia{}} \\
\cmidrule(lr){2-3}\cmidrule(lr){4-5}
Predictor features / decoding & Base & +\method{} & Base & +\method{} \\
\midrule
Full / connected & 0.0362 & 0.0351 & 0.5882 & 0.5984 \\
Full / raw & 0.0275 & 0.0302 & 0.6144 & 0.6299 \\
Geometry/context / connected & 0.1102 & 0.1087 & 0.2781 & 0.2877 \\
Type / connected & 0.0407 & 0.0414 & 0.6093 & 0.6165 \\
\bottomrule
\end{tabular}
\end{minipage}\par\addvspace{8pt}

Removing room-type features lowers Base median \pia{} from 0.5882 to 0.2781, whereas Type retains 0.6093. The high PIA assigned by completion thus depends strongly on room-type inputs, within the same geometric candidate structure. Raw decoding yields higher median \pia{} than connected decoding for both models, so connectivity repair alone does not explain the high scores. Post-training \ipd{} improves under Full/connected and Geometry/context/connected, but worsens under the other two configurations on this shared subset.

\subsubsection{Post-Training through the Fixed Completion Interface}

Final post-training evaluation uses five blocks totaling 5,000 conditions, separate from the 1,000-condition completion analysis above. With the full completion model fixed, \ipd{} decreases from 0.0386 to 0.0360, \piad{} from 0.0903 to 0.0769, median \pia{} increases from 0.5766 to 0.5934, and accepted outputs increase from 3,186 to 3,304 (Table~\ref{tab:portable-full}). Across the five blocks, Base \ipd{} is \(0.0386\pm0.0005\), block \piad{} is \(0.0916\pm0.0032\), and block median \pia{} is \(0.5771\pm0.0047\), with population standard deviations.

For the 2,462 conditions accepted both before and after post-training, \(\Delta\pia{}\) has mean \(+0.0167\) and median zero: 1,225 improve, 248 tie, and 989 worsen. Aggregate improvement therefore coexists with unchanged or worse individual layouts. Here the generator is updated while access completion stays fixed; the separate held-out test in Table~\ref{tab:completion-fidelity} measures the accuracy of that completion. Using a fixed feedback procedure and recovering real access relations accurately are distinct requirements.

\FloatBarrier
\Needspace{6\baselineskip}
\section{Generative AI for Design: Literature Review and Synthesis}
\label{app:survey-overview}
\label{app:literature-review}

\raggedbottom
To place the floorplan study in a broader design context, we review how 55 studies across 14 design categories use domain knowledge in generative modeling. Three questions organize the comparison:

\begin{enumerate}
\item \textbf{RQ1: What counts as design progress, and what evidence supports it?} We compare synthesis-oriented and purpose-oriented contributions across research communities (Section~\ref{app:survey-priorities}).
\item \textbf{RQ2: Where do design criteria come from, and how is their meaning made computable?} We trace knowledge sources to analytical objects and measurements (Section~\ref{app:survey-evaluation}).
\item \textbf{RQ3: How do design criteria change generation, and what makes them reusable learning objectives?} We distinguish assessment, inference, and model updates, then examine adaptation across generators (Section~\ref{app:survey-feedback}).
\end{enumerate}

The synthesis in Section~\ref{app:survey-synthesis} uses these comparisons to locate the floorplan study within existing work. Appendix~\ref{app:design-learning-principles} then proposes a workflow for applying design theory in other tasks.

\subsection{Review Scope and Evidence Coding}
\label{app:survey-coding}

\begin{figure}[!htb]
  \centering
  \includegraphics[width=\linewidth]{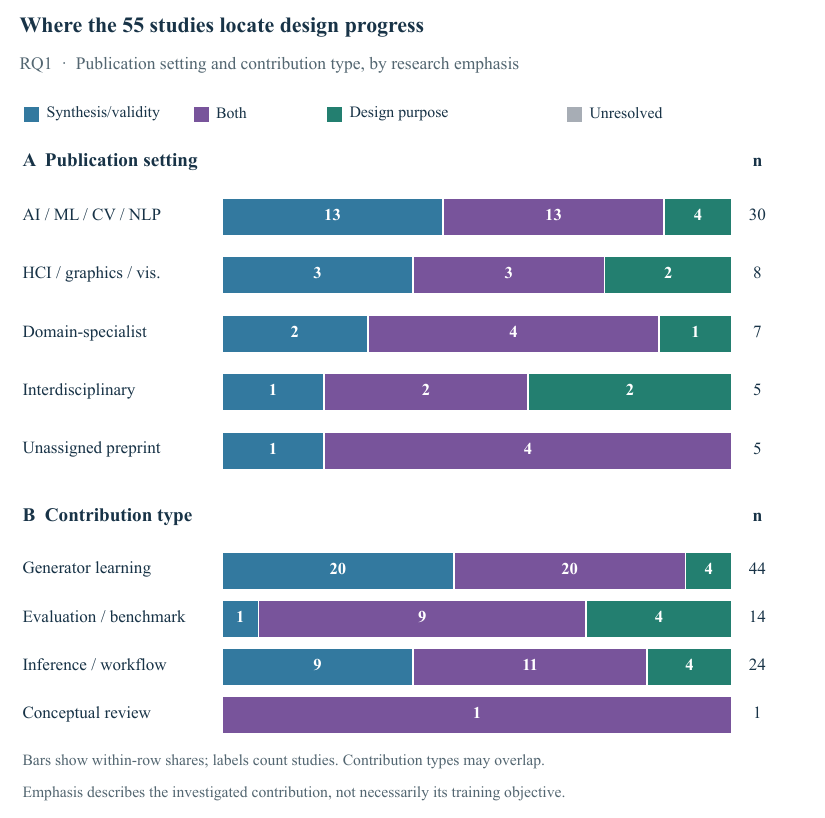}
  \caption{\textbf{Research emphasis across all 55 reviewed studies.} Publication setting and contribution type are classified independently of the investigated objective. A study may contribute to multiple types in panel B. Study-level assignments and references appear in Section~\ref{app:survey-inventory}; Figure~\ref{fig:design-survey-research-evidence} explains representative mechanisms.}
  \label{fig:design-survey-full-research}
\end{figure}

\paragraph{Selection and coverage.} We purposively sampled work published or released from 2021 to September 13, 2026, starting from recent systems and tracing representative synthesis, evaluation, control, and learning approaches. The full-sample summaries include all 55 studies in 14 design categories (Figures~\ref{fig:design-survey-full-research} and~\ref{fig:design-survey-full-knowledge}). Fifty have method-level evidence for detailed mapping; three abstract-level records and two contextual reviews or benchmarks are marked separately. Earlier design theories and general learning methods provide context outside this inventory. Counts describe the selected sample.

\paragraph{What we record.} Each record identifies the year, publication setting, design category, contribution type, represented objects, knowledge source, criterion use, updated component, and validation evidence. These fields distinguish generator development, evaluation, and human--AI workflows. Publication setting describes the research community; methods and experiments determine the recorded contribution. Section~\ref{app:survey-inventory} provides the coding key and complete inventory.

\paragraph{Computational roles and knowledge sources.} We code representation as implicit, explicit elements, or explicit relations. We separately record whether a criterion is used for example fitting, evaluation, inference, or learning, and which component is updated. Learning includes supervision, losses, and rewards. RQ2 identifies the criterion's source: domain theories, established principles, author-defined criteria, task specifications, physical models, or preference judgments. Representation and use are thus classified separately from theoretical grounding and design quality.

\paragraph{Full-sample classification.} Research emphasis records whether a study examines synthesis quality and condition validity, domain-purpose performance, or both. Criterion use records what the criterion changes: a study may evaluate design purpose while training its generator only to fit examples. We record each criterion's source and documented uses, including learned evaluators or surrogates. Example-based records fit data without a separately established design criterion; unknown sources remain unresolved. Expert evaluation is distinguished from expert-provided training labels. A study can use multiple knowledge sources and apply a criterion in several ways.

\paragraph{Evidence and interpretation.} We assign codes from reported representations, losses, rewards, updates, and evaluation protocols. Fixed conditioning, interpretation of underspecified requests, interactive revision, and preference learning are recorded separately; stated aims alone do not establish a computational or learning mechanism. Coverage is task-specific: editorial examples concern single pages, visualization examples mainly charts, and exhibition examples spatial layouts. Section~\ref{app:survey-inventory} lists every study, citation, and evidence status.

\subsection{RQ1: Design Progress and Its Evidence}
\label{app:survey-priorities}

\paragraph{Two standards of progress.} Generative modeling asks whether an artifact is faithful and controllable; design asks what it accomplishes. Visual and geometric quality make a proposal recognizable, coherent, or constructible. Design quality also concerns how its elements serve a purpose: communicating an argument, supporting movement, assembling a garment, or carrying a load. Statistical similarity and functional performance therefore require different evidence~\citep{regenwetter2023metrics}. We ask how these design purposes can guide generator learning.

\paragraph{Different research starting points.} A recurring contrast in the surveyed work is what defines progress. AI synthesis papers foreground representations, conditioning, and generation quality: LayoutTransformer models page layouts, DeepCAD construction sequences, and HouseDiffusion vector coordinates~\citep{gupta2021layouttransformer,wu2021deepcad,shabani2023housediffusion}. Domain-oriented studies foreground use, professional judgment, or physical performance~\citep{regenwetter2023metrics,park2024quality}. The former can advance generation without explicitly measuring the design purpose; the latter can articulate that purpose without making it a generator's learning objective.

\begin{figure}[!htb]
  \centering
  \includegraphics[width=\linewidth]{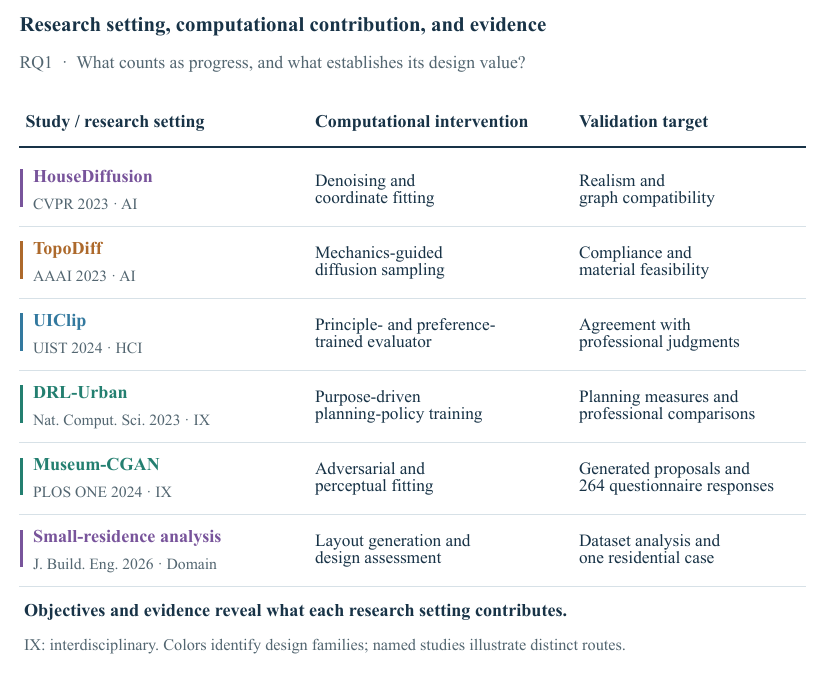}
  \caption{\textbf{Research setting, computational intervention, and validation.} Representative cases explain the 55-study distribution in Figure~\ref{fig:design-survey-full-research}; the small-residence analysis adds architectural context outside that coded sample. References and validation details appear in Section~\ref{app:survey-inventory} and Table~\ref{tab:survey-validation}.}
  \label{fig:design-survey-research-evidence}
\end{figure}

\paragraph{Model advances and design evidence.} Museum-CGAN illustrates the separation: a conventional conditional GAN learns adversarial and perceptual objectives, while questionnaires assess the resulting museum proposals~\citep{tang2024museumcgan}. TopoDiff, published at an AI venue, brings mechanical performance into diffusion sampling~\citep{maze2023topodiff}; interdisciplinary DRL-Urban directly trains a graph policy for planning goals~\citep{zheng2023urbanplanning}. Figure~\ref{fig:design-survey-research-evidence} makes these distinctions visible; Table~\ref{tab:survey-validation} separates data volume, design cases, and human assessments. These cases show why design evaluation, inference-time guidance, and generator learning must be recorded separately.

\begin{table}[!htbp]
\centering
\caption{Synthesis quality and design purpose ask different questions about the same artifact.}
\label{tab:survey-purpose-distinction}
\small
\setlength{\tabcolsep}{4pt}
\renewcommand{\arraystretch}{1.15}
\begin{tabular}{@{}>{\raggedright\arraybackslash}p{0.17\linewidth}>{\raggedright\arraybackslash}p{0.38\linewidth}>{\raggedright\arraybackslash}p{\dimexpr0.45\linewidth-16pt\relax}@{}}
\toprule
Design object & Synthesis-oriented question & Purpose-oriented question \\
\midrule
Graphic layout & Are elements and placements~plausible? & Does their organization establish a~readable~hierarchy? \\
Mechanical part & Is the geometry valid and similar~to~examples? & Does the structure perform under its intended~loads? \\
Garment & Can a coherent pattern be~generated? & Do panel and sewing relations support assembly and wearing? \\
Floorplan & Are rooms and boundaries geometrically~coherent? & Does their access organization support the intended use of spaces? \\
\bottomrule
\end{tabular}
\end{table}

\paragraph{Comparing purposes across tasks.} The following examples compare what each design organizes, the purpose of that organization, and whether it affects evaluation, inference, or learning.
\subsubsection{Organization for Communication and Interaction}
\label{app:survey-domains}

Graphic and information designs organize elements to direct attention, convey meaning, and support action. A coherent arrangement is the starting point; reading hierarchy, scientific fidelity, and~interaction introduce more specific standards of success.

\begingroup\interlinepenalty=10000
\paragraph{Posters and advertising.} PosterLLaVa learns placements from reference layouts; LaySPA post-trains a generator with explicit spacing, alignment, overlap, and text--underlay rewards; PosterReward adapts an image generator through learned judgments of rendered posters~\citep{yang2024posterllava,li2026layspa,lai2026posterreward}. These methods respectively learn from example layouts, explicit relation rewards, and learned quality judgments.
\par
\endgroup

\paragraph{Editorial and document layout.} LayoutTransformer models page-element sequences; LayoutDM adds relation energies during sampling, LayoutFormer++ constrains decoding, and LayoutNUWA generates editable code~\citep{gupta2021layouttransformer,inoue2023layoutdm,jiang2023layoutformerplusplus,tang2024layoutnuwa}. These mechanisms control placement and expose elements needed to study information hierarchy, but their reported objectives do not directly measure reading or argumentation.

\paragraph{Scientific diagrams.} AutomaTikZ and DiagramAgent improve graphical programs through compilation or checking; AutoFigure critiques and refines an entity--relation blueprint before rendering~\citep{belouadi2024automatikz,wei2025diagramagent,zhu2026autofigure}. The inspectable blueprint separates scientific organization from appearance, although correction chiefly occurs during inference.

\paragraph{Data visualization.} LIDA translates analytical goals into executable charts; ChartGPT produces data transformations and chart specifications; Data Formulator 2 exposes transformations and visual encodings for revision~\citep{dibia2023lida,tian2025chartgpt,wang2025dataformulator2}. Here organization serves a data question, and users revise analytical meaning together with visual form.

\paragraph{Interfaces and~interaction.} UIClip uses design principles and professional preferences to train a screenshot evaluator; UICoder fine-tunes a UI-code generator through compilation and prompt-relevance feedback~\citep{wu2024uiclip,wu2024uicoder}. UIClip trains an evaluator; UICoder updates a generator. Screenshot quality, code execution, and successful interaction require separate evaluation.

\paragraph{Packaging.} A kumis-packaging workflow combines cultural elements, shape grammar, requirement weighting, prompting, and proposal selection; a craft-beer study compares visual and brand expression across tools~\citep{zhao2026kumis,chen2026craftbeer}. Both center cultural communication and identity in evaluating proposals, without updating the underlying generator or addressing structural packaging engineering.

\subsubsection{Organization for Construction and Physical Performance}

For physical designs, organization must support assembly and behavior. Valid geometry establishes an object that can be manipulated or constructed; performance criteria determine whether that object is useful under its intended conditions.

\paragraph{Industrial products and CAD.} DeepCAD learns construction sequences, BrepGen boundary geometry and associations, and Text2CAD language-conditioned parametric construction~\citep{wu2021deepcad,xu2024brepgen,khan2024text2cad}. MO-PaDGAN instead trains with performance and diversity, while TopoDiff guides sampling through mechanical criteria~\citep{chen2021mopadgan,maze2023topodiff}. Because similarity does not establish engineering value~\citep{regenwetter2023metrics}, both the measured property and its effect on generation matter.

\paragraph{Garments.} DressCode, SewingLDM, and AIpparel represent panels and sewing relations---structure tied to assembly, not only appearance~\citep{he2024dresscode,liu2024sewingldm,nakayama2025aipparel}. Reference patterns teach structured generation, and SewingLDM adds integrity terms to its representation model. Simulation, reconstruction, and user studies then test different aspects of assembly and wearing.

\paragraph{Textiles and patterns.} Textile diffusion and Knit-FLUX target pattern appearance and knitted-fabric imagery, whereas functional-textile work represents interlacing and yarn fields under surrogate performance losses~\citep{karagoz2023textile,liu2025knitflux,tang2026functionaltextile}. This separates organizing a surface motif from organizing material for physical behavior; implementation and validation evidence are recorded in the inventory.

\subsubsection{Organization for Spatial and Sequential Experience}

Spatial and temporal designs organize routes through space or events. Accessibility, playability, continuity, and narrative transition supply purposes that visual consistency alone cannot establish.

\paragraph{Urban and landscape design.} UrbanGAN fits block morphology and evaluates spatial attributes; DRL-Urban trains policies for accessibility, ecology, and road efficiency~\citep{fedorova2021urbangan,zheng2023urbanplanning}. PlantoGraphy connects intentions, scene graphs, layouts, and rendering interactively, whereas GenFusion combines diffusion candidates with performance prediction and evolutionary search~\citep{huang2024plantography,gao2025genfusion}. Design purpose thus enters through policy learning, interaction, search, or assessment.

\paragraph{Film, storyboards, and animation.} TaleCrafter controls character layouts; CineVerse plans scenes and shots for consistent keyframes; ShotPlan learns shot-control tokens; and STAGE preference-optimizes frame pairs before a separate video model renders transitions~\citep{gong2023talecrafter,phung2026cineverse,guo2026shotplan,zhang2025stage}. These studies evaluate different targets: identity consistency, shot organization, or transitions. Their mechanisms also differ in whether they control a pretrained model or update its weights.

\paragraph{Game levels.} Controllable PCGRL trains from path, region, and playability objectives; later work scales this setting and adds language control~\citep{earle2021contpcgrl,earle2024pcgrljax,baek2025ipcgrl}. DiffusionCraft instead fits local patterns from one level and evaluates playability afterward~\citep{dai2024diffusioncraft}. Purpose-driven RL and example-fitting diffusion therefore coexist; the distinction is which design capability enters learning.

\paragraph{Exhibition spaces.} Museum-CGAN fits layouts adversarially and perceptually before human evaluation; Rule-Exhibit turns connectivity, salience, and artwork importance into topology and placement supervision; later work compares image and spatial properties across generators~\citep{tang2024museumcgan,xu2025ruleexhibit,li2026museum}. These are different interventions: judging proposals, constructing design-informed targets, and comparing systems. Rules can therefore enter learning without an RL reward.

\begin{figure}[p]
  \centering
  \includegraphics[width=\linewidth]{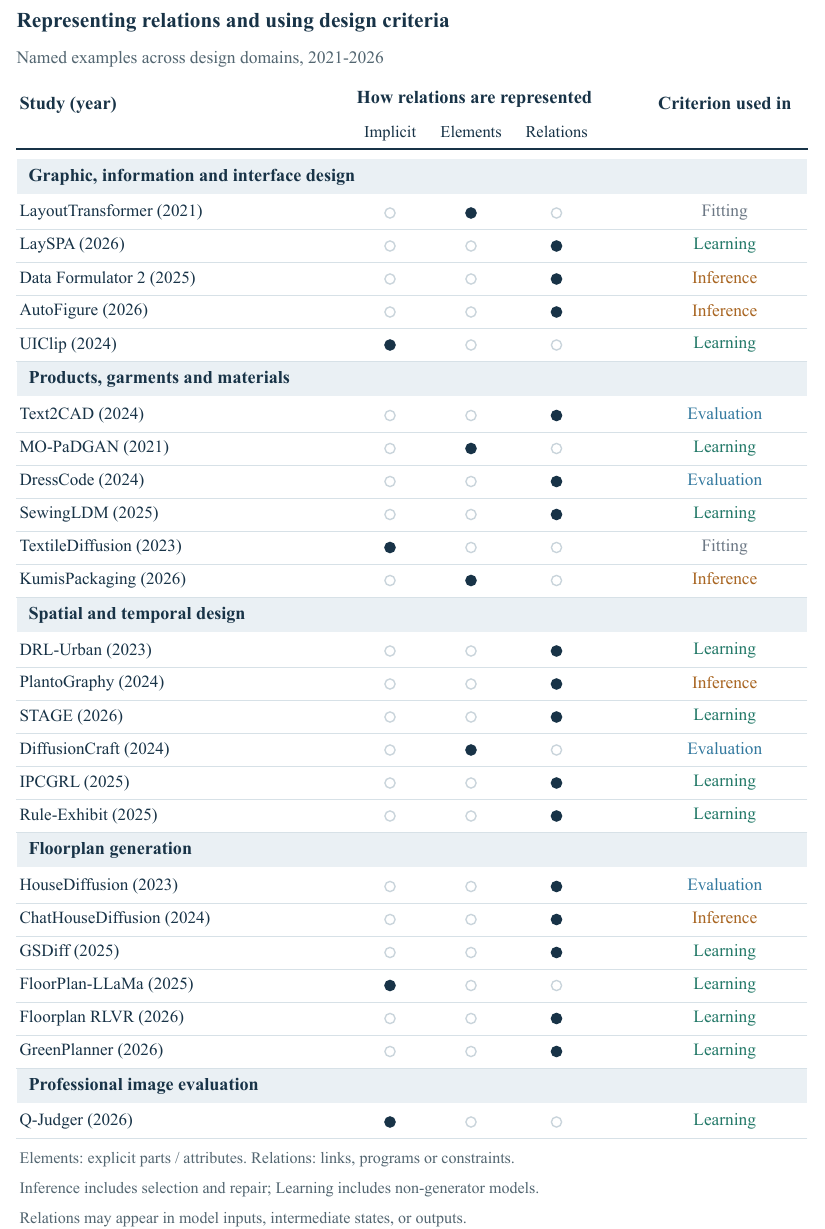}
  \caption{\textbf{Representation and criterion use are distinct.} Named studies connect represented organization to criterion use. Figure~\ref{fig:design-survey-knowledge-basis} distinguishes theory-derived measures from other criteria; Section~\ref{app:survey-inventory} gives the complete inventory.}
  \label{fig:design-survey-map}
\end{figure}

\paragraph{RQ1 finding: design interest and design learning must be distinguished.} Of the 55 records, 20 emphasize synthesis or validity, nine design purpose, and 26 examine both (Figure~\ref{fig:design-survey-full-research}). Both emphases occur across publication communities. Evaluating a design purpose does not necessarily make it a training objective: professional assessment can follow example fitting, and performance criteria can guide inference without updating model weights. Figure~\ref{fig:design-survey-map} therefore compares what each method represents and where its design criteria act.

\subsection{RQ2: From Knowledge Sources to Design Measurement}
\label{app:survey-evaluation}

\begingroup\interlinepenalty=10000
\paragraph{Evaluate the purpose of the particular design.} IDEA-Bench tests professional task completion; Qwen-Image-Bench and ArtiMuse train evaluators from professional rubrics or aesthetic annotations; PosterReward targets poster quality~\citep{liang2025ideabench,li2026qwenimagebench,cao2026artimuse,lai2026posterreward}. The relevant purposes differ: a poster organizes attention and communication, while a dwelling organizes access, shared activities, and privacy. Appearance and organization contribute according to the task; neither a universal aesthetic score nor geometric validity defines success across domains.
\par
\endgroup

\begin{figure}[!htb]
  \centering
  \includegraphics[width=\linewidth]{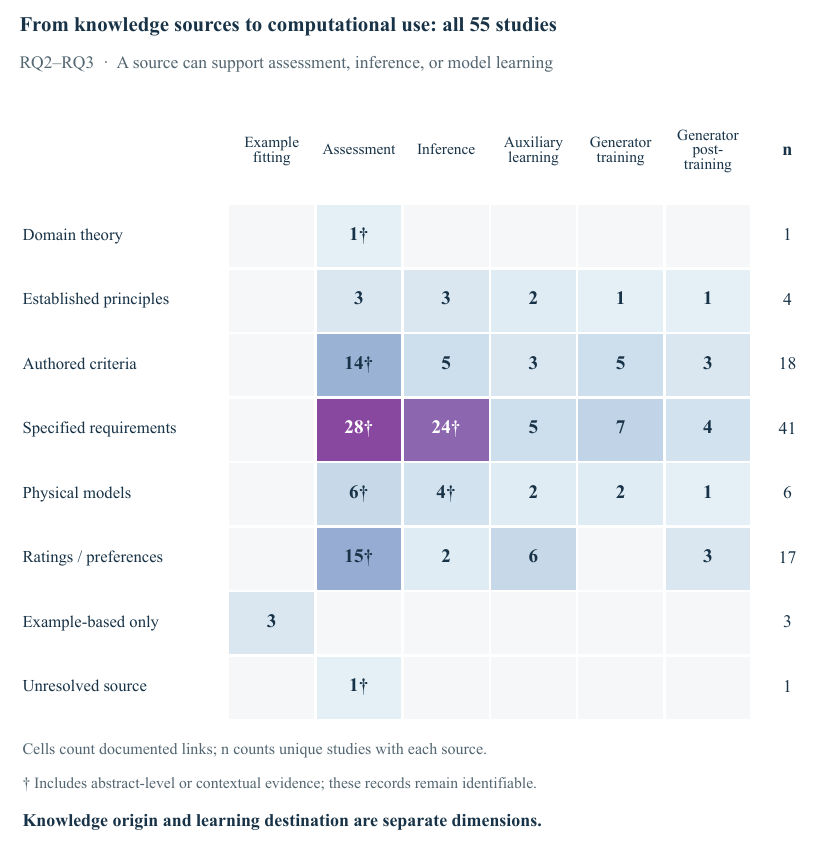}
  \caption{\textbf{Knowledge provenance and computational destination across all 55 studies.} Cells count documented source-to-use links, including learned implementations; both dimensions may overlap. Marked cells include abstract-level or contextual evidence. Our framework is excluded from these counts and appears separately in Figure~\ref{fig:design-survey-knowledge-basis}.}
  \label{fig:design-survey-full-knowledge}
\end{figure}

\begin{figure}[!htb]
  \centering
  \includegraphics[width=\linewidth]{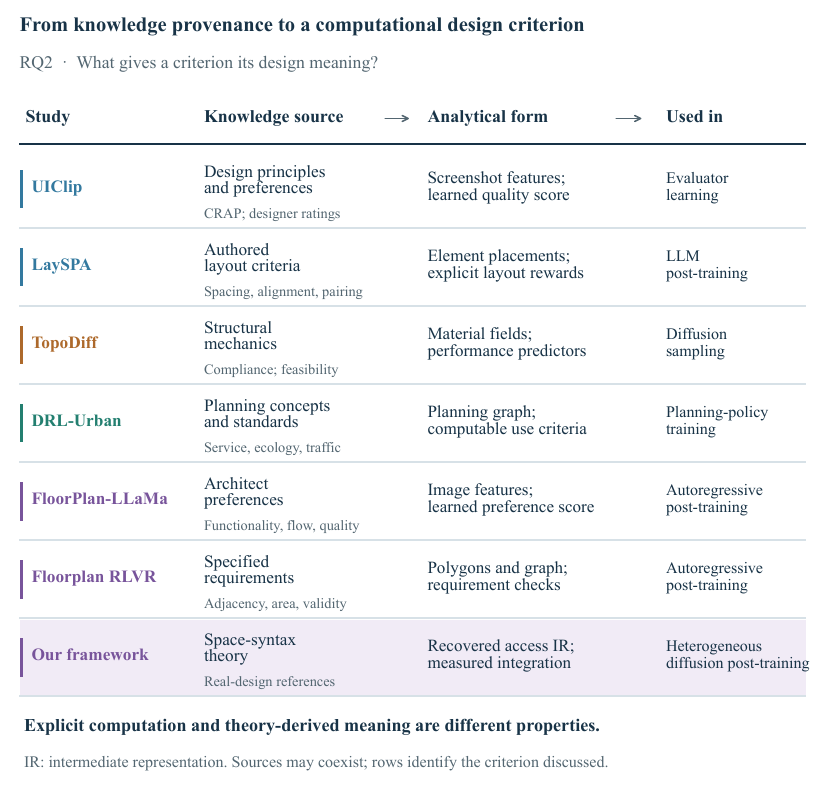}
  \caption{\textbf{Representative knowledge-to-computation connections.} These cases explain the categories in Figure~\ref{fig:design-survey-full-knowledge}. Our pathway carries space-syntax objects and measurements through empirical grounding into heterogeneous-generator post-training. Study references are given in the accompanying analysis and inventory.}
  \label{fig:design-survey-knowledge-basis}
\end{figure}

\begingroup\interlinepenalty=10000
\paragraph{Identify where the criterion comes from.} We call a measure \emph{theory-derived} when its analytical objects and interpreted property can be traced to an established domain theory. Other criteria may operationalize design principles, check task requirements, predict physical behavior, or learn judgments from preference labels. A system may combine these sources. Figure~\ref{fig:design-survey-full-knowledge} classifies all 55 records by knowledge source and use; Figure~\ref{fig:design-survey-knowledge-basis} explains representative examples.
\par
\endgroup

\Needspace{8\baselineskip}
\paragraph{Separate theoretical meaning from implementation.} UIClip uses contrast, repetition, alignment, and proximity (CRAP) principles to construct perturbations, then learns judgment from screenshots and preferences. LaySPA explicitly computes authored spacing, alignment, overlap, and text--underlay rewards~\citep{wu2024uiclip,li2026layspa}. TopoDiff uses structural mechanics; DRL-Urban operationalizes planning concepts and service standards, including the 15-minute life circle~\citep{maze2023topodiff,zheng2023urbanplanning}. Space syntax supplies both an analytical representation and an interpretation of integration before either is embedded in a learning system~\citep{hillier1984sociallogic,hillier1996space}.

\begingroup\interlinepenalty=10000
\paragraph{Architectural assessment and its remaining gap.} Floorplan evaluation covers image similarity, geometry, program constraints, and architect-informed rules~\citep{nauata2021houseganpp,shabani2023housediffusion,zeng2025floorplanmetrics,luo2024dstruct2design,vanengelenburg2025msd,park2024quality}. FloorPlan-LLaMa learns functionality, flow, and overall judgments from expert labels~\citep{yin2025floorplanllama}, but the spatial basis of each score remains implicit. The evaluator ranks perceived quality without exposing the access organization responsible for it.
\par
\endgroup

\paragraph{Space syntax connects organization to use.} Space syntax analyzes spatial configuration through explicit relations and measures~\citep{hillier1984sociallogic,hillier1996space,jiang2000integration}, grounded in architectural studies of domestic organization and design transformation~\citep{hanson1998decoding,edgu2003preferences,derix2014spatialconfiguration,eloy2016transforming,azizi2023latent}. \ssio{} recovers convex spaces and access relations, then measures relative accessibility. Theory defines the property; residential corpora provide the empirical reference needed to interpret it.

Residential studies connect accessibility to particular uses: living-room-centered configurations in Korean apartments~\citep{byun2016typology}, collective activities in integrated spaces in Inuit households~\citep{dawson2003inuit}, and private and shared living-room experiences~\citep{rechavi2009livingroom}. These findings motivate context-specific reference distributions in our cross-corpus analysis.

\begin{figure}[p]
  \centering
  \includegraphics[width=\linewidth]{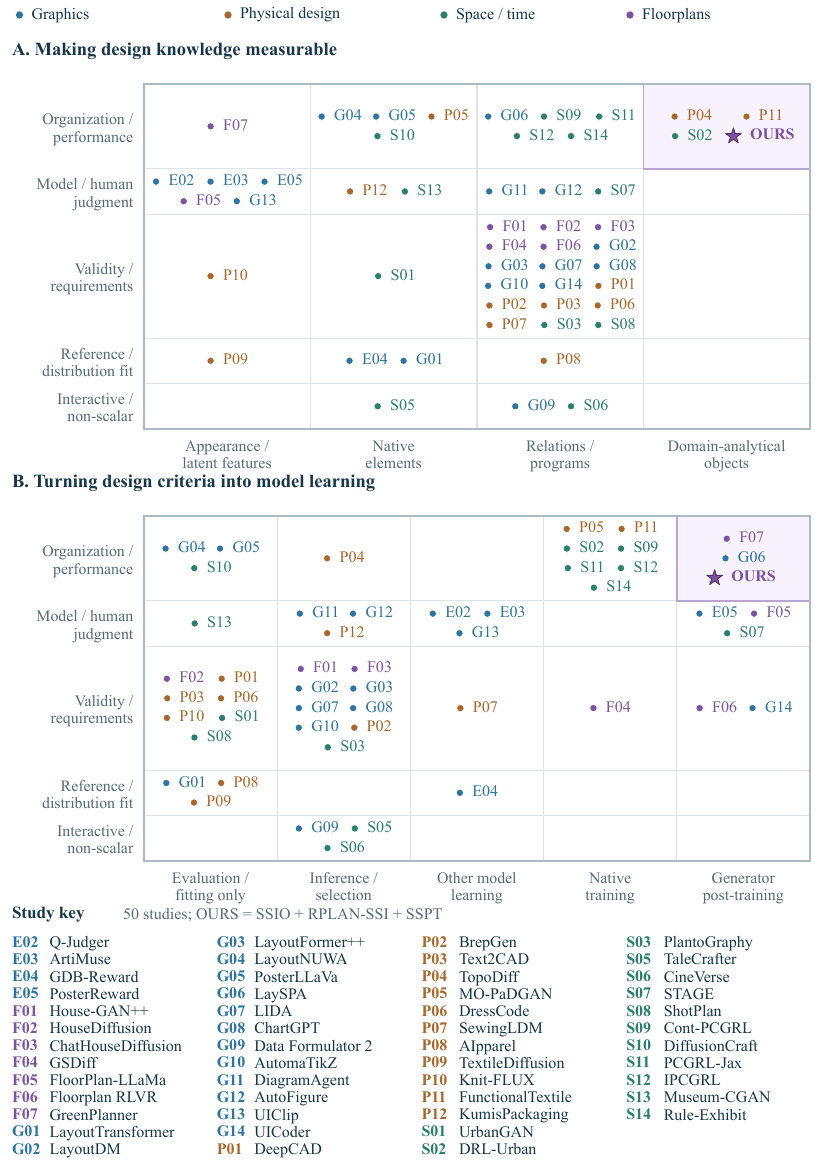}
  \caption{\textbf{From design signals to model learning.} Fifty studies are classified by (A) analysis interface and signal basis and (B) the component affected. Organization/performance includes authored criteria and physical objectives as well as theory-derived measures; Figure~\ref{fig:design-survey-knowledge-basis} separates these foundations. The star denotes our \ssio{}/\rplanssi{}/\sspt{} pathway. The key names each study; Section~\ref{app:survey-inventory} provides references.}
  \label{fig:design-survey-scatter}
\end{figure}

\paragraph{RQ2 finding: explicit computation and theoretical grounding are independent.} A graph exposes relations, a rule computes a score, and preference labels supply judgments. None alone explains why the measured property matters for a design task. Theory-derived measures use the theory's analytical objects and interpretation to make that connection. Figure~\ref{fig:design-survey-scatter} compares representations and signal use; Figure~\ref{fig:design-survey-knowledge-basis} compares their knowledge sources. The next step depends on the missing connection: formalize existing knowledge, connect an established measure to learning, or validate a proposed measure against its intended purpose.

\paragraph{Knowledge sources behind computable criteria.} Specified requirements occur in 41 records, authored criteria in 18, and ratings or preferences in 17; these sources overlap (Figure~\ref{fig:design-survey-full-knowledge}). Established principles and physical models also enter learning. The named space-syntax use in this inventory concerns museum assessment and is documented at abstract level~\citep{li2026museum}.

\subsection{RQ3: From Design Criteria to Reusable Generator Learning}
\label{app:survey-feedback}

Having distinguished what a design criterion measures and where it comes from, we now ask how it affects generation. We compare the objective, training stage, and component updated.

\subsubsection{Training Stage and Design Objective}

Pretraining, continued pretraining (CPT), supervised fine-tuning (SFT), preference optimization, and RL describe how or when a model learns, not what design property it learns. CPT may extend domain modeling~\citep{gururangan2020dontstop}; SFT may itself be post-training; RL may train a policy from scratch. UICoder combines compiler- and relevance-filtered SFT with preference optimization~\citep{wu2024uicoder}. We therefore record the optimized property and updated component alongside the training route.

LayoutTransformer fits structured layouts and HouseDiffusion denoising targets~\citep{gupta2021layouttransformer,shabani2023housediffusion}; their examples carry organization even when the objective does not measure its purpose. LayoutNUWA and PosterLLaVa similarly fine-tune pretrained models on reference specifications~\citep{tang2024layoutnuwa,yang2024posterllava}. Explicit design purpose can also enter initial training: MO-PaDGAN adds performance and diversity objectives, DRL-Urban rewards accessibility, ecology, and road efficiency, and Rule-Exhibit converts design rules into topology and placement supervision~\citep{chen2021mopadgan,zheng2023urbanplanning,xu2025ruleexhibit}. The distinction lies in the property driving updates, whether supplied through examples, losses, rewards, or constructed targets.

\subsubsection{Where Design Feedback Acts}

\begin{figure}[p]
  \centering
  \includegraphics[width=\linewidth]{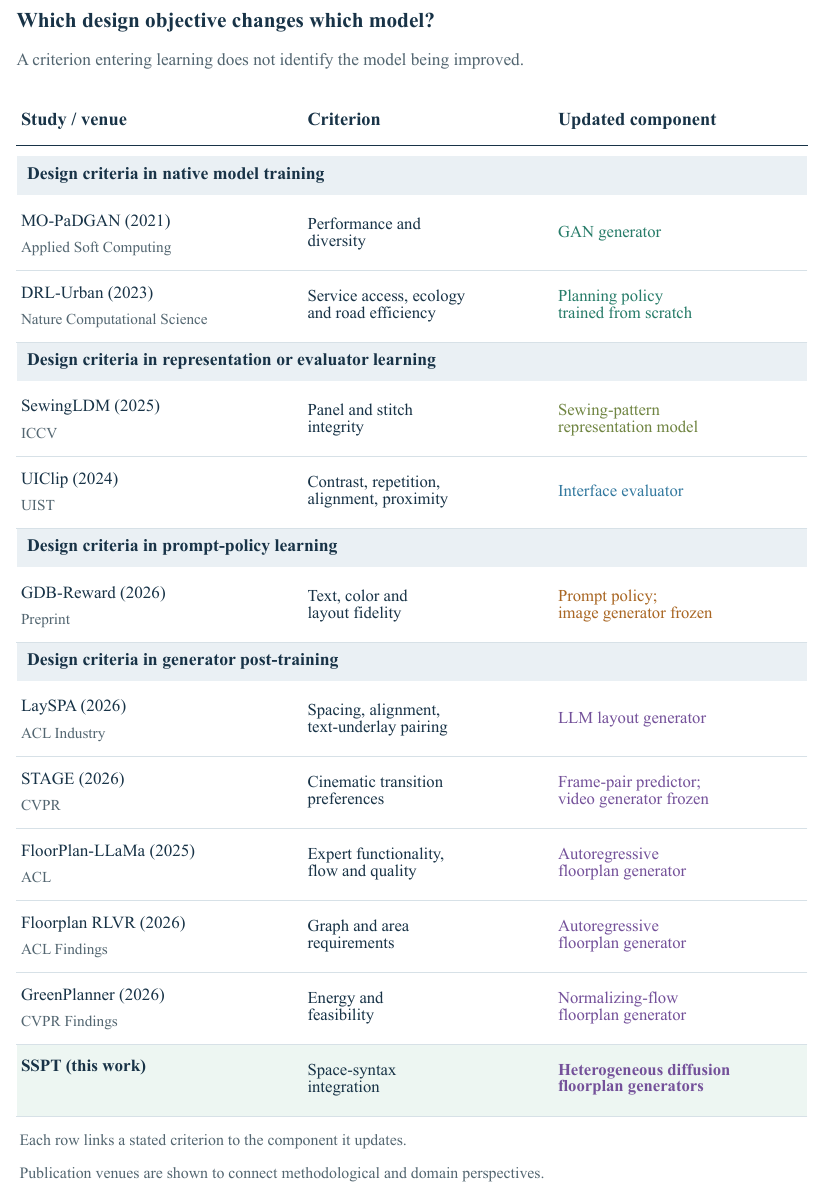}
  \caption{\textbf{A design criterion can update different components.} The examples separate native training, representation or evaluator learning, prompt-policy learning, and generator post-training. \sspt{} connects one architectural measurement to heterogeneous diffusion policies.}
  \label{fig:design-survey-timeline}
\end{figure}

A criterion may evaluate outputs, guide sampling, select candidates, train an evaluator, optimize prompts, or update the generator. LayoutDM and TopoDiff guide inference; UIClip trains an evaluator; GDB-Reward updates a prompt policy; LaySPA and PosterReward update generators~\citep{inoue2023layoutdm,maze2023topodiff,wu2024uiclip,deganutti2026gdbreward,li2026layspa,lai2026posterreward}. Figure~\ref{fig:design-survey-timeline} distinguishes these updated components from evaluation and selection with fixed generator weights.

\begingroup\interlinepenalty=10000
Differentiable rewards, policy gradients, and preference or group-relative objectives provide adaptation mechanisms~\citep{clark2024draft,prabhudesai2024alignprop,fan2023dpok,black2024ddpo,zhao2024scoresasactions,zhao2025ctrl,hu2025b2diffurl,christiano2017preferences,ouyang2022instructgpt,wallace2024diffusiondpo,yang2024d3po,liu2025flowgrpo}; DPPO extends diffusion policy gradients to control and robotics~\citep{ren2024dppo}. Reward-ranked and synthetic-data retraining offer further routes~\citep{dong2023raft,ferbach2024selfconsuming,alemohammad2024mad}. These methods specify how a signal changes a model. Design theory supplies the property to improve and the objects on which it is measured.
\par
\endgroup

\subsubsection{Post-Training Existing Generators}

Purpose-driven initial training builds a generator around a design objective. Post-training asks how that objective can improve an established generator while retaining its geometric quality and response to conditioning.

\begin{figure}[!htb]
  \centering
  \includegraphics[width=\linewidth]{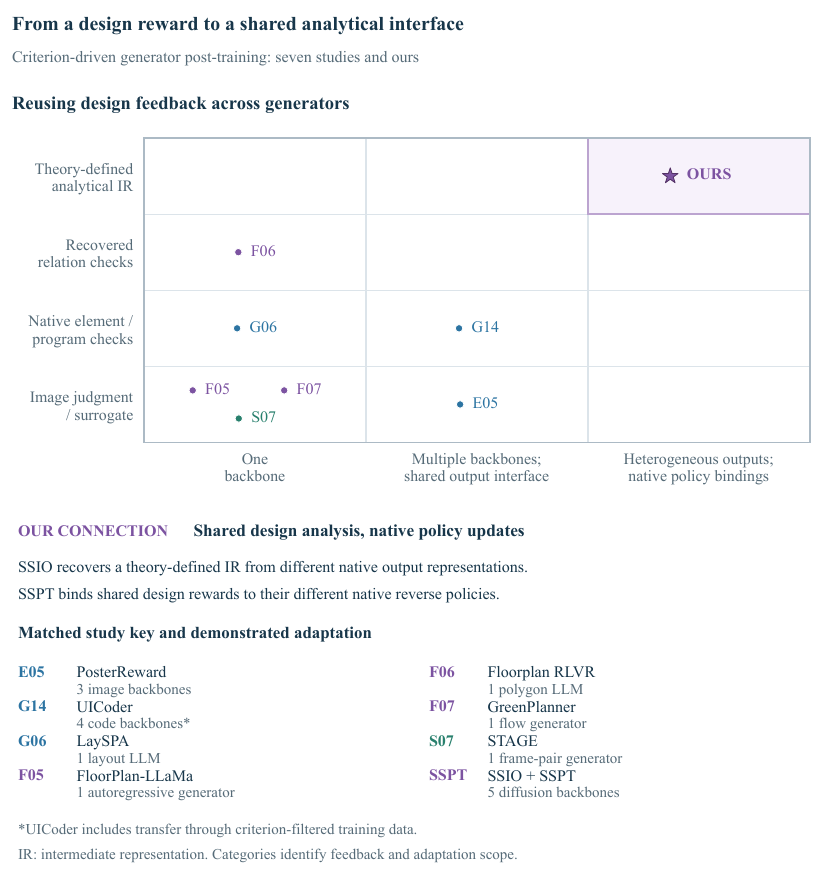}
  \caption{\textbf{Post-training scope and feedback interface.} Seven prior methods and our pathway are classified by their feedback basis and demonstrated generator adaptation. IR denotes intermediate representation. \ssio{} and \sspt{} connect theory-defined analytical objects to native policy updates across heterogeneous outputs.}
  \label{fig:design-survey-posttraining-scope}
\end{figure}

In architecture, prior work applies visibility or space-syntax analysis to reconstructed plans, small residences, and museum layouts~\citep{heo2023visibility,man2026smallspace,li2026museum}. FloorPlan-LLaMa instead post-trains an autoregressive image-token generator through learned architect preferences; floorplan RLVR adapts an autoregressive layout model through graph, area, and validity rewards~\citep{yin2025floorplanllama,lara2026floorplanrlvr}. The former leaves the spatial basis implicit; the latter verifies specified conditions rather than a theory-defined design property. GreenPlanner targets energy and feasibility in a flow generator~\citep{zeng2026greenplanner}; other methods alter conditions, outputs, topology, or latent control~\citep{zeng2025card,zong2026housetune,zhang2026doubledqn,guldilek2026ppogan}. \sspt{} connects explicit architectural measurement to post-training across heterogeneous floorplan diffusion policies.

PosterReward adapts several image-generator backbones, and UICoder several UI-code backbones~\citep{lai2026posterreward,wu2024uicoder}. Our setting additionally requires recovering spaces and access relations from different floorplan formats, so that one accessibility measure can guide generators with different policies (Figure~\ref{fig:design-survey-posttraining-scope}). Each generator retains its own synthesis and update mechanisms.

\paragraph{RQ3 finding: the learning destination determines the capability changed.} Assessment and selection can improve the artifacts delivered to users; generator updates change the model that produces them. Figure~\ref{fig:design-survey-full-knowledge} records which component each criterion affects. Reusing a design objective across generators requires a common measurement and an update method suited to each generator.

\subsection{Cross-RQ Synthesis: The Floorplan Study in Context}
\label{app:survey-synthesis}

The three questions connect design purpose, measurement, and learning. A study needs evidence that a design property matters (RQ1), a computable measure with a clear domain meaning (RQ2), and a way to use that measure to update a generator (RQ3). Floorplan generation lets us examine all three connections in one task.

\paragraph{Existing generation capabilities make the question testable.} Floorplan methods condition on boundaries, room programs, relations, partial specifications, and activities~\citep{wu2019rplan,he2022iplan,hu2020graph2plan,nauata2021houseganpp,para2021constraintgraphs,zhang2024maskplan,hong2024cons2plan,wang2023actfloorgan}. Explicit room and wall modeling~\citep{sun2022wallplan,liu2022panoptic,tang2023graphtransformergan,wang2025diffplanner,hu2025gsdiff} supports heterogeneous diffusion, autoregressive, and multimodal generators~\citep{shabani2023housediffusion,zeng2024floorplandiffusion,wang2026ce2eplan,lian2026cgfloor,qin2024chathousediffusion,shiohara2026fmlm,qin2026housemind}. This diversity motivates studying whether architectural feedback can improve established generators without replacing their architectures; our experiments test this question on five diffusion backbones.

Graph-constrained generation~\citep{upadhyay2022flnet,aalaei2023graphcgan} and neural-guided optimization~\citep{zheng2023neuralguided} exemplify different routes to satisfying specified relations. A shared theory-defined evaluation can assess the spatial organization produced by either route.

\paragraph{The design objective is ready for explicit translation.} Space syntax defines spatial objects and accessibility measures; residential corpora show how these measures vary across residential contexts. Existing heterogeneous generators then let us test whether shared architectural feedback improves spatial organization while retaining required output quality. These resources make the theory-to-learning question in Section~\ref{app:survey-feedback} concrete.

\paragraph{Implications and open directions.} The floorplan study suggests a sequence for other tasks: represent the objects defined by domain knowledge, ground their measurements in design evidence, and use the resulting feedback to improve existing generators. Where a suitable theory exists, its objects and measures need computational implementation and empirical references. Where it does not, proposed representations and scores first need evidence that they capture the intended design purpose. Appendix~\ref{app:design-learning-principles} develops this procedure and the longer-term construction of AI-native design theories (Section~\ref{app:ai-native-design-theories}).

\FloatBarrier
\subsection{Validation Evidence and Complete Study Inventory}

\subsubsection{Validation at the Level of the Claimed Contribution}
\label{app:survey-validation}

Held-out synthesis tests measure generator behavior across examples; professional studies examine decisions, experience, or use in a particular setting. Table~\ref{tab:survey-validation} separates training or source-data scale, independent test cases, and human evaluation because these counts support different claims.

\begingroup
\small
\setlength{\tabcolsep}{3pt}
\setlength{\LTcapwidth}{\linewidth}
\renewcommand{\arraystretch}{1.12}
\begin{longtable}{@{}>{\raggedright\arraybackslash}p{0.24\linewidth}>{\raggedright\arraybackslash}p{0.36\linewidth}>{\raggedright\arraybackslash}p{0.36\linewidth}@{}}
\caption{\normalsize Validation protocols answer different questions. Source-data volume, independent design cases, and human assessments are reported separately.}\label{tab:survey-validation}\\
\toprule
Study / community & Reported scale or protocol & What the evidence tests \\
\midrule
\endfirsthead
\caption[]{\normalsize Validation protocols answer different questions (continued).}\\
\toprule
Study / community & Reported scale or protocol & What the evidence tests \\
\midrule
\endhead
\midrule
\multicolumn{3}{r}{\emph{Continued on the next page.}}\\
\endfoot
\bottomrule
\endlastfoot
DRL-Urban / interdisciplinary\newline\citep{zheng2023urbanplanning} & One synthetic scenario; two real communities; repeated policy runs; eight professional-planner comparisons; 100 design-trained evaluators. & Planning objectives and professional~comparisons. \\
\addlinespace[3pt]
Museum-CGAN / interdisciplinary\newline\citep{tang2024museumcgan} & 100 images; 264 valid questionnaire responses; no conventional train/test split reported. & Perceived design quality; held-out generalization not~established. \\
\addlinespace[3pt]
Small-residence analysis / specialist journal\newline\citep{man2026smallspace} & 14,373 extracted small-layout samples; one 30\,m$^2$ residential evaluation case. & Dataset analysis and a design-assessment case; no large-scale adaptation test. \\
\addlinespace[3pt]
Rule-Exhibit / graphics\newline\citep{xu2025ruleexhibit} & Rule-generated supervision; quantitative comparisons; dataset size/split unconfirmed in the inspected report. & Rule-based learning; independent testing~breadth~unspecified. \\
\addlinespace[3pt]
Data Formulator 2 / HCI\newline\citep{wang2025dataformulator2} & User study: eight participants. & Iterative analysis and~interaction. \\
\addlinespace[3pt]
LaySPA / AI\newline\citep{li2026layspa} & 3,000 training layouts per dataset; 1,000 and 905 test layouts. & Post-training and held-out layout~criteria. \\
\addlinespace[3pt]
Kumis packaging / interdisciplinary\newline\citep{zhao2026kumis} & 106 cultural items; 38 selected items; three final packaging designs. & Knowledge-guided human--AI proposal~development. \\
\end{longtable}
\endgroup
\subsubsection{Cited Study Inventory}
\label{app:survey-inventory}

Each entry gives the study name and citation, represented objects, design criterion, and the step affected by that criterion. IDs link studies to the figures. Abstract-level and contextual records are marked separately. Research focus (F) and knowledge basis (K) are coded independently of representation (R) and criterion use (P).

\textbf{Publication setting:} AI = AI/ML/CV/NLP; HG = HCI, graphics, and visualization; DS = domain-specialist; IX = interdisciplinary; PP = preprint without an assigned venue.

\textbf{Representation (R):} R0 = implicit in images or latent tokens; R1 = explicit elements and attributes; R2 = explicit relations, programs, or operational constraints.
\textbf{Criterion use (P):} P0 = example fitting; P1 = separate evaluation; P2 = inference or selection; P3 = criterion-driven learning.

\textbf{Research focus (F):} S = synthesis/validity, D = design purpose, B = both, U = unresolved.
\textbf{K:} T = domain theory, P = established principles, A = authored criteria, R = specified requirements, M = physical models, J = ratings/preferences, E = example-based only, U = unresolved. Multiple K codes indicate combined sources, not a ranking of theoretical maturity.

\begingroup
\small
\setlength{\tabcolsep}{3pt}
\setlength{\LTcapwidth}{\linewidth}
\setlength{\LTpre}{8pt}
\setlength{\LTpost}{10pt}
\renewcommand{\arraystretch}{1.1}

\Needspace{8\baselineskip}
\begin{longtable}{@{}>{\raggedright\arraybackslash}p{0.30\linewidth}>{\raggedright\arraybackslash}p{0.37\linewidth}>{\raggedright\arraybackslash}p{\dimexpr0.33\linewidth-12pt\relax}@{}}
\caption{\normalsize Layout generation and spatial control in graphic design.}\label{tab:survey-graphic-1}\\
\toprule
Study / source & Representation and criterion & Where the criterion acts \\
\midrule
\endfirsthead
\caption[]{\normalsize Layout generation and spatial control in graphic design (continued).}\\
\toprule
Study / source & Representation and criterion & Where the criterion acts \\
\midrule
\endhead
\midrule
\multicolumn{3}{r}{\emph{Continued on the next page.}}\\
\endfoot
\bottomrule
\endlastfoot
\textbf{LayoutTransformer}\newline\citep{gupta2021layouttransformer}\newline{\color{protocoltext}G01; AI; R1, P0; F:S; K:E} & Element boxes and labels. Layout distribution and~completion. & Autoregressive likelihood~fitting. \\*
\addlinespace[2pt]
\textbf{LayoutDM}\newline\citep{inoue2023layoutdm}\newline{\color{protocoltext}G02; AI; R2, P2; F:S; K:R} & Elements with operational pairwise constraints. Size / position relation~satisfaction. & Denoising; relation energies guide~sampling. \\
\addlinespace[2pt]
\textbf{LayoutFormer++}\newline\citep{jiang2023layoutformerplusplus}\newline{\color{protocoltext}G03; AI; R2, P2; F:S; K:R} & Serialized element and relation constraints. Constraint~satisfaction. & Likelihood fitting; restricted~decoding. \\
\addlinespace[2pt]
\textbf{LayoutNUWA}\newline\citep{tang2024layoutnuwa}\newline{\color{protocoltext}G04; AI; R1, P1; F:S; K:A} & Layout attributes in editable code. Alignment and overlap, evaluated. & Reference-code instruction~tuning. \\
\addlinespace[2pt]
\textbf{PosterLLaVa}\newline\citep{yang2024posterllava}\newline{\color{protocoltext}G05; PP; R1, P1; F:B; K:A+R} & Poster elements as JSON boxes. Saliency-aware placement, evaluated. & Visual instruction~tuning to reference~layouts. \\
\addlinespace[2pt]
\textbf{LaySPA}\newline\citep{li2026layspa}\newline{\color{protocoltext}G06; AI; R2, P3; F:B; K:A+R} & Explicit element environment and relations. Spacing, alignment and text-underlay~pairing. & GRPO updates the LLM layout~generator. \\*
\addlinespace[2pt]
\textbf{LIDA}\newline\citep{dibia2023lida}\newline{\color{protocoltext}G07; AI; R2, P2; F:B; K:A+R} & Data goals and executable chart code. Executable and relevant~visualizations. & LLM inference, execution and~selection. \\
\end{longtable}

\Needspace{8\baselineskip}
\begin{longtable}{@{}>{\raggedright\arraybackslash}p{0.30\linewidth}>{\raggedright\arraybackslash}p{0.40\linewidth}>{\raggedright\arraybackslash}p{\dimexpr0.30\linewidth-12pt\relax}@{}}
\caption{\normalsize Intermediate specifications for visualization, diagrams, and interface evaluation.}\label{tab:survey-graphic-2}\\
\toprule
Study / source & Representation and criterion & Where the criterion acts \\
\midrule
\endfirsthead
\caption[]{\normalsize Intermediate specifications for visualization, diagrams, and interface evaluation (continued).}\\
\toprule
Study / source & Representation and criterion & Where the criterion acts \\
\midrule
\endhead
\midrule
\multicolumn{3}{r}{\emph{Continued on the next page.}}\\
\endfoot
\bottomrule
\endlastfoot
\textbf{ChartGPT}\newline\citep{tian2025chartgpt}\newline{\color{protocoltext}G08; HG; R2, P2; F:B; K:R} & Data transformations and chart specifications. Valid data fields and chart~encodings. & Specification SFT; inference validity~filtering. \\*
\addlinespace[2pt]
\textbf{Data Formulator 2}\newline\citep{wang2025dataformulator2}\newline{\color{protocoltext}G09; HG; R2, P2; F:D; K:R} & Data transformations and encoding specifications. Iterative analytical task~completion. & User and LLM revise data / code; no new~weights. \\
\addlinespace[2pt]
\textbf{AutomaTikZ}\newline\citep{belouadi2024automatikz}\newline{\color{protocoltext}G10; AI; R2, P2; F:S; K:R} & Executable TikZ programs. Compilable scientific vector~graphics. & Code SFT; compiler-guided~retry. \\
\addlinespace[2pt]
\textbf{DiagramAgent}\newline\citep{wei2025diagramagent}\newline{\color{protocoltext}G11; AI; R2, P2; F:S; K:A+R} & Diagram plans and executable diagram code. Completeness and~compilability. & Code-agent SFT; inference~checking. \\
\addlinespace[2pt]
\textbf{AutoFigure}\newline\citep{zhu2026autofigure}\newline{\color{protocoltext}G12; AI; R2, P2; F:B; K:A+R} & Entity-relation blueprint and rendered diagram. Scientific fidelity and diagram~clarity. & Inference critique and~refinement. \\*
\addlinespace[2pt]
\textbf{UIClip}\newline\citep{wu2024uiclip}\newline{\color{protocoltext}G13; HG; R0, P3; F:D; K:P+A+J} & Screenshots and text assessments. Contrast, repetition, alignment,~proximity. & CLIP evaluator learns perturbations / preferences. \\
\end{longtable}

\Needspace{8\baselineskip}
\begin{longtable}{@{}>{\raggedright\arraybackslash}p{0.30\linewidth}>{\raggedright\arraybackslash}p{0.36\linewidth}>{\raggedright\arraybackslash}p{\dimexpr0.34\linewidth-12pt\relax}@{}}
\caption{\normalsize Interface generation and professional design evaluation.}\label{tab:survey-graphic-3}\\
\toprule
Study / source & Representation and criterion & Where the criterion acts \\
\midrule
\endfirsthead
\caption[]{\normalsize Interface generation and professional design evaluation (continued).}\\
\toprule
Study / source & Representation and criterion & Where the criterion acts \\
\midrule
\endhead
\midrule
\multicolumn{3}{r}{\emph{Continued on the next page.}}\\
\endfoot
\bottomrule
\endlastfoot
\textbf{UICoder}\newline\citep{wu2024uicoder}\newline{\color{protocoltext}G14; AI; R2, P3; F:S; K:R} & SwiftUI code and rendered interfaces. Compilation and prompt~relevance. & Filtered SFT and DPO update code~generator. \\*
\addlinespace[2pt]
\textbf{IDEA-Bench}\newline\citep{liang2025ideabench}\newline{\color{protocoltext}E01; AI; Context; F:B; K:R+J} & Images and task settings; professional task~completion and~quality. & Benchmark; no generator update. Not in coordinate~map. \\
\addlinespace[2pt]
\textbf{Q-Judger}\newline\citep{li2026qwenimagebench}\newline{\color{protocoltext}E02; PP; R0, P3; F:B; K:R+J} & Images and explicit rubric text. 56 professional image-evaluation~facets. & Supervised Q-Judger evaluator~training. \\
\addlinespace[2pt]
\textbf{ArtiMuse}\newline\citep{cao2026artimuse}\newline{\color{protocoltext}E03; AI; R0, P3; F:D; K:J} & Images with aesthetic analyses. Eight aesthetic attributes and holistic~judgment. & Expert-supervised aesthetic~evaluator. \\
\addlinespace[2pt]
\textbf{GDB-Reward}\newline\citep{deganutti2026gdbreward}\newline{\color{protocoltext}E04; PP; R1, P3; F:B; K:A+R+J} & Layout metadata, prompts and images. Text / color / layout reference~fidelity. & GRPO updates prompt policy; FLUX remains~frozen. \\*
\addlinespace[2pt]
\textbf{PosterReward}\newline\citep{lai2026posterreward}\newline{\color{protocoltext}E05; AI; R0, P3; F:B; K:A+J} & Poster images and learned textual judgments. Typography, composition and~preferences. & Reward learning; downstream generator RL~adaptation. \\
\end{longtable}

\Needspace{8\baselineskip}
\begin{longtable}{@{}>{\raggedright\arraybackslash}p{0.29\linewidth}>{\raggedright\arraybackslash}p{0.40\linewidth}>{\raggedright\arraybackslash}p{\dimexpr0.31\linewidth-12pt\relax}@{}}
\caption{\normalsize CAD, engineering performance, and garment assembly.}\label{tab:survey-physical-1}\\
\toprule
Study / source & Representation and criterion & Where the criterion acts \\
\midrule
\endfirsthead
\caption[]{\normalsize CAD, engineering performance, and garment assembly (continued).}\\
\toprule
Study / source & Representation and criterion & Where the criterion acts \\
\midrule
\endhead
\midrule
\multicolumn{3}{r}{\emph{Continued on the next page.}}\\
\endfoot
\bottomrule
\endlastfoot
\textbf{DeepCAD}\newline\citep{wu2021deepcad}\newline{\color{protocoltext}P01; AI; R2, P1; F:S; K:R} & Sketch / extrusion command sequences. CAD execution~validity. & Command and parameter~fitting. \\*
\addlinespace[2pt]
\textbf{BrepGen}\newline\citep{xu2024brepgen}\newline{\color{protocoltext}P02; HG; R2, P2; F:S; K:R} & Face-edge-vertex B-rep structure. Geometrically valid~assembly. & Diffusion fitting; geometry repaired at~inference. \\
\addlinespace[2pt]
\textbf{Text2CAD}\newline\citep{khan2024text2cad}\newline{\color{protocoltext}P03; AI; R2, P1; F:B; K:R+J} & Text-conditioned CAD programs. Valid execution and prompt~correspondence. & Conditional CAD sequence~fitting. \\
\addlinespace[2pt]
\textbf{TopoDiff}\newline\citep{maze2023topodiff}\newline{\color{protocoltext}P04; AI; R1, P2; F:B; K:R+M} & Material occupancy and load fields. Compliance and absence of floating~material. & Surrogate gradients guide diffusion~sampling. \\
\addlinespace[2pt]
\textbf{MO-PaDGAN}\newline\citep{chen2021mopadgan}\newline{\color{protocoltext}P05; IX; R1, P3; F:B; K:A+M} & Parametric engineering shapes. Performance and~diversity. & Additional objectives train GAN~generator. \\
\addlinespace[2pt]
\textbf{DressCode}\newline\citep{he2024dresscode}\newline{\color{protocoltext}P06; HG; R2, P1; F:S; K:R+J} & Panels and sewing connections. Draping and prompt~correspondence,~evaluated. & Autoregressive pattern~fitting. \\*
\addlinespace[2pt]
\textbf{SewingLDM}\newline\citep{liu2024sewingldm}\newline{\color{protocoltext}P07; AI; R2, P3; F:B; K:R+J} & Panels and stitch connections. Panel / stitch~integrity. & Integrity losses shape learned~representation. \\
\end{longtable}

\Needspace{8\baselineskip}
\begin{longtable}{@{}>{\raggedright\arraybackslash}p{0.30\linewidth}>{\raggedright\arraybackslash}p{0.37\linewidth}>{\raggedright\arraybackslash}p{\dimexpr0.33\linewidth-12pt\relax}@{}}
\caption{\normalsize Garments, textiles, packaging, and engineering-design metrics.}\label{tab:survey-physical-2}\\
\toprule
Study / source & Representation and criterion & Where the criterion acts \\
\midrule
\endfirsthead
\caption[]{\normalsize Garments, textiles, packaging, and engineering-design metrics (continued).}\\
\toprule
Study / source & Representation and criterion & Where the criterion acts \\
\midrule
\endhead
\midrule
\multicolumn{3}{r}{\emph{Continued on the next page.}}\\
\endfoot
\bottomrule
\endlastfoot
\textbf{AIpparel}\newline\citep{nakayama2025aipparel}\newline{\color{protocoltext}P08; AI; R2, P0; F:S; K:E} & Multimodal sewing-pattern tokens. Paired panel and stitch~reconstruction. & Supervised multimodal generation / editing. \\*
\addlinespace[2pt]
\textbf{TextileDiffusion}\newline\citep{karagoz2023textile}\newline{\color{protocoltext}P09; DS; R0, P0; F:S; K:E} & Textile-pattern images. Appearance and text~similarity. & Image-domain diffusion~adaptation. \\
\addlinespace[2pt]
\textbf{Knit-FLUX}\newline\citep{liu2025knitflux}\newline{\color{protocoltext}P10; IX; R0, P1; F:S; K:A+J} & Rendered knit images. Knit-unit appearance and yarn coverage, evaluated. & Image-domain LoRA~adaptation. \\
\addlinespace[2pt]
\textbf{FunctionalTextile}\newline\citep{tang2026functionaltextile}\newline{\color{protocoltext}P11; DS; R2, P3; F:B; K:R+M} & Reported yarn / interlacing fields; porosity, stiffness and~conductivity. & Reported functional-surrogate loss trains~generator. \\
\addlinespace[2pt]
\textbf{KumisPackaging}\newline\citep{zhao2026kumis}\newline{\color{protocoltext}P12; IX; R1, P2; F:D; K:R+J} & Cultural elements and shape-grammar variants. Cultural communication and packaging~preference. & Human prompting and candidate~selection. \\
\addlinespace[2pt]
\textbf{CraftBeerPackaging}\newline\citep{chen2026craftbeer}\newline{\color{protocoltext}P13; DS; Abstract only\newline F:D; K:U} & Packaging images from existing tools. Visual and brand-expression~assessment. & Tool comparison; abstract-level evidence. Not in coordinate~map. \\*
\addlinespace[2pt]
\textbf{Beyond Statistical Similarity}\newline\citep{regenwetter2023metrics}\newline{\color{protocoltext}P14; DS; Context\newline F:B; K:A+R+M} & Engineering evaluation taxonomy: feasibility, performance, novelty and~conditioning. & Comparative metric study; contextual reference. Not in coordinate~map. \\
\end{longtable}

\Needspace{8\baselineskip}
\begin{longtable}{@{}>{\raggedright\arraybackslash}p{0.30\linewidth}>{\raggedright\arraybackslash}p{0.37\linewidth}>{\raggedright\arraybackslash}p{\dimexpr0.33\linewidth-12pt\relax}@{}}
\caption{\normalsize Urban, landscape, and story visualization.}\label{tab:survey-spatial-1}\\
\toprule
Study / source & Representation and criterion & Where the criterion acts \\
\midrule
\endfirsthead
\caption[]{\normalsize Urban, landscape, and story visualization (continued).}\\
\toprule
Study / source & Representation and criterion & Where the criterion acts \\
\midrule
\endhead
\midrule
\multicolumn{3}{r}{\emph{Continued on the next page.}}\\
\endfoot
\bottomrule
\endlastfoot
\textbf{UrbanGAN}\newline\citep{fedorova2021urbangan}\newline{\color{protocoltext}S01; DS; R1, P1; F:S; K:A+R} & Urban-block rasters and parcel context. Building spacing and urban morphology,~evaluated. & Conditional adversarial~fitting. \\*
\addlinespace[2pt]
\textbf{DRL-Urban}\newline\citep{zheng2023urbanplanning}\newline{\color{protocoltext}S02; IX; R2, P3; F:D; K:P+R} & Parcel-road-intersection graph. Service access, ecology and road~efficiency. & PPO trains a planning policy from~scratch. \\
\addlinespace[2pt]
\textbf{PlantoGraphy}\newline\citep{huang2024plantography}\newline{\color{protocoltext}S03; HG; R2, P2; F:B; K:R+J} & Scene graph and plant-layout constraints. User-specified landscape~composition. & Interactive graph planning; rendering~LoRA. \\
\addlinespace[2pt]
\textbf{GenFusion-Urban}\newline\citep{gao2025genfusion}\newline{\color{protocoltext}S04; DS; Abstract only\newline F:B; K:R+M} & Reported urban layout / performance pipeline. Urban building~performance. & Reported diffusion + surrogate + NSGA-II. Not in coordinate~map. \\*
\addlinespace[2pt]
\textbf{TaleCrafter}\newline\citep{gong2023talecrafter}\newline{\color{protocoltext}S05; HG; R1, P2; F:S; K:R} & Character attributes, layouts and story sequence. Character consistency and story~visualization. & Adapted rendering plus inference~control. \\
\end{longtable}

\Needspace{8\baselineskip}
\begin{longtable}{@{}>{\raggedright\arraybackslash}p{0.30\linewidth}>{\raggedright\arraybackslash}p{0.37\linewidth}>{\raggedright\arraybackslash}p{\dimexpr0.33\linewidth-12pt\relax}@{}}
\caption{\normalsize Cinematic organization and playable levels.}\label{tab:survey-spatial-2}\\
\toprule
Study / source & Representation and criterion & Where the criterion acts \\
\midrule
\endfirsthead
\caption[]{\normalsize Cinematic organization and playable levels (continued).}\\
\toprule
Study / source & Representation and criterion & Where the criterion acts \\
\midrule
\endhead
\midrule
\multicolumn{3}{r}{\emph{Continued on the next page.}}\\
\endfoot
\bottomrule
\endlastfoot
\textbf{CineVerse}\newline\citep{phung2026cineverse}\newline{\color{protocoltext}S06; AI; R2, P2; F:B; K:P+R} & Structured cinematic shot plans. Cinematic composition and~continuity. & Plan-guided inference; keyframe~fitting. \\*
\addlinespace[2pt]
\textbf{STAGE}\newline\citep{zhang2025stage}\newline{\color{protocoltext}S07; AI; R2, P3; F:B; K:A+J} & Intra-shot and inter-shot frame pairs. Cinematic transition~preference. & DPO updates frame-pair predictor; video model~frozen. \\
\addlinespace[2pt]
\textbf{ShotPlan}\newline\citep{guo2026shotplan}\newline{\color{protocoltext}S08; PP; R2, P1; F:B; K:R+J} & Explicit shot / cut planning tokens. Shot controllability and consistency,~evaluated. & Conditional flow matching; no separate planning~loss. \\
\addlinespace[2pt]
\textbf{Cont-PCGRL}\newline\citep{earle2021contpcgrl}\newline{\color{protocoltext}S09; AI; R2, P3; F:D; K:A+R} & Grid regions, paths and edit actions. Target difficulty / path and region~objectives. & From-scratch goal-conditioned~PPO. \\*
\addlinespace[2pt]
\textbf{DiffusionCraft}\newline\citep{dai2024diffusioncraft}\newline{\color{protocoltext}S10; AI; R1, P1; F:B; K:R} & Level tiles and local spatial patterns. A-star playability, evaluated. & Diffusion fitting from one~level. \\
\end{longtable}

\Needspace{8\baselineskip}
\begin{longtable}{@{}>{\raggedright\arraybackslash}p{0.29\linewidth}>{\raggedright\arraybackslash}p{0.39\linewidth}>{\raggedright\arraybackslash}p{\dimexpr0.32\linewidth-12pt\relax}@{}}
\caption{\normalsize Controllable levels and exhibition layouts.}\label{tab:survey-spatial-3}\\
\toprule
Study / source & Representation and criterion & Where the criterion acts \\
\midrule
\endfirsthead
\caption[]{\normalsize Controllable levels and exhibition layouts (continued).}\\
\toprule
Study / source & Representation and criterion & Where the criterion acts \\
\midrule
\endhead
\midrule
\multicolumn{3}{r}{\emph{Continued on the next page.}}\\
\endfoot
\bottomrule
\endlastfoot
\textbf{PCGRL-Jax}\newline\citep{earle2024pcgrljax}\newline{\color{protocoltext}S11; AI; R2, P3; F:D; K:A+R} & Grid edits, goals and fixed elements. Path / region control and~generalization. & From-scratch PPO with scalable~environments. \\*
\addlinespace[2pt]
\textbf{IPCGRL}\newline\citep{baek2025ipcgrl}\newline{\color{protocoltext}S12; AI; R2, P3; F:D; K:A+R} & Language-derived tasks and level edits. Instruction-specified paths and~regions. & Task encoding plus reward-trained~PPO. \\
\addlinespace[2pt]
\textbf{Museum-CGAN}\newline\citep{tang2024museumcgan}\newline{\color{protocoltext}S13; IX; R1, P1; F:B; K:J} & Museum zones and rendered plans. Spatial / visual qualities, evaluated. & Adversarial and perceptual image~fitting. \\
\addlinespace[2pt]
\textbf{Rule-Exhibit}\newline\citep{xu2025ruleexhibit}\newline{\color{protocoltext}S14; HG; R2, P3; F:B; K:A+R} & Hall connectivity and exhibit placement. Connectivity, saliency and exhibit~importance. & Design rules generate supervised training~labels. \\*
\addlinespace[2pt]
\textbf{Museum-LoRA-Compare}\newline\citep{li2026museum}\newline{\color{protocoltext}S15; DS; Abstract only\newline F:B; K:T+J} & Generated museum-plan images. Spatial connectivity, evaluated. & LoRA / GAN comparison; abstract-level evidence. Not in coordinate~map. \\
\end{longtable}

\Needspace{8\baselineskip}
\begin{longtable}{@{}>{\raggedright\arraybackslash}p{0.30\linewidth}>{\raggedright\arraybackslash}p{0.36\linewidth}>{\raggedright\arraybackslash}p{\dimexpr0.34\linewidth-12pt\relax}@{}}
\caption{\normalsize Floorplan synthesis and design-directed learning.}\label{tab:survey-floorplan-1}\\
\toprule
Study / source & Representation and criterion & Where the criterion acts \\
\midrule
\endfirsthead
\caption[]{\normalsize Floorplan synthesis and design-directed learning (continued).}\\
\toprule
Study / source & Representation and criterion & Where the criterion acts \\
\midrule
\endhead
\midrule
\multicolumn{3}{r}{\emph{Continued on the next page.}}\\
\endfoot
\bottomrule
\endlastfoot
\textbf{House-GAN++}\newline\citep{nauata2021houseganpp}\newline{\color{protocoltext}F01; AI; R2, P2; F:S; K:R} & Room graph and room masks. Graph compatibility and geometric~quality. & GAN training; schedule~meta-optimization. \\*
\addlinespace[2pt]
\textbf{HouseDiffusion}\newline\citep{shabani2023housediffusion}\newline{\color{protocoltext}F02; AI; R2, P1; F:S; K:R} & Graph-conditioned vector polygons. Geometry and compatibility, evaluated. & Noise and binary-coordinate~regression. \\
\addlinespace[2pt]
\textbf{ChatHouseDiffusion}\newline\citep{qin2024chathousediffusion}\newline{\color{protocoltext}F03; PP; R2, P2; F:S; K:R} & Language-derived relations and raster plan. User conditions and controlled~editing. & Conditional denoising and inference~editing. \\
\addlinespace[2pt]
\textbf{GSDiff}\newline\citep{hu2025gsdiff}\newline{\color{protocoltext}F04; AI; R2, P3; F:S; K:R} & Boundary graph and vector geometry. Alignment and edge~validity. & Native geometric losses train~generator. \\
\addlinespace[2pt]
\textbf{FloorPlan-LLaMa}\newline\citep{yin2025floorplanllama}\newline{\color{protocoltext}F05; AI; R0, P3; F:B; K:J} & Image tokens; image-based preference evaluator. Expert functionality, flow and overall~judgment. & PPO updates autoregressive image-token~generator. \\
\addlinespace[2pt]
\textbf{Floorplan RLVR}\newline\citep{lara2026floorplanrlvr}\newline{\color{protocoltext}F06; AI; R2, P3; F:S; K:R} & JSON polygons and reconstructed relations. Connectivity-graph and area~requirements. & GRPO updates autoregressive~generator. \\*
\addlinespace[2pt]
\textbf{GreenPlanner}\newline\citep{zeng2026greenplanner}\newline{\color{protocoltext}F07; AI; R2, P3; F:B; K:P+R+M} & Conditioned vector / topology representation. Learned energy and feasibility~criteria. & Gradient feedback adapts normalizing-flow~generator. \\
\end{longtable}
\endgroup

\FloatBarrier
\Needspace{6\baselineskip}

\section{Design Theory Translation: Principles for Model Improvement}
\label{app:design-learning-principles}

\begin{figure}[!ht]
  \centering
  \includegraphics[width=\linewidth]{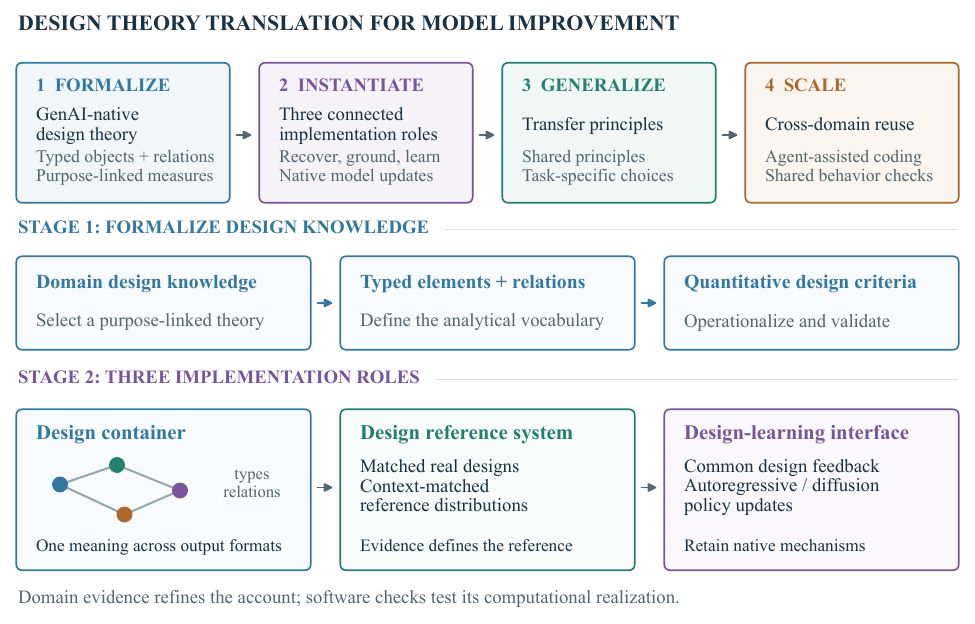}
  \caption{\textbf{A proposed workflow for design theory translation.} Informed by the floorplan study, the overview connects formalization, implementation, and cross-domain extension. Stage 1 defines typed analytical objects and quantitative criteria grounded in domain meaning. Stage 2 recovers the design container from artifacts, grounds measurements in empirical references, and connects shared feedback to native model updates.}
  \label{fig:theory-translation-procedure}
\end{figure}

The floorplan study and review motivate a proposed workflow for using design theory to improve generators in other tasks. It addresses three questions: what evidence supports the intended design benefit (RQ1, Section~\ref{app:survey-priorities}), how a measure captures that benefit (RQ2, Section~\ref{app:survey-evaluation}), and how feedback updates the generator (RQ3, Section~\ref{app:survey-feedback}). We first define the objects, relations, and measures required by the theory, then organize their use into a design container, an empirical reference system, and a learning interface (Figure~\ref{fig:theory-translation-procedure}). The sections below explain these roles, derive construction questions from the floorplan evidence, and illustrate agent-assisted implementation.

For task-specific construction routes, see Section~\ref{app:domain-operators}; for worked implementations, see Section~\ref{app:agent-theory-translation}.

\subsection{Generative AI-Native Design Theory Construction}
\label{app:ai-native-design-theories}

\paragraph{The translation challenge.}
Design theories explain how artifacts serve their purposes, but their concepts do not always specify what a model should represent or measure. Using them in generator learning requires explicit design objects, relations, context, and quantitative properties. We call theories developed or reformulated in this form \emph{Generative AI-native design theories}. Here, native means usable by a generative learning process; the theory itself may predate GenAI.

Space syntax illustrates this translation. It already connects spatial organization to graphs and quantitative accessibility measures~\citep{hillier1984sociallogic,hillier1996space}. Its existing objects and measures define the design container; \ssio{} recovers them from floorplans, residential corpora provide references, and \sspt{} connects measurement to learning. Other domains can follow the same sequence---\emph{design purpose, relational representation, quantitative analysis, and evidence of design consequences}---using properties appropriate to their tasks.

\paragraph{Constructing a relational account.}
Start with a theory that explains the intended design purpose and identifies relevant relations. Use it to define the elements, functional types, and directed, spatial, temporal, or functional relations to represent. A typed, attributed graph can store these relations together with any required geometry, material, time, or user context. Types carry design meaning: a label differs from a process, a control from a navigation destination, and a service space from a public space. Visual adjacency alone does not express these distinctions.

Next, turn the theory's claims into measurable properties. Use established formulas when available. For a proposed measure, specify its definition and the controlled design changes it should distinguish. For example, a grouping principle can motivate a measure of disagreement between intended groups and displayed regions; a task-performance model can predict the duration of an executable interaction sequence. The first proposes a theory-inspired metric; the second implements an existing quantitative model. Domain tasks, professional judgments, or observed use must establish whether the measure captures the intended benefit. Real-design distributions then provide context-specific reference values.

\paragraph{Designing for the generative process.}
The same design should receive the same interpretation across output formats. Specify how to recover its elements and relations, identify outputs that cannot be measured, and compare alternatives under the same brief. A completed design can then supply feedback to autoregressive or diffusion updates even when graph recovery is nondifferentiable. Generation experiments may reveal missing distinctions, prompting revisions to the types, relations, or measures that require further domain evidence.

Section~\ref{app:domain-operators} identifies quantitative theories that can be implemented directly, theories requiring new measurements, and complementary standards or physical models across fourteen domains. Several sources may serve different purposes within one domain. The next three sections explain how to recover the resulting representation, interpret its scores against real designs, and use them in model updates.

\subsection{The Design Container: A Theory-Defined Intermediate Representation}
\label{app:design-container}

A \emph{design container} is a theory-defined intermediate representation of the objects, relations, attributes, and context needed to compute design properties. The theory defines what these objects and relations mean; recovery operators extract them from each generator's native output. A useful common form is a typed, attributed graph \(z_d=(V_d,E_d,\tau_d,a_d,c)\): \(V_d\) contains design elements, \(E_d\) their relations, \(\tau_d\) assigns element and relation types, \(a_d\) stores attributes, and \(c\) records the design context. Higher-order relations can be represented by relation nodes. Examples include spaces linked by access, information units by reading order, garment panels by seams, and components by functional interfaces.

Compiler intermediate representations provide a useful analogy: LLVM and MLIR support shared analyses across different source languages through explicit semantics~\citep{lattner2004llvm,lattner2021mlir,llvm2026langref}. Here, $\mathcal A_{d,m}$ recovers a domain representation $z=\mathcal A_{d,m}(x,c)$ from artifact $x$ and context $c$, and theory-defined measure $\mathcal M_d$ computes $s=\mathcal M_d(z,c)$. Comparisons across generators therefore need consistent object and relation definitions, format-specific recovery rules, and shared scoring and aggregation.

\ssio{} instantiates this container with convex-space access graphs and integration. The GSDiff case shows why construction must distinguish recoverable information from missing relations (Section~\ref{sec:gsdiff-case}; Appendix~\ref{app:gsdiff-completion}): a room-boundary graph does not determine access. When completion supplies those relations, its recovery fidelity and the generator's subsequent improvement require separate evaluation.

\subsection{The Design Reference System: Theory, Evidence, and Evaluation}
\label{app:design-grounding-protocol}

A \emph{design reference system} combines theory-defined properties with evidence from real designs. Theory specifies what to measure; intended use explains why it matters; matched real designs provide reference values. Comparisons should match relevant requirements, scale, functions, users, and environment, and recover the same design container from real and generated designs.

Comparing corpora helps distinguish recurring patterns from collection-specific conventions. Task performance, simulation, professional assessment, or theory can support their interpretation. References can thus describe a range of valid designs rather than prescribe one numerical target.

In the floorplan study, Public-first organization recurs across corpora, while their score distributions differ. \rplanssi{} therefore uses matched residential references to evaluate population-level profile alignment and within-plan Public anchoring. The generator--corpus comparison separately identifies the generator's weaker functional hierarchy.

For another task, this suggests three evaluation steps: check recovered objects and relations, assess individual designs against their purpose, and compare distributions, coverage, and variability with matched real designs. Publish the recovery rules, reference splits, matching criteria, metrics, coverage, and uncertainty, including recovery failures.

\subsection{The Shared Design-Learning Interface: Feedback and Native Updates}
\label{app:design-native-learning}

A \emph{shared design-learning interface} uses the same design measurements to guide different generators' native updates. Reference distributions calibrate and evaluate the objective; feedback on individual outputs drives learning. The measurement stays common while each model retains its action space.

For an \emph{autoregressive generator}, actions are sequence decisions. A complete artifact can be decoded into its design container and measured to supply rewards for policy-gradient updates or rankings for preference optimization~\citep{schulman2017ppo,ouyang2022instructgpt,rafailov2023dpo}. Tokenization, likelihoods, and action masks remain native to the sequence model.

For a \emph{diffusion generator}, actions are stochastic reverse transitions. DDPO supplies this policy-learning view~\citep{black2024ddpo}; each backbone's reverse variables, transition probabilities, conditioning, masks, and variances determine its policy update. \sspt{} demonstrates these bindings across five floorplan backbones, and its HouseDiffusion support ablation shows that update placement matters under a fixed transition budget.

\sspt{} implements this interface for diffusion models. Applying it to autoregressive generators would require sequence-specific updates using the same design feedback.

\subsection{Retaining Required Quality During Adaptation}
\label{app:design-quality-protocol}

First, specify how much output quality may deteriorate and which validity requirements must hold. Use development experiments to tune update strength and control reward shortcuts. On validation conditions, discard checkpoints that violate these requirements, then select the remaining checkpoint with the best design performance. Finally, evaluate the selected checkpoint on independent test conditions, reporting output quality and design performance separately.

HouseDiffusion illustrates this trade-off. Retrospective calibration from frozen Base evaluations establishes a native-quality reference range (Appendix~\ref{app:selection}). The selected model remains within this range while \ipd{} and \piad{} decrease by 27.0\% and 32.2\%, respectively; FID and Compatibility increase relative to matched Base. Area control reduces the native-quality cost of configurational feedback. Sensitivity analysis shows how changing the quality requirement changes checkpoint choice (Appendix~\ref{app:selection-results}).

Other domains need task-specific quality requirements, such as rendering validity, geometric consistency, program execution, or simulation feasibility. Candidate ways to preserve quality include limiting adaptation, retaining the native training objective as a regularizer, and constraining changes from a reference policy. Their effects require independent testing.

\subsection{From Floorplan Evidence to a Reusable Research Procedure}
\label{app:design-transfer-procedure}

The procedure starts with a design purpose and a theory that explains the relevant properties. If no operational theory exists, first formulate one from domain knowledge. Define its objects and measures, recover them from generated outputs, compare them with matched real designs, and use the resulting feedback for model updates. Independent design outcomes and output-quality tests determine whether those updates improve the task.

The floorplan experiments suggest checks for each step: compare corpora to distinguish recurring patterns from context-specific distributions; use matched cases to separate organization from link count; ablate update location and shortcut control; and evaluate relation recovery separately from learning, as in GSDiff. The domain sketches below ask how these checks could apply elsewhere.
\subsection{A Fourteen-Domain Map of Construction Routes}
\label{app:domain-operators}

Whether a theory is ready to use depends on the design purpose. The fourteen sketches distinguish three starting points: quantitative theories with defined objects and measures; explanatory theories or principles needing computational definitions; and complementary standards or physical models. These distinctions follow the knowledge-source classification in Appendix~\ref{app:survey-evaluation}. References and generated outputs should use the same recovery rules; controlled design changes and task outcomes can then test what the scores mean.

Table~\ref{tab:domain-theory-construction} summarizes the starting points. Spatial access, expert task execution, requirement-defined products, and specified graphical encodings have relevant quantitative models. Communication, continuity, and apparel needs require further task-specific measures. Formal notation, symmetry, folding mechanics, and graph grammars provide other useful tools. Each sketch distinguishes the existing foundation from the work still needed.

\begin{table}[!t]
\centering
\caption{\textbf{Construction entry points across fourteen design categories.} Knowledge sources and construction needs are distinguished; sources and proposed measurements are detailed below.}
\label{tab:domain-theory-construction}
\small
\setlength{\tabcolsep}{4pt}
\renewcommand{\arraystretch}{1.12}
\begin{tabular}{@{}>{\raggedright\arraybackslash}p{0.25\linewidth}>{\raggedright\arraybackslash}p{0.35\linewidth}>{\raggedright\arraybackslash}p{\dimexpr0.40\linewidth-16pt\relax}@{}}
\toprule
Design category & Available foundation & Construction focus \\
\midrule
\multicolumn{3}{@{}l}{\emph{Existing quantitative or formal analytical accounts}} \\*
Floorplans / interiors & Space syntax & Configuration recovery and~grounding \\
Urban / landscape & Space syntax; cooling models & Purpose-specific network or~climate~inputs \\
UI / interaction & Keystroke-Level Model & Executable methods and~calibrated~timings \\
Products / CAD & Axiomatic Design & Recover dependencies; estimate requirement~success \\
Visualization / maps & Expressiveness; projection theory & Task-matched encodings and~distortion \\
Exhibition / wayfinding & Access and cross-visibility analysis & Visitor goals and interpretive~outcomes \\
\addlinespace
\multicolumn{3}{@{}l}{\emph{Explanatory theories and principles requiring operationalization}} \\*
Posters / graphics & Common-region grouping & Intended versus perceived~grouping \\
Editorial & Discourse relations; spatial contiguity & Content-to-layout correspondence \\
Cinema / storyboards & Attentional continuity theory & Cut-specific continuity~measurements \\
Garments & Functional--expressive--aesthetic needs & Purpose-specific fit and~movement~criteria \\
\addlinespace
\multicolumn{3}{@{}l}{\emph{Complementary formal methods, standards, and physical models}} \\*
Scientific diagrams & SBGN typed notation & Semantic conformance and~comprehension \\*
Textiles / patterns & Symmetry; textile simulation & Repeat structure or physical~performance \\
Packaging & Fold kinematics; construction/test~standards & Assembly, use, and protection~objectives \\*
Game levels & Mission--space grammars; MDA & Progression constraints and player~outcomes \\
\bottomrule
\end{tabular}
\end{table}

\subsubsection{Communication and Interaction}
\label{app:design-operators-communication}

Here, organization primarily serves interpretation or task completion. Typography, data mappings, temporal order, and executable transitions belong in the container whenever they carry design meaning; references must be matched by content, audience, and task.

\paragraph{Poster and graphic layout.}
Common-region grouping is an established perceptual principle explaining association~\citep{palmer1992commonregion}, while learned layout analyses already quantify importance, alignment, and grouping~\citep{odonovan2014learninglayouts}. Text, images, visible regions, typography, and intended priority form a candidate container. Priority inversions and disagreement between intended groups and region membership are proposed task-specific measures, rather than formulas inherited from the perceptual principle. Brief-matched posters and message-identification or recall tasks can ground and validate them.

\Needspace{7\baselineskip}
\paragraph{Editorial design.}
Rhetorical Structure Theory supplies a theory of discourse functions~\citep{mann1987rst}, and spatial contiguity relates text--figure placement to comprehension~\citep{mayer1999instructional}. Document hierarchy, reading order, references, page geometry, and language direction provide the analytical objects. Reading-order violations, broken references, and text--figure separation can be evaluated against complete publications with equivalent content, then checked through retrieval time and comprehension.

\paragraph{Scientific and mechanism diagrams.}
Congruence and apprehension connect graphical organization to conceptual understanding~\citep{tversky2002animation}; Systems Biology Graphical Notation (SBGN) supplies typed semantic and layout rules for biological mechanisms~\citep{rougny2019sbgn}. Entities, states, processes, compartments, and directed interactions form a domain container. Conformance checks inherit the standard; a weighted violation score is an additional operational choice. Expert mechanism specifications and process-tracing tasks separately evaluate scientific correctness and comprehension; other sciences require their own ontologies.

\paragraph{Data visualization and maps.}
Graphical expressiveness formalizes faithful encoding, while effectiveness concerns task-dependent perceptual readability~\citep{mackinlay1986graphical}; projection theory quantifies geographic distortion~\citep{snyder1987projections}. Typed fields, marks, scales, geographic features, units, and projection parameters expose these properties. Encoding checks and projection equations provide existing analytical foundations. A combined perceptual score requires its own task assumptions and evidence, rather than treating all effectiveness rankings as established quantitative laws.

\paragraph{UI and interaction design.}
The Keystroke-Level Model (KLM) predicts time for a specified expert, routine, error-free method~\citep{card1980keystroke}. Ordered operations, devices, mental preparation, and response delays therefore supply a directly quantitative container. Screens, controls, and executable transitions add task structure: reachability is a separate property, and error rates require observed interaction. Method recovery and calibrated timings connect the model to generated interfaces; broader usability and accessibility need corresponding users and outcome evidence.

\paragraph{Cinematic composition and storyboards.}
The Attentional Theory of Cinematic Continuity explains how attention and perceptual expectations support continuity across cuts~\citep{smith2012continuity}. Shots, events, characters, trajectories, gaze targets, and cut types expose that organization. Action-phase mismatch and target displacement are candidate operationalizations, not ready-made scores supplied by the theory. Their interpretation requires cut-specific continuity goals, timed sequences, and gaze or comprehension evidence, including cases where discontinuity is intentional.

\paragraph{Exhibition and wayfinding design.}
Exhibition studies distinguish accessibility from directed cross-visibility~\citep{peponis2004exhibitions}, while interpretive guidance links information to visible site features~\citep{nps2019wayside}. Separate movement and visibility graphs can represent exhibits, paths, signs, destinations, occlusion, and thematic links. Route success, engagement, and interpretive understanding provide stronger grounding than visibility alone.

\subsubsection{Spatial and Physical Design}
\label{app:design-operators-physical}

Physical relations must be interpreted under use conditions: loads for parts, bodies and materials for garments, climate for landscapes, and movement rules for playable levels. Reference artifacts provide organizational context; simulation, testing, or observed use supplies performance evidence.

\paragraph{Floorplans and interiors.}
Space syntax formalizes access organization through spaces, connections, and integration~\citep{hillier1984sociallogic,hillier1996space}. SSIO already recovers this structure and relates functional integration to residential reference distributions. An interior extension could add furniture, activity locations, and mobility envelopes, grounding activity reachability and route length in furnished plans and observed use.

\paragraph{Urban and landscape design.}
Space syntax relates street-network configuration to pedestrian movement and land-use attraction~\citep{hillier1993naturalmovement}. Street segments, crossings, destinations, transit stops, and pedestrian evidence would support accessibility and service coverage. A separate performance-modeling branch could use InVEST to estimate urban heat mitigation from land cover, shade, albedo, and evapotranspiration under reference climate conditions~\citep{bosch2021urbancooling}; the planning brief determines which branch defines improvement.

\paragraph{Industrial products and mechanical CAD.}
Product architecture maps functions to components and interfaces~\citep{ulrich1995architecture}. Axiomatic Design adds a quantitative foundation: functional requirements and design parameters form a dependency matrix, while independence and information axioms analyze their coupling and probability of satisfying requirements~\citep{suh1996axiomaticchapter}. A container can preserve these typed dependencies alongside parts, joints, materials, loads, and tolerances. Recovering functional assignments and estimating requirement-success probabilities are the adaptation tasks; mechanical simulation supplies purpose-specific performance evidence~\citep{autodeskstaticstress}.

\paragraph{Garments.}
The functional, expressive, and aesthetic framework organizes apparel needs but does not itself provide quantitative fit or comfort formulas~\citep{lamb1992apparel}. Panels, seams, bodies, materials, and activity provide a candidate container; GarmentCodeData supplies pattern representations and draping checks~\citep{korosteleva2024garmentcodedata}. Pattern feasibility is already computable. Connecting ease, fit, movement, or expression to the framework requires matched bodies, fabrics, activities, and wearer evidence.

\paragraph{Textiles and surface patterns.}
TexGen constructs yarn geometry for mechanical and permeability simulations~\citep{brown2022texgen}, whereas plane-group symmetry formalizes motif repetition~\citep{iucrspacegrouppatterns}. These purposes require distinct containers: yarn paths, interlacing, contacts, material properties, and loads for physical performance; motifs, repeat boundaries, and symmetry for surface organization. Manufactured swatches or repeat-ready designs provide the corresponding references.

\paragraph{Packaging.}
Fold kinematics offers a narrow quantitative foundation: faces and hinge relations can determine whether a specified package admits the required folding motion~\citep{balkcom2006paperbags}. FEFCO supplies a construction vocabulary and ISTA supplies distribution-performance tests~\citep{fefco2022code,istatestprocedures}; these are complementary standards. Panels, folds, tabs, closures, inserts, and product geometry expose assembly and retention. Protection and opening objectives additionally require material, product, shipment, and use evidence.

\paragraph{Game levels.}
Mission--space graph grammars provide a formal route from progression structure to playable space~\citep{dormans2010adventures}. Regions, hazards, resources, gates, prerequisites, and state-dependent transitions form a container for progression checks. MDA relates mechanics and dynamics to player experience~\citep{hunicke2004mda}, while expressive-range analysis characterizes generator distributions~\citep{smith2010expressiverange}. Reachability and prerequisite consistency are directly testable; challenge and experiential improvement require a declared player model and play traces.

\subsection{Agent-Assisted Translation and Worked Examples}
\label{app:agent-theory-translation}

\begin{figure}[!ht]
  \centering
  \includegraphics[width=\linewidth]{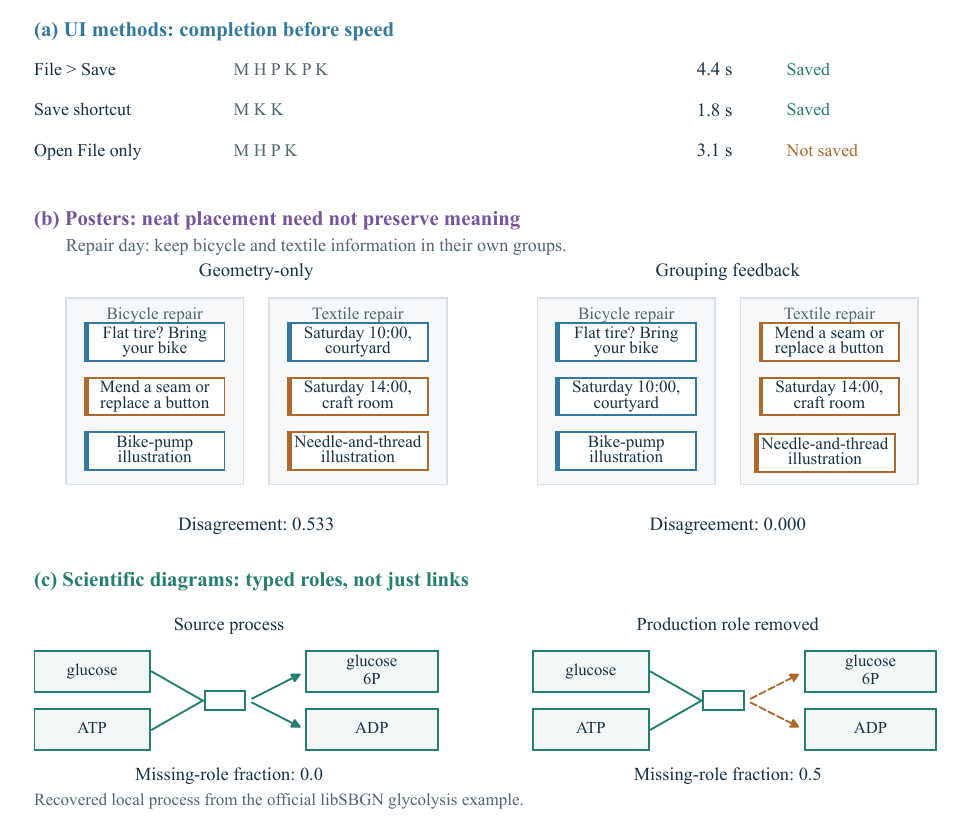}
  \caption{\textbf{Worked theory-translation examples.} (a) Task completion qualifies KLM time comparisons; M/H/P/K denote mental preparation, homing, pointing, and keystrokes. (b) Two editors start from the same development poster. Colors identify intended groups; image contents appear as labels in the recovered container. (c) A process recovered from the libSBGN glycolysis example; dashed arcs mark deleted production relations.}
  \label{fig:worked-theory-translation}
\end{figure}

Three examples connect design goals to computable feedback: task completion and predicted effort in UI using KLM~\citep{card1980keystroke}; typed process roles in scientific diagrams using SBGN~\citep{rougny2019sbgn}; and intended poster grouping using an authored metric motivated by common-region perception~\citep{palmer1992commonregion}. Figure~\ref{fig:worked-theory-translation} shows their analytical objects and controlled changes. The examples show how recovered design structure supports task-specific evaluation and feedback-guided editing, without generator training or human-performance evaluation.

\begingroup\interlinepenalty=10000
\paragraph{UI: completing the task before reducing effort.}
Three authored tasks---saving a document, formatting a selection, and exporting a PDF---each have a menu method, a shortcut, and an incomplete alternative. An executable state model verifies the requested result independently of the KLM time calculation. For saving, supplied operation times predict 4.4 seconds for the menu method and 1.8 for the shortcut; merely opening File takes 3.1 seconds but leaves the document unsaved. Across nine methods, six complete their tasks and three incomplete alternatives are excluded from time-based improvement. The reference is a complete method for the same task, not a pooled distribution of unrelated interfaces.
\par
\endgroup

\paragraph{Scientific diagrams: recovering the meaning of a connection.}
Three official \href{https://github.com/sbgn/libsbgn/tree/2aae05bb5c28b6d61757f8f3123bea505d57a20b/example-files}{libSBGN example maps}---glycolysis, MAPK, and insulin-like growth-factor signaling---supply 19 compatible irreversible process neighborhoods. Removing an entire consumption or production role is detected in all 38 paired changes; all 38 wrong-direction or wrong-type variants are rejected. However, 16 single-arc deletions leave another arc fulfilling the same role and do not change the score. Thus, role completeness differs from preserving every reaction participant: the latter requires a richer, task-specific account of the process, not simply more connectivity.

\paragraph{Feedback-guided poster repair.}
Seven authored event briefs use six content elements in two or three panels: three development briefs and four evaluation briefs, with each brief's variants kept together. We compare no editing, geometry-only editing, and grouping-feedback editing. Both editors evaluate 22 candidates per round for four rounds (88 evaluations per editor and brief), using the same operations: no change, all pair swaps, and individual grid snaps. Geometry minimizes grid displacement and overlap; grouping feedback first minimizes pairwise disagreement between intended groups and displayed regions, then geometry. Equal scores retain the current design.

\begin{table}[!ht]
  \centering
  \caption{\textbf{Poster repair under a shared editing budget.} Means over four authored evaluation briefs. Named-panel accuracy is evaluated separately from the grouping objective.}
  \label{tab:worked-poster-repair}
  \small
  \setlength{\tabcolsep}{7pt}
  \begin{tabular}{lccc}
    \toprule
    Selection criterion & Group disagreement $\downarrow$ & Grid error $\downarrow$ & Named panel $\uparrow$ \\
    \midrule
    No editing & 0.350 & 0.463 & 58.3\% \\
    Geometry only & 0.350 & 0.093 & 58.3\% \\
    Grouping feedback & 0.083 & 0.278 & 79.2\% \\
    \bottomrule
  \end{tabular}
\end{table}

\FloatBarrier
Grouping feedback reduces mean disagreement from 0.350 to 0.083, whereas geometry-only editing improves alignment without repairing groups (Table~\ref{tab:worked-poster-repair}). Both preserve content and avoid overlap or containment errors; grouping repair leaves more alignment error under the shared budget. The library poster reaches zero disagreement with only one-third of its elements in their intended named panels, distinguishing grouping from semantic assignment. The clinic's groups of four and two cannot fit two three-slot panels: exhaustive enumeration confirms that disagreement of $1/3$ is unavoidable without changing the template. Effective feedback therefore needs panel semantics, and the edit space must permit the intended organization.

\paragraph{A reusable implementation workflow.}
Researchers specify the design purpose, objects, relations, and measurement sources. Agents then help implement the container, matched references, and feedback-guided editing. In the worked examples, equivalent formats and renamed elements preserve structure and scores; poster grouping also remains unchanged under whole-layout translation and scaling. Implementation checks cover finite scores, context matching, and reference/test separation. The accompanying materials provide the workflow instructions, implementation, sources, and results.

Assessing design gains and reductions in expert effort will require experiments with trained generators and expert users.

\end{document}